\documentclass[jair,twoside,11pt,theapa]{article}
\usepackage{jair, theapa, rawfonts}
\usepackage[utf8]{inputenc}
\usepackage{url}
\usepackage{amsmath}
\usepackage{amssymb}
\usepackage{mathtools}
\usepackage{esdiff}
\usepackage[noend]{algpseudocode}
\usepackage{algorithm}
\usepackage{tikz}
\usepackage{tikz-qtree}
\usepackage{neuralnetwork}
\usepackage{listings}

\newcommand{\m}[1]{\mathbf{#1}}%
\newcommand{\ygcomment}[1]{\textbf{[TODO: #1]}}
\renewcommand{\ygcomment}[1]{}
\renewcommand{\shortcite}[0]{\citeyear}
\newcommand{\ignore}[1]{}

  \usepackage{courier}
\DeclareMathOperator*{\argmax}{arg\,max}

\title{A Primer on Neural Network Models\\ for Natural Language Processing}
\author{Yoav Goldberg \\ \small{Draft as of \today.}}

\begin{document}

\maketitle

\ygcomment{
data size: the compression work requires 2m sentences.
collobert and weston
    }

\fbox{
\parbox{0.9\textwidth}{%
\small{
The most up-to-date version of this manuscript is available at
\url{http://www.cs.biu.ac.il/~yogo/nnlp.pdf}. Major updates will be published on
arxiv periodically.\\  I welcome any comments you may have regarding the content and
presentation.  If you spot a missing reference or have relevant work you'd
like to see mentioned, do let me know.\\ \texttt{first.last@gmail}}}}
\vspace{1em}
\vspace{2em}

\begin{abstract}
    Over the past few years, neural networks have re-emerged as powerful
    machine-learning models, yielding state-of-the-art results in fields such as image
    recognition and speech processing. More recently, neural network models
    started to be applied also to textual natural language signals, again with very
    promising results.  This tutorial surveys neural network models from the
    perspective of natural language processing research, in an attempt to bring
    natural-language researchers up to speed with the neural techniques. The
    tutorial covers input encoding for natural language tasks, feed-forward
    networks, convolutional networks, recurrent networks and recursive networks,
    as well as the computation graph abstraction for
    automatic gradient computation.
\end{abstract}

\section{Introduction}
For a long time, core NLP techniques were dominated by machine-learning
approaches that used linear models such as support vector machines or logistic
regression, trained over very high dimensional yet very sparse feature vectors. 

Recently, the field has seen some success in switching from such linear models
over sparse inputs to non-linear neural-network models over dense inputs.  While
most of the neural network techniques are easy to apply, sometimes as almost
drop-in replacements of the old linear classifiers, there is in many cases a
strong barrier of entry.  In this tutorial I attempt to provide NLP
practitioners (as well as newcomers) with the basic background, jargon, tools
and methodology that will allow them to understand the principles behind the
neural network models and apply them to their own work.  This tutorial is
expected to be self-contained, while presenting the different approaches under a
unified notation and framework.  It repeats a lot of material which is available
elsewhere. It also points to external sources for more advanced topics when
appropriate.

This primer is not intended as a comprehensive resource for those that will go
on and develop the next advances in neural-network machinery (though it may
serve as a good entry point). Rather, it is aimed at those readers who are
interested in taking the existing, useful technology and applying it in useful
and creative ways to their favourite NLP problems.  For more in-depth, general
discussion of neural networks, the theory behind them, advanced optimization
methods and other advanced topics, the reader is referred to other existing
resources. In particular, the book by Bengio et al \shortcite{bengio2015deep} is highly
recommended.

\paragraph{Scope} The focus is on applications of neural networks to language
processing tasks.  However, some subareas of language processing with neural
networks were decidedly left out of scope of this tutorial.
These include the vast literature of language modeling and acoustic modeling,
the use of neural networks for machine translation, and multi-modal applications
combining language and other signals such as images and videos
(e.g. caption generation).  Caching methods for efficient runtime performance,
methods for efficient training with
large output vocabularies and attention models are also not discussed.
Word embeddings are discussed only to the extent that is needed to understand in order
to use them as inputs for other models.  Other unsupervised approaches, including
autoencoders and recursive autoencoders, also fall out of scope.
While some
applications of neural networks for language modeling and machine translation are
mentioned in the text, their treatment is by no means comprehensive.

\paragraph{A Note on Terminology}
The word ``feature'' is used to refer to a concrete, linguistic input such
as a word, a suffix, or a part-of-speech tag. For example, in a first-order
part-of-speech tagger, the features might be ``current word, previous word, next
word, previous part of speech''. The term ``input vector'' is used to refer to
the actual input that is fed to the neural-network classifier. Similarly,
``input vector entry'' refers to a specific value of the input. This is in contrast to
a lot of the neural networks literature in which the word ``feature'' is overloaded
between the two uses, and is used primarily to refer to an input-vector entry.

\paragraph{Mathematical Notation} I use bold upper case letters to represent
matrices ($\m{X}$, $\m{Y}$, $\m{Z}$), and bold lower-case letters to 
represent vectors ($\m{b}$). When there
are series of related matrices and vectors (for example, where each matrix
corresponds to a different layer in the network), superscript indices are used
($\m{W^1}$, $\m{W^2}$). For the rare cases in which we want indicate the power of a
matrix or a vector, a pair of brackets is added around the item to be
exponentiated: $(\m{W})^2, (\m{W^3})^2$. 
Unless otherwise stated, vectors are assumed to be row vectors.
We use $[\m{v_1};\m{v_2}]$ to denote vector concatenation.

\clearpage
\section{Neural Network Architectures}

Neural networks are powerful learning models.
We will discuss two kinds of neural network architectures, that can be mixed and
matched -- feed-forward networks and Recurrent / Recursive networks. 
Feed-forward networks include networks with fully connected layers, such as the
multi-layer perceptron, as well as networks with convolutional and pooling
layers. All of the networks act as classifiers, but each with different
strengths.

Fully connected feed-forward neural networks (Section \ref{sec:ff}) are non-linear learners that
can, for the most part, be used as a drop-in replacement wherever a linear
learner is used. This includes binary and multiclass classification problems,
as well as more complex structured prediction problems (Section
\ref{sec:structured}). The non-linearity of the
network, as well as the ability to easily integrate pre-trained word embeddings,
often lead to superior classification accuracy.  A series of works
\cite{chen2014fast,weiss2015structured,pei2015effective,durrett2015neural} managed to obtain improved syntactic parsing
results by simply replacing the linear model of a parser with a fully connected 
feed-forward network.  Straight-forward applications of a feed-forward network as a
classifier replacement (usually coupled with the use of pre-trained word
vectors) provide
benefits also for CCG supertagging \cite{lewis2014improved}, dialog state tracking \cite{henderson2013deep}, pre-ordering
for statistical machine translation \cite{degispert2015fast} and language
modeling \cite{bengio2003neural,vaswani2013decoding}.  Iyyer et al
\shortcite{iyyer2015deep} demonstrate that multi-layer feed-forward networks can
provide competitive results on sentiment classification and factoid question
answering.

Networks with convolutional and pooling layers (Section \ref{sec:convnet}) are useful for classification tasks in which
we expect to find strong local clues regarding class membership, but these clues
can appear in different places in the input. For example, in a document
classification task, a single key phrase (or an ngram) can help in determining the
topic of the document \cite{johnson2015effective}.  We would like to learn that certain sequences of words
are good indicators of the topic, and do not necessarily care where they appear
in the document.  Convolutional and pooling layers allow the model to learn to
find such local indicators, regardless of their position.  Convolutional and
pooling architecture show promising results on many tasks, including document classification
\cite{johnson2015effective}, short-text categorization \cite{wang2015semantic},
sentiment classification \cite{kalchbrenner2014convolutional,kim2014convolutional},
relation type classification between entities
\cite{zeng2014relation,dossantos2015classifying}, event detection
\cite{chen2015event,nguyen2015event}, paraphrase identification \cite{yin2015convolutional}
semantic role labeling \cite{collobert2011natural}, question answering
\cite{dong2015question}, predicting box-office revenues of movies based on
critic reviews \cite{bitvai2015nonlinear} modeling text interestingness
\cite{gao2014modeling}, and modeling the relation between character-sequences
and part-of-speech tags {\cite{santos2014learning}.

In natural language we often work with structured data of arbitrary sizes, 
such as sequences and trees.  We would like to be able to capture regularities
in such structures, or to model similarities between such structures.  In many
cases, this means encoding the structure as a fixed width vector, which we can
then pass on to another statistical learner for further processing. While
convolutional and pooling architectures allow us to encode arbitrary large items
as fixed size vectors capturing their most salient features, they do so by sacrificing
most of the structural information. Recurrent (Section \ref{sec:rnn}) and
recursive (Section \ref{sec:recnn}) architectures, on
the other hand, allow us to work with sequences and trees while preserving a lot
of the structural information. Recurrent networks \cite{elman1990finding} are designed to model
sequences, while recursive networks \cite{goller1996learning} are generalizations of recurrent networks
that can handle trees.  We will also discuss an extension of recurrent networks that
allow them to model stacks
\cite{dyer2015transitionbased,watanabe2015transitionbased}. 

Recurrent models have been shown to produce very
strong results for language modeling, including
\cite{mikolov2010recurrent,mikolov2011extensions,mikolov2012statistical,duh2013adaptation,adel2013combination,auli2013joint,auli2014decoder};
as well as for sequence tagging 
\cite{irsoy2014opinion,xu2015ccg,ling2015finding}, machine translation
\cite{sundermeyer2014translation,tamura2014recurrent,sutskever2014sequence,cho2014learning},
dependency parsing \cite{dyer2015transitionbased,watanabe2015transitionbased},
sentiment analysis \cite{wang2015predicting}, noisy text normalization
\cite{chrupala2014normalizing}, dialog state tracking
\cite{mrksic2015multidomain}, response generation \cite{sordoni2015neural}, and
modeling the relation between character sequences and part-of-speech
tags \cite{ling2015finding}.  \ygcomment{mention the multi-modal stuff?}

Recursive models were shown to produce state-of-the-art or near state-of-the-art results for 
constituency 
\cite{socher2013parsing} and dependency \cite{le2014insideoutside,zhu2015reranking}
parse re-ranking, discourse parsing \cite{li2014recursive}, semantic relation classification
\cite{hashimoto2013simple,liu2015dependencybased},
political ideology detection based on parse trees \cite{iyyer2014political},
sentiment classification \cite{socher2013recursive,hermann2013role},
target-dependent sentiment classification \cite{dong2014adaptive} and
question answering \cite{iyyer2014neural}.

\clearpage
\section{Feature Representation}

Before discussing the network structure in more depth, it is important to pay
attention to how features are represented.
For now, we can think of a feed-forward neural network as a function $NN(\m{x})$ that
takes as input a $d_{in}$ dimensional vector $\m{x}$ and produces a $d_{out}$
dimensional output vector. The function is often used as a \emph{classifier},
assigning the input $\m{x}$ a degree of membership in one or more of $d_{out}$
classes.  The function can be complex, and is almost always
non-linear.  Common structures of this function will be discussed in Section
\ref{sec:ff}. Here, we focus on the input, $\m{x}$.
When dealing with natural language, the 
input $\m{x}$ encodes features such as words, part-of-speech tags or other
linguistic information. Perhaps the biggest jump when
moving from sparse-input linear models to neural-network based models is to stop
representing each feature as a unique dimension (the so called one-hot
representation) and representing them instead as dense vectors.  That is,
each core feature is \emph{embedded} into a $d$ dimensional
space, and
represented as a vector in that space.\footnote{Different feature types may be embedded into different spaces.
For example, one may represent word features using 100 dimensions, and
part-of-speech features using 20 dimensions.}
The embeddings (the vector
representation of each core feature) can then be trained like the other parameter
of the function $NN$. Figure \ref{fig:sparse-vs-dense} shows the two approaches
to feature representation.

\begin{figure}[t]
\begin{center}
\includegraphics[width=0.7\textwidth]{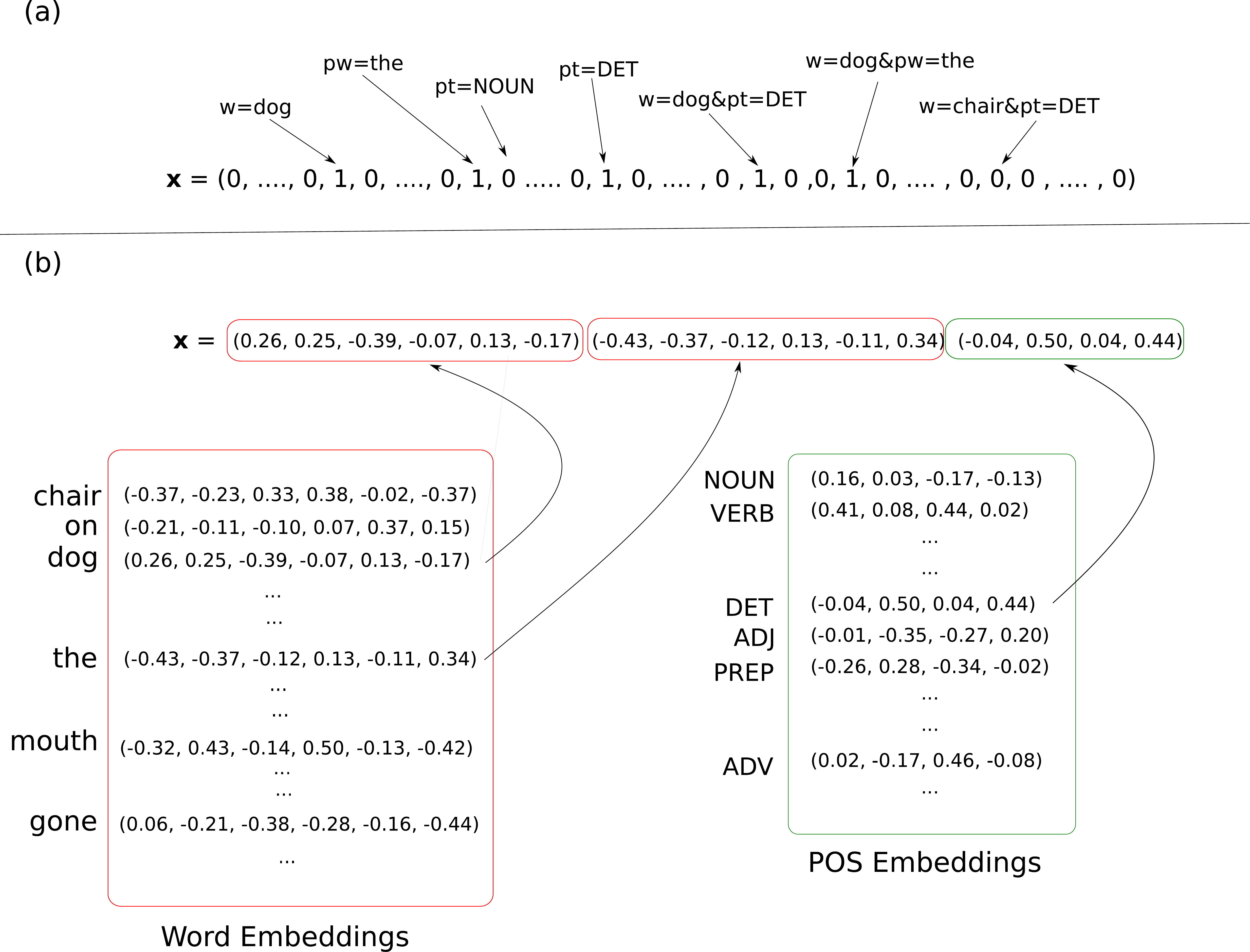}
\end{center}
\caption{\textbf{Sparse vs. dense feature representations}. 
Two encodings of the information: \emph{current word is ``dog''; previous word is ``the''; previous pos-tag is ``DET''}.
(a) Sparse feature vector.
Each dimension represents a feature. Feature combinations receive their own
dimensions. Feature values are binary. Dimensionality is very high.
(b) Dense, embeddings-based feature vector. Each core feature is represented as a
vector. Each feature corresponds to several input vector entries. No explicit
encoding of feature combinations. Dimensionality is low. The feature-to-vector
mappings come from an embedding table.}
\label{fig:sparse-vs-dense}
\end{figure}

The feature embeddings (the values of the vector entries for each feature) are treated as \emph{model
parameters} that need to be trained together with the other components of the
network. Methods of training (or obtaining) the feature embeddings will be
discussed later. For now, consider the feature embeddings as given.

The general structure for an NLP classification system based on a feed-forward neural network is thus: 
\begin{enumerate}
    \item Extract a set of core linguistic features $f_1,\dots,f_k$ that are relevant
        for predicting the output class.
    \item For each feature $f_i$ of interest, retrieve the corresponding vector $v(f_i)$. 
    \item Combine the vectors (either by concatenation, summation or a
        combination of both) into an input vector $\m{x}$.
    \item Feed $\m{x}$ into a non-linear classifier (feed-forward neural network).
\end{enumerate}

The biggest change in the input, then, is the move from sparse representations in which each
feature is its own dimension, to a dense representation in which each feature
is mapped to a vector.  Another difference is that we extract only \emph{core
features} and not feature combinations.  We will elaborate on both these changes
briefly.

\paragraph{Dense Vectors vs. One-hot Representations} 

What are the benefits of representing our features as vectors instead of as
unique IDs? Should we always represent features as dense vectors?
Let's consider the two kinds of representations:
\begin{description}
    \item[\textbf{One Hot}] Each feature is its own dimension.
        \begin{itemize}
            \item Dimensionality of one-hot vector is same as number of distinct
                features.
            \item Features are completely independent from one another.  The
                feature ``word is `dog' '' is as dis-similar to ``word is
                `thinking' ''
                than it is to ``word is `cat' ''.
        \end{itemize}
    \item[\textbf{Dense}] Each feature is a $d$-dimensional vector.
        \begin{itemize}
            \item Dimensionality of vector is $d$.
            \item Similar features will have similar vectors --
                information is shared between similar features.
        \end{itemize}
\end{description}

One benefit of using dense and low-dimensional vectors is computational: the
majority of neural network toolkits do not play well with
very high-dimensional, sparse vectors. However, this is just a technical
obstacle, which can be resolved with some engineering effort.

The main benefit of the dense representations is in generalization power: if we
believe some features may provide similar clues, it is worthwhile to provide a
representation that is able to capture these similarities.
For example, assume we have observed the word `dog' many times during
training, but only observed the word `cat' a handful of times, or not at all.
If each of the words is associated with its own dimension, occurrences of
`dog' will not tell us anything about the occurrences of `cat'. However, in
the dense vectors representation the learned vector for `dog' may be similar
to the learned vector from `cat', allowing the model to share statistical
strength between the two events.  This argument assumes that ``good'' vectors are
somehow given to us. Section \ref{sec:word-embed} describes ways of
obtaining such vector representations.

In cases where we have relatively few distinct features in the category, and we
believe there are no correlations between the different features, we may use the
one-hot representation. However, if we believe there are going to be correlations
between the different features in the group (for example, for part-of-speech
tags, we may believe that the different verb inflections \texttt{VB} and
\texttt{VBZ} may
behave similarly as far as our task is concerned) it may be worthwhile to let
the network figure out the correlations and gain some statistical strength by
sharing the parameters.
It may be the case that under some circumstances, when
the feature space is relatively small and the training data is plentiful, or
when we do not wish to share statistical information between distinct words, there
are gains to be made from using the one-hot representations. However, this is
still an open research question, and there are no strong evidence to either side.
The majority of work (pioneered by
\cite{collobert2008unified,collobert2011natural,chen2014fast})
advocate the use of dense, trainable embedding vectors for all features.
For work using neural network architecture with sparse vector encodings see
\cite{johnson2015effective}.

Finally, it is important to note that representing features as dense vectors
is an integral part of the neural network framework, and that consequentially 
the differences between using sparse and dense feature representations are
subtler than they may appear at first.  In fact, using sparse, one-hot vectors
as input when training a neural network amounts to dedicating the first layer
of the network to learning a dense embedding vector for each feature 
based on the training data.  We touch on this in Section \ref{sec:embed-layer}.

\paragraph{Variable Number of Features: Continuous Bag of Words}
Feed-forward networks assume a fixed dimensional input.
This can easily accommodate the case of a feature-extraction function that extracts a fixed
number of features: each feature is represented as a vector, and the vectors are
concatenated. This way, each region of the resulting input vector corresponds to a
different feature. However, in some cases the number of features is not known in
advance (for example, in document classification it is common that each word in
the sentence is a feature). 
We thus need to represent an unbounded number of features using a fixed size
vector.
One way of achieving this is through a so-called \emph{continuous bag of words}
(CBOW) representation \cite{mikolov2013efficient}. The CBOW is very similar to the traditional bag-of-words
representation in which we discard order information, and works by either summing or
averaging the
embedding vectors of the corresponding features:\footnote{Note that if the
$v(f_i)$s were one-hot vectors rather than dense feature representations, the
$CBOW$ and $WCBOW$ equations above would reduce to the traditional (weighted) bag-of-words
representations, which is in turn equivalent to a sparse feature-vector
representation in which
each binary indicator feature corresponds to a unique ``word''.}

\[CBOW(f_1,...,f_k) = \frac{1}{k}\sum_{i=1}^{k} v(f_i) \] 

A simple variation on the CBOW representation is weighted CBOW, in which different vectors receive different weights:

\[WCBOW(f_1,...,f_k) = \frac{1}{\sum_{i=1}^{k}a_i}\sum_{i=1}^{k} a_i v(f_i) \]

Here, each feature $f_i$ has an associated weight $a_i$, indicating the relative
importance of the feature. For example, in a document classification task, a
feature $f_i$ may correspond to a word in the document, and the associated
weight $a_i$ could be the word's TF-IDF score.

\paragraph{Distance and Position Features} The linear distance in between two words in a
sentence may serve as an informative feature. For example, in an event
extraction task\footnote{The event extraction task involves identification of
events from a predefined set of event types. For example identification of
``purchase'' events or ``terror-attack'' events.  Each event type can be
triggered by various triggering words (commonly verbs), and has several slots
(arguments)
that needs to be filled (i.e. who purchased? what was purchased? at what
amount?).}
we may be given a trigger word and a candidate argument word, and asked to
predict if the argument word is indeed an argument of the trigger. The distance (or relative position)
between the trigger and the argument is a strong signal for this prediction task.  In the
``traditional'' NLP setup, distances are usually encoded by binning the
distances into several groups (i.e. 1, 2, 3, 4, 5--10, 10+) and associating each bin
with a one-hot vector.  In a neural architecture, where the input vector is not composed of binary indicator features, it may seem natural to allocate a single
input vector entry to the distance feature, where the numeric value of that entry is the distance. However,
this approach is not taken in practice. Instead,
distance features are encoded
similarly to the other feature types: each bin is associated with a
$d$-dimensional vector, and these distance-embedding vectors are then trained as
regular parameters in the
network \cite{zeng2014relation,dossantos2015classifying,zhu2015reranking,nguyen2015event}.

\paragraph{Feature Combinations} Note that the feature extraction stage in the
neural-network settings deals only with extraction of \emph{core} features.
This is in contrast to the traditional linear-model-based NLP systems in which the
feature designer had to manually specify not only the core features of interests but
also interactions between them (e.g., introducing not only a feature stating
``word is X'' and a feature
stating ``tag is Y'' but also combined feature stating ``word is X and tag is Y'' or
sometimes even ``word is X, tag is Y and previous word is Z'').  The combination
features are crucial in linear models because they introduce more
dimensions to the input, transforming it into a space where the data-points are
closer to being linearly separable. On the other hand, the space of possible
combinations is very large, and the feature designer has to spend a lot of time
coming up with an effective set of feature combinations.\ygcomment{also not
stat-efficient. very high dim.}
One of the
promises of the non-linear neural network models is that one needs to define 
only the core features. The non-linearity of the classifier, as defined by the
network structure, is expected to take care of finding the indicative feature
combinations, alleviating the need for feature combination engineering.

Kernel methods \cite{shawe-taylor2004kernel}, and in particular polynomial
kernels \cite{kudo2003fast}, also allow the
feature designer to specify only core features, leaving the feature combination
aspect to the learning algorithm.  In contrast to neural-network models, kernels
methods are convex, admitting exact solutions to the optimization problem.
However, the classification efficiency in kernel methods scales linearly with the size
of the training data, making them too slow for most practical purposes, and not
suitable for training with large datasets.
On the other hand, neural network
classification efficiency scales linearly with the size of the network,
regardless of the training data size.

\paragraph{Dimensionality} How many dimensions should we allocate for each
feature? Unfortunately, there are no theoretical bounds or even established best-practices
in this space.
Clearly, the dimensionality should grow with the number of the members in the class
(you probably want to assign more dimensions to word embeddings than to
part-of-speech embeddings) but how much is enough? In current research, the
dimensionality of word-embedding vectors range between about 50 to a few hundreds,
and, in some extreme cases, thousands. Since the dimensionality of the vectors
has a direct effect on memory requirements and processing time, a good rule of
thumb would be to experiment with a few different sizes, and choose a good
trade-off between speed and task accuracy.

\paragraph{Vector Sharing} Consider a case where you have a few features that
share the same vocabulary. For example, when assigning a part-of-speech to a
given word, we may have a set of features considering the
previous word, and a set of features considering the next word. When building
the input to the classifier, we will concatenate the vector representation of
the previous word to the vector representation of the next word. The classifier
will then be able to distinguish the two different indicators, and treat them
differently. But should the two features share the same vectors? Should the
vector for ``dog:previous-word'' be the same as the vector of ``dog:next-word''?
Or should we assign them two distinct vectors?
This, again, is mostly an empirical question. If you believe words behave
differently when they appear in different positions (e.g., word X behaves like
word Y when in the previous position, but X behaves like Z when in the next
position) then it may be a good idea to use two different vocabularies and
assign a different set of vectors for each feature type. However, if you believe
the words behave similarly in both locations, then something may be gained by
using a shared vocabulary for both feature types.

\paragraph{Network's Output} For multi-class classification problems with $k$
classes, the
network's output is a $k$-dimensional vector in which every dimension represents
the strength of a particular output class. That is, the output remains as in the
traditional linear models -- scalar scores to items in a discrete set.  However,
as we will see in Section
\ref{sec:ff}, there is a $d \times k$ matrix associated with the
output layer. The columns of this matrix can be thought of as $d$ dimensional
embeddings of the output classes.  The vector similarities between the vector
representations of the $k$ classes indicate the model's learned similarities between the
output classes.

\paragraph{Historical Note}  Representing words as dense vectors for input to a
neural network was introduced by Bengio et al \cite{bengio2003neural} in the
context of neural language modeling.  It was introduced to NLP tasks in the
pioneering work of Collobert, Weston and colleagues
\shortcite{collobert2008unified,collobert2011natural}.  Using embeddings for
representing not only words but arbitrary features was popularized following Chen and Manning \shortcite{chen2014fast}.

\clearpage
\section{Feed-forward Neural Networks}
\label{sec:ff}
\ygcomment{linear vs nonlinear, the xor example\ldots}

\paragraph{A Brain-inspired metaphor} As the name suggest, neural-networks are
inspired by the brain's computation mechanism, which consists of computation
units called neurons. In the metaphor, a neuron is a computational unit that has
scalar inputs and outputs.  Each input has an associated weight. The neuron
multiplies each input by its weight, and then sums\footnote{While summing is the
most common operation, other functions, such as a max, are also possible} them,
applies a non-linear function to the result, and passes it to its output. 
The neurons are connected to each other, forming a network:  the output of a
neuron may feed into the inputs of one or more neurons.  Such networks were shown to
be very capable computational devices.  If the weights are set correctly, a neural network
with enough neurons and a non-linear activation function can approximate a very wide range of mathematical functions (we will be more precise
about this later).

\begin{figure}[h!]
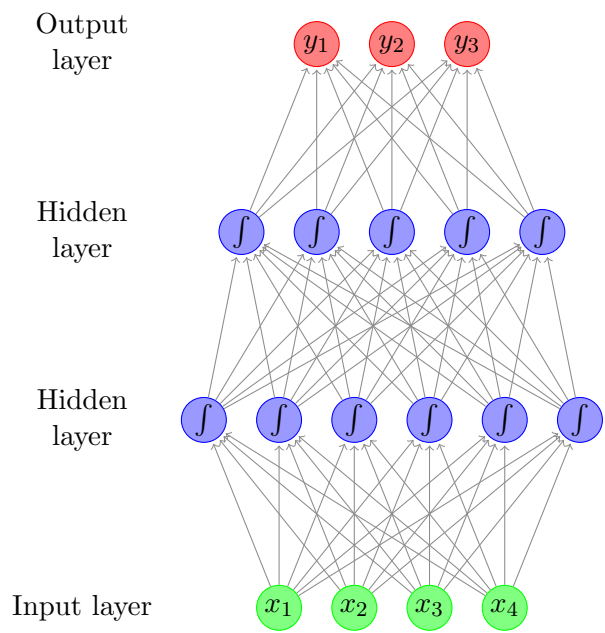

\begin{center}
\begin{neuralnetwork}[height=4,style={rotate=90},layertitleheight=5.5em,toprow=true]
		\newcommand{\nodetextclear}[2]{}
		\newcommand{\nodetextx}[2]{$x_#2$}
		\newcommand{\nodetexty}[2]{$y_#2$}
		\newcommand{\nodetexts}[2]{$\int$}
		\inputlayer[count=4, bias=false, title=Input layer, text=\nodetextx]
		\hiddenlayer[count=6, bias=false, title=Hidden layer, text=\nodetexts] \linklayers
		\hiddenlayer[count=5, bias=false, title=Hidden layer, text=\nodetexts] \linklayers
		\outputlayer[count=3, title=Output layer, text=\nodetexty] \linklayers
\end{neuralnetwork}
\end{center}
\caption{Feed-forward neural network with two hidden layers.}
\label{fig:nn1}
\end{figure}

A typical feed-forward neural network may be drawn as in Figure \ref{fig:nn1}.
Each circle is a neuron, with incoming arrows being the neuron's inputs and
outgoing arrows being the neuron's outputs.  Each arrow carries a weight,
reflecting its importance (not shown).
Neurons are arranged in layers,
reflecting the flow of information.  The bottom layer has no incoming arrows,
and is the input to the network.  The top-most layer has no outgoing arrows, and
is the output of the network. The other layers are considered ``hidden''.
The sigmoid shape inside the neurons in the
middle layers represent a non-linear function (typically a $1/(1+e^{-x}$)) that
is applied to the neuron's value before passing it to the output.
In the figure, each neuron is connected to all of the neurons in the next layer
-- this is called a \emph{fully-connected layer} or an \emph{affine layer}.

While the brain metaphor is sexy and
intriguing, it is also distracting and cumbersome to manipulate mathematically.
We therefore switch to using more concise mathematic notation.
The values of each row of neurons in the network can be thought of as a vector.  In Figure
\ref{fig:nn1} the input layer is a $4$ dimensional vector ($\m{x}$), and the
layer above it is a $6$ dimensional vector ($\m{h^1}$).  
The fully connected layer can be
thought of as a linear transformation from $4$ dimensions to $6$ dimensions.
A fully-connected layer implements a vector-matrix multiplication, $\m{h} = \m{x}\m{W}$ where
the weight of the connection from the $i$th neuron in the input row to the $j$th
neuron in the output row is $W_{ij}$.\footnote{To see why this is the case,
denote the weight of the $i$th input of the $j$th neuron in $\m{h}$ as $w_{ij}$.
The value of $h_j$ is then $h_j = \sum_{i=1}^{4}x_i\cdot w_{ij}$.} The values of
$\m{h}$ are then transformed by a non-linear function $g$ that is applied to
each value before being passed on to the next input. The whole computation from
input to output can be written as: $(g(\m{x}\m{W^1}))\m{W^2}$ where $\m{W^1}$
are the weights of the first layer and $\m{W^2}$ are the weights of the second
one.

\paragraph{In Mathematical Notation}
From this point on, we will abandon the brain metaphor and describe networks exclusively
in terms of vector-matrix operations.

\noindent The simplest neural network is the perceptron, which is a linear function of its inputs:

\[
NN_{Perceptron}(\m{x}) = \m{x}\m{W} + \m{b} 
\]
\[
\m{x} \in \mathbb{R}^{d_{in}}, \;\; \m{W} \in \mathbb{R}^{d_{in} \times d_{out}}, \;\; \m{b} \in \mathbb{R}^{d_{out}}
\]

\noindent $\m{W}$ is the weight matrix, and $\m{b}$ is a bias term.\footnote{The
network in figure \ref{fig:nn1} does not include bias terms. A bias term can be
added to a layer by adding to it an additional neuron that does not have any incoming connections,
whose value is always $1$.}
In order to go beyond linear functions, we introduce a non-linear hidden layer (the network in Figure \ref{fig:nn1} has two such layers),
resulting in the 1-layer Multi Layer Perceptron (MLP1).
A one-layer feed-forward neural network has the form:

\[ NN_{MLP1}(\m{x}) = g(\m{x}\m{W^1} + \m{b^1})\m{W^2} + \m{b^2} \]
\[ \m{x} \in \mathbb{R}^{d_{in}}, \;\;
   \m{W^1} \in \mathbb{R}^{d_{in} \times d_{1}}, \;\; \m{b^1} \in \mathbb{R}^{d_{1}}, \;\;
   \m{W^2} \in \mathbb{R}^{d_{1} \times d_{2}}, \;\; \m{b^2} \in \mathbb{R}^{d_{2}} 
\]

Here $\m{W^1}$ and $\m{b^1}$ are a matrix and a bias term for the first linear
transformation of the input, $g$ is a non-linear function that is applied element-wise 
(also called a \emph{non-linearity} or an \emph{activation function}), and
$\m{W^2}$ and $\m{b^2}$ are the matrix and bias term for a second linear transform.

Breaking it down, $\m{x}\m{W^1}+\m{b^1}$ is a linear transformation of the input $\m{x}$
from $d_{in}$ dimensions to $d_1$ dimensions. $g$ is then applied to each of the
$d_1$ dimensions, and the matrix $\m{W^2}$ together with bias vector $\m{b}^2$ are then
used to transform the result into the $d_2$ dimensional output vector.
The non-linear activation function $g$ has a crucial role in the network's
ability to represent complex functions. Without the non-linearity in $g$, the
neural network can only represent linear transformations of the
input.\footnote{To see why, consider that a sequence of linear transformations
is still a linear transformation.}

We can add additional linear-transformations and non-linearities, resulting in a
2-layer MLP (the network in Figure \ref{fig:nn1} is of this form):

\[ NN_{MLP2}(\m{x}) = (g^2(g^1(\m{x}\m{W^1} + \m{b^1})\m{W^2}+\m{b^2}))\m{W^3} \]

\noindent It is perhaps clearer to write deeper networks like this using intermediary
variables:

\begin{align*}
    NN_{MLP2}(\m{x}) =& \m{y}\\
    \m{h^1} =& g^1(\m{x}\m{W^1}+\m{b^1}) \\
    \m{h^2} =& g^2(\m{h^1}\m{W^2} + \m{b^2}) \\
    \m{y} =& \m{h^2} \m{W^3}
\end{align*}

The vector resulting from each linear transform is referred to as a \emph{layer}.
The outer-most linear transform results in the \emph{output layer} and the other
linear transforms result in \emph{hidden layers}. Each hidden layer is followed
by a non-linear activation.
In some cases, such as in the last layer of our example,
the bias vectors are forced to 0 (``dropped'').

\ygcomment{repeated sentence from start of section}
Layers resulting from linear transformations are often referred to as \emph{fully
connected}, or \emph{affine}.
Other types of architectures exist.
In particular, image
recognition problems benefit from \emph{convolutional} and \emph{pooling} layers. 
Such layers have uses also in language processing, and will be discussed in
Section \ref{sec:convnet}.  Networks with more than one hidden layer are said to
be \emph{deep} networks, hence the name \emph{deep learning}.

When describing a neural network, one should specify the \emph{dimensions} of
the layers and the input. A layer will expect a $d_{in}$ dimensional vector as
its input, and transform it into a $d_{out}$ dimensional vector. The
dimensionality of the layer is taken to be the dimensionality of its output.
For a fully connected layer $l(\m{x}) = \m{x}\m{W}+\m{b}$ with input
dimensionality $d_{in}$ and output dimensionality $d_{out}$, the dimensions of
$\m{x}$ is $1\times d_{in}$, of $\m{W}$ is $d_{in}\times d_{out}$ and of $\m{b}$
is $1\times d_{out}$.

The output of the network is a $d_{out}$ dimensional vector. In case $d_{out} =
1$, the network's output is a scalar. Such networks can be used for regression
(or scoring)
by considering the value of the output, or for binary
classification by consulting the sign of the output.  Networks with $d_{out} =
k > 1$ can be used for $k$-class classification, by associating each dimension
with a class, and looking for the dimension with maximal value.  Similarly, if
the output vector entries are positive and sum to one, the output can be
interpreted as a distribution over class assignments (such output 
normalization is typically achieved by applying a softmax transformation on
the output layer, see Section \ref{sec:softmax}).
\ygcomment{Added this paragraph per Miguel's comment, but is it needed here?
repetition..}

The matrices and the bias terms that define the linear transformations are the
\emph{parameters} of the network. It is common to refer to the collection of all
parameters as $\theta$. Together with the input, the parameters
determine the network's output. The training algorithm is responsible for
setting their values such that the network's predictions are correct.
Training is discussed in Section \ref{sec:training}.

\subsection{Representation Power} In terms of representation power, it was shown
by \cite{hornik1989multilayer,cybenko1989approximation} that MLP1 is a universal
approximator -- it can approximate with any desired non-zero amount of error a
family of functions\footnote{Specifically, a feed-forward network with linear
output layer and at least one hidden layer with a ``squashing'' activation function
can approximate any Borel measurable function from one finite dimensional space to another.} 
that include all continuous functions on a closed and bounded
subset of $\mathbb{R}^n$, and any function mapping from any finite dimensional
discrete space to another. This may suggest there is no reason to go
beyond MLP1 to more complex architectures. However, the theoretical result does not state 
how large the hidden layer should be, nor does it say anything about the
learnability of the neural network (it states that a representation exists, but
does not say how easy or hard it is to set the parameters based on training data
and a specific learning algorithm).  It also does not guarantee that a
training algorithm will find the \emph{correct} function generating our training data.
Since in practice we train neural networks on relatively small amounts of data,
using a combination of the backpropagation algorithm and variants of stochastic gradient descent,
and use hidden layers of relatively modest sizes (up to several thousands), there is
benefit to be had in trying out more complex architectures than MLP1.
In many
cases, however, MLP1 does indeed provide very strong results.
For further discussion on the representation power of feed-forward neural
networks, see \cite[Section 6.5]{bengio2015deep}.

\subsection{Common Non-linearities}

The non-linearity $g$ can take many forms. There is currently no good theory as
to which non-linearity to apply in which conditions, and choosing the correct
non-linearity for a given task is for the most part an empirical question. I
will now go over the common non-linearities from the literature: the sigmoid,
tanh, hard tanh and the rectified linear unit (ReLU). Some NLP researchers also
experimented with other forms of non-linearities such as cube and tanh-cube.

\paragraph{Sigmoid} The sigmoid activation function $\sigma(x) = 1/(1+e^{-x})$ is an
S-shaped function, transforming each value $x$ into the range $[0,1]$.

\paragraph{Hyperbolic tangent (tanh)} The hyperbolic tangent $tanh(x) =
\frac{e^{2x}-1}{e^{2x}+1}$  activation
function is an S-shaped function, transforming the values $x$ into the range $[-1,1]$.

\paragraph{Hard tanh} The hard-tanh activation function is an approximation of
the $tanh$ function which is faster to compute and take derivatives of:
\begin{align*}
    hardtanh(x) = \begin{cases}
        -1 &  x < -1 \\
        1  &  x > 1 \\
        x  & \text{otherwise}
    \end{cases}
\end{align*}

\paragraph{Rectifier (ReLU)} The Rectifier activation function \cite{glorot2011deep}, also known as
the rectified linear unit is a very simple activation function that is easy to
work with and was shown many times to produce excellent results.%
\footnote{The technical 
advantages of the ReLU over the sigmoid and tanh activation functions is that it
does not involve expensive-to-compute functions, and more importantly that it 
does not saturate.  The sigmoid and tanh activation are capped at $1$, and
the gradients at this region of the functions are near zero, driving the entire
gradient near zero.  The ReLU activation does not have this problem, making it
especially suitable for networks with multiple layers, which are 
susceptible to the vanishing gradients problem when trained with the saturating
units.}
The ReLU unit
clips each value $x < 0$ at 0.  Despite its simplicity, it performs well for many
tasks, especially when combined with the dropout regularization technique 
(see Section \ref{sec:dropout}).

\begin{align*}
    ReLU(x) = \max(0,x) = \begin{cases}
        0 & x < 0 \\
        x & \text{otherwise}
    \end{cases}
\end{align*}

As a rule of thumb, ReLU units work better than tanh, and tanh works better than
sigmoid.\footnote{
In addition to these activation functions, recent works from the NLP community
experiment with and reported success with other forms of non-linearities.
The \textbf{Cube} activation function, $g(x) = (x)^3$, was suggested by
\cite{chen2014fast}, who found it to be more effective than other non-linearities
in a feed-forward network that was used to predict the actions in a greedy
transition-based dependency parser.
The \textbf{tanh cube} activation function $g(x) = \tanh( (x)^3 +
x)$ was proposed by \cite{pei2015effective},
who found it to be more effective than other non-linearities
in a feed-forward network that was used as a component in a
structured-prediction graph-based dependency parser.

The cube and tanh-cube activation functions are motivated by the desire
to better capture interactions between different features.
While these activation functions are reported to improve performance in certain
situations, their general applicability is still to be determined.}

\subsection{Output Transformations}
\label{sec:softmax}

In many cases, the output layer vector is also transformed.
A common transformation is the \emph{softmax}:

\begin{align*}
           \m{x} =& x_1,\ldots,x_k \\
    softmax(x_i) =& \frac{e^{x_i}}{\sum_{j=1}^{k} e^{x_j}} 
\end{align*}

The result is a vector of non-negative real numbers that sum to one, making it a
discrete probability distribution over $k$ possible outcomes.

The $softmax$ output transformation is used when we are
interested in modeling a probability distribution over the possible output
classes.  To be effective, it should be used in conjunction with a probabilistic
training objective such as cross-entropy (see Section \ref{sec:crossent} below).

When the softmax transformation is applied to the output of a network without a
hidden layer, the result is the well known multinomial logistic regression
model, also known as a maximum-entropy classifier.

\subsection{Embedding Layers}
\label{sec:embed-layer}

Up until now, the discussion ignored the source of $\m{x}$, treating it as an arbitrary
vector. In an NLP application, $\m{x}$ is usually composed of various embeddings
vectors.
We can be explicit about the source of $\m{x}$, and include it in the network's definition.
We introduce $c(\cdot)$, a function from core features to an input vector.

It is common for $c$ to extract the embedding vector associated with each feature, and concatenate them:

\begin{align*}
    \m{x} = c(f_1,f_2,f_3) =& [v(f_1);v(f_2);v(f_3)] \\
    NN_{MLP1}(\m{x}) =& NN_{MLP1}(c(f_1,f_2,f_3)) \\
    =& NN_{MLP1}([v(f_1);v(f_2);v(f_3)]) \\
    =& (g([v(f_1);v(f_2);v(f_3)]\m{W^1} + \m{b^1}))\m{W^2} + \m{b^2}  
\end{align*}

Another common choice is for $c$ to sum the embedding vectors (this assumes the
embedding vectors all share the same dimensionality):

\begin{align*}
    \m{x} = c(f_1,f_2,f_3) =& v(f_1)+v(f_2)+v(f_3) \\
    NN_{MLP1}(\m{x}) =& NN_{MLP1}(c(f_1,f_2,f_3)) \\ 
    =& NN_{MLP1}(v(f_1)+v(f_2)+v(f_3)) \\
    =& (g((v(f_1)+v(f_2)+v(f_3))\m{W^1} + \m{b^1}))\m{W^2} + \m{b^2}  
\end{align*}


The form of $c$ is an essential part of the network's design.
In many papers, it is common to refer to $c$ as part of the network,
and likewise treat the word embeddings $v(f_i)$ as resulting
from an ``embedding layer'' or ``lookup layer''. Consider a vocabulary of $|V|$
words, each embedded as a $d$ dimensional vector. The collection of vectors can
then be thought of as a $|V| \times d$ embedding matrix $\m{E}$ in which each row
corresponds to an embedded feature.
Let $\m{f_i}$ be a $|V|$-dimensional vector, which is all zeros except from one
index, corresponding to the value of the $i$th feature, in which the value is
$1$ (this is called a one-hot vector). The multiplication $\m{f_i}\m{E}$ will then
``select'' the corresponding row of $\m{E}$. Thus, $v(f_i)$ can be defined in
terms of $\m{E}$ and $\m{f_i}$:
\[ v(f_i) = \m{f_i} \m{E}  \]

And similarly:

\[ CBOW(f_1,...,f_k) = \sum_{i=1}^{k} (\m{f_i} \m{E}) = (\sum_{i=1}^{k} \m{f_i}) \m{E} \]

The input to the network is then considered to be a collection of one-hot
vectors.  While this is elegant and well defined mathematically, an efficient
implementation typically involves a hash-based data structure mapping features
to their corresponding embedding vectors, without going through the one-hot
representation.

In this tutorial, we take $c$ to be separate from the
network architecture: the network's inputs are always dense real-valued input
vectors, and $c$ is applied before the input is passed the network, similar to a
``feature function'' in the familiar linear-models terminology.  However, when
training a network, the input vector $\m{x}$ does remember how it was
constructed, and can propagate error gradients back to its component embedding
vectors, as appropriate. 

\paragraph{A note on notation} When describing network layers that get
concatenated vectors $\m{x}$, $\m{y}$ and $\m{z}$ as input, some authors
use explicit concatenation
($[\m{x};\m{y};\m{z}]\m{W} + \m{b}$) while others use an affine transformation
($\m{x}\m{U} + \m{y}\m{V} + \m{z}\m{W} + \m{b})$.  If the weight matrices
$\m{U}$, $\m{V}$, $\m{W}$ in the affine transformation are different than one
another, the two notations are equivalent.

\paragraph{A note on sparse vs. dense features} Consider a network which uses
a ``traditional'' sparse representation for its input vectors,
and no embedding layer.  
Assuming the set of all available features is $V$ and we have $k$ ``on''
features $f_1,\ldots,f_k$, $f_i \in V$,
the network's input is:
\[
\m{x} = \sum_{i=1}^{k} \m{f_i} \;\;\;\;\;\;\;\;\;\; \m{x} \in \mathbb{N}_+^{|V|}
\] 
\noindent and so the first layer (ignoring the non-linear activation) is:
\[
\m{x}\m{W} + \m{b} = (\sum_{i=1}^{k} \m{f_i})\m{W}
\]
\[
\m{W} \in \mathbb{R}^{|V|\times d}, \;\;\; \m{b}\in\mathbb{R}^d
\]

This layer selects rows of $\m{W}$ corresponding to the input features in
$\m{x}$ and sums them, then adding a bias term.  This is very similar to
an embedding layer that produces a CBOW representation over the features,
where the matrix $\m{W}$ acts as the embedding matrix.  The main difference
is the introduction of the bias vector $\m{b}$, and the fact that the embedding
layer typically does not undergo a non-linear activation but rather passed on
directly to the first layer. Another difference is that this scenario forces
each feature to receive a separate vector (row in $\m{W}$) while the embedding
layer provides more flexibility, allowing for example for the features ``next
word is dog'' and ``previous word is dog'' to share the same vector. However,
these differences are small and subtle. When it comes to multi-layer
feed-forward networks, the difference between dense and sparse inputs
is smaller than it may seem at first sight.

\subsection{Loss Functions}

When training a neural network (more on training in Section \ref{sec:training} below),
much like when training a linear classifier, 
one defines a loss function $L(\m{\hat{y}},\m{y})$, stating the loss of
predicting $\m{\hat{y}}$ when the true output is $\m{y}$.  The training
objective is then to minimize the loss across the different training examples.
The loss $L(\m{\hat{y}},\m{y})$ assigns a numerical score (a scalar) for the
network's output $\m{\hat{y}}$ given the true expected output
$\m{y}$.\footnote{In our notation, both the model's output and the expected
output are vectors, while in many cases it is more natural to think of the
expected output as a scalar (class assignment). In such cases, $\m{y}$ is simply
the corresponding one-hot vector.} The loss is always positive, and should be
zero only for cases where the network's output is correct.

The parameters of the network (the matrices $\m{W^i}$, the biases $\m{b^i}$ and
commonly the embeddings $\m{E}$) are then set in order to minimize the loss $L$
over the training examples (usually, it is the sum of the losses over the
different training examples that is being minimized).

The loss can be an arbitrary function mapping two vectors to a scalar. For
practical purposes of optimization, we restrict ourselves to functions for which
we can easily compute gradients (or sub-gradients).  In most cases, it is
sufficient and advisable to rely on a common loss function rather than defining
your own.  For a detailed discussion on loss functions for neural networks see
\cite{lecun2006tutorial,lecun2005loss,bengio2015deep}.
We now discuss some loss functions that are commonly used in neural networks for
NLP.

\paragraph{Hinge (binary)} 
For binary classification problems, the network's output is a single scalar
$\hat{y}$ and the intended output $y$ is in $\{+1,-1\}$.  The classification rule
is $sign(\hat{y})$, and a classification is
considered correct if $y\cdot\hat{y} > 0$, meaning that $y$ and $\hat{y}$ share
the same sign. 
The hinge loss, also known as margin loss or SVM loss, is defined as:
\begin{align*}
    L_{hinge(binary)}(\hat{y},y) = \max(0, 1 - y\cdot\hat{y})
\end{align*}

The loss is 0 when $y$ and $\hat{y}$ share the same sign and $|\hat{y}| \geq 1$.
Otherwise, the loss is linear.  In other words, the binary hinge loss attempts to achieve a
correct classification, with a margin of at least 1.

\paragraph{Hinge (multiclass)} 
The hinge loss was extended to the multiclass setting by Crammer and Singer
\shortcite{crammer2002algorithmic}.
Let $\m{\hat{y}} = \hat{y}_1,\ldots,\hat{y}_n$ be the network's output vector, and
$\m{y}$ be the one-hot vector for the correct output class.

The classification rule is defined as
selecting the class with the highest score:
\begin{align*}
    \text{prediction} = \arg\max_i \hat{y}_i
\end{align*},

Denote by $t = \argmax_i y_i$ the correct
class, and by $k = \argmax_{i \neq t} \hat{y}_i$ the highest scoring class such that $k\neq t$. 
The multiclass hinge loss is defined as:
\begin{align*}
    L_{hinge(multiclass)}(\m{\hat{y}},\m{y}) = \max(0, 1 - (\hat{y}_t -
    \hat{y}_k))
\end{align*}

\noindent The multiclass hinge loss attempts to score the correct class above
all other classes with a margin of at least 1.

Both the binary and multiclass hinge losses are intended to be used with a
linear output layer.  The hinge losses are useful whenever we require a hard
decision rule, and do not attempt to model class membership probability.

\paragraph{Log loss} The log loss is a common variation of the hinge loss, which
can be seen as a ``soft'' version of the hinge loss with an infinite margin
\cite{lecun2006tutorial}.

\begin{align*}
    L_{log}(\m{\hat{y}},\m{y}) = \log(1 + exp(-(\hat{y}_t - \hat{y}_k) )
\end{align*}

%
%
%

\paragraph{Categorical cross-entropy loss} \label{sec:crossent}
The categorical cross-entropy loss (also referred to as \emph{negative log
likelihood}) is
used when a probabilistic interpretation of the scores is desired.

Let $\m{y}=y_1,\ldots,y_n$ be a vector representing
the true multinomial distribution over the labels $1,\ldots,n$, and let
$\m{\hat{y}}=\hat{y_1},\ldots,\hat{y_n}$ be the network's output, which was
transformed by the $softmax$ activation function, and represent the class
membership conditional distribution $\hat{y}_i = P(y=i|\m{x})$.
The categorical cross entropy loss measures the dissimilarity between the true label distribution
$\m{y}$ and the predicted label distribution $\m{\hat{y}}$, and is defined as
cross entropy:

\begin{align*}
    L_{cross-entropy}(\m{\hat{y}},\m{y}) = -\sum_i y_i\log(\hat{y}_i)
\end{align*}

For hard classification problems in which each training example has a single
correct class assignment, $\m{y}$ is a one-hot vector representing the true
class. In such cases, the cross entropy can be simplified to:

\begin{align*}
    L_{cross-entropy(\text{hard classification})}(\m{\hat{y}},\m{y}) = -\log(\hat{y}_t)
\end{align*}

\noindent where $t$ is the correct class assignment. This attempts to set the
probability mass assigned to the correct class $t$ to 1. Because the scores
$\m{\hat{y}}$ have been transformed using the $softmax$ function and represent a
conditional distribution, increasing the mass assigned to the correct class
means decreasing the mass assigned to all the other classes.

The cross-entropy loss is very common in the neural networks literature, and
produces a multi-class classifier which does not only predict the one-best
class label but but also predicts a distribution over the possible labels.  When
using the cross-entropy loss, it is assumed that the network's output is transformed
using the $softmax$ transformation.

\paragraph{Ranking losses}  In some settings, we are not given supervision in
term of labels, but rather as pairs of correct and incorrect items $\m{x}$ and
$\m{x'}$, and our goal is to score correct items above incorrect ones. Such
training situations arise when we have only positive examples, and generate
negative examples by corrupting a positive example.  
A useful loss in such scenarios is the margin-based ranking
loss, defined for a pair of correct and incorrect examples:

\begin{align*}
    L_{ranking(margin)}(\m{x},\m{x'}) = \max(0, 1 - (NN(\m{x}) - NN(\m{x'})))
\end{align*}

\noindent where $NN(\m{x})$ is the score assigned by the network for input vector
$\m{x}$.  The objective is to score (rank) correct inputs over incorrect ones with a
margin of at least 1.  

A common variation is to use the log version of the ranking loss:

\begin{align*}
    L_{ranking(log)}(\m{x},\m{x'}) = \log(1 + exp(-(NN(\m{x}) - NN(\m{x'}))))
\end{align*}

Examples using the ranking hinge loss in language tasks include training with the
auxiliary tasks used for deriving pre-trained word embeddings
(see section \ref{sec:word-embed}), in which we are given a correct word sequence and a
corrupted word sequence, and our goal is to score the correct sequence above
the corrupt one \cite{collobert2008unified}.  Similarly, Van de Cruys \shortcite{vandecruys2014neural} used the
ranking loss in a selectional-preferences task, in which the network was trained
to rank correct verb-object pairs above incorrect, automatically derived ones,
and \cite{weston2013connecting} trained a model to score correct
(head,relation,trail) triplets above corrupted ones in an information-extraction
setting.  An example of using the ranking log loss can be found in
\cite{gao2014modeling}.  A variation of the ranking log loss allowing for a
different margin for the negative and positive class is given in \cite{dossantos2015classifying}.
\ygcomment{Do we really need dos santos here?}

\clearpage
\section{Word Embeddings}
\label{sec:word-embed}

A main component of the neural-network approach is the use of embeddings -- representing each feature as a vector in a low dimensional space.
But where do the vectors come from? This section will survey the common approaches.

\subsection{Random Initialization}

When enough supervised training data is available, one can just treat the feature
embeddings the same as the other model parameters: initialize the embedding
vectors to random values, and let the network-training procedure tune them into ``good'' vectors. 

Some care has to be taken in the way the random initialization is performed. The
method used by the effective word2vec implementation
\cite{mikolov2013efficient,mikolov2013distributed} is to initialize the word
vectors to uniformly sampled random numbers in the range $[-\frac{1}{2d},
\frac{1}{2d}]$ where $d$ is the number of dimensions.  Another option is to use
\emph{xavier initialization} (see Section \ref{sec:glorot-init}) and initialize with
uniformly
sampled values from $\left[-\frac{\sqrt{6}}{\sqrt{d}}, \frac{\sqrt{6}}{\sqrt{d}}\right]$.

In practice, one will often use the random initialization approach to initialize
the embedding vectors of commonly occurring features, such as part-of-speech
tags or individual letters, while using some form of supervised or unsupervised
pre-training to initialize the potentially rare features, such as features for
individual words.  The pre-trained vectors can then either be treated as fixed
during the network training process, or, more commonly, treated like the
randomly-initialized vectors and further tuned to the task at hand.  

\subsection{Supervised Task-specific Pre-training}

If we are interested in task A, for which we only have a limited amount of
labeled data (for example, syntactic parsing), but there is an auxiliary task B
(say, part-of-speech tagging) for which we have much more labeled data, we may
want to pre-train our word vectors so that they perform well as predictors for task B, and
then use the trained vectors for training task A.  In this way, we can utilize
the larger amounts of labeled data we have for task B.  When training
task A we can either treat the pre-trained vectors as fixed, or tune them
further for task A.
Another option is to train jointly for both objectives, see Section \ref{sec:joint} for more details.

\subsection{Unsupervised Pre-training}

The common case is that we do not have an auxiliary task with large enough
amounts of annotated data (or maybe we want to help bootstrap the auxiliary task
training with better vectors). In such cases, we resort to ``unsupervised''
methods, which can be trained on huge amounts of unannotated text.  

The techniques for training the word vectors are essentially those of supervised
learning, but instead of supervision for the task that we care about, we instead
create practically unlimited number of supervised training instances from raw
text, hoping that the tasks that we created will match (or be close enough to) the
final task we care about.\footnote{The interpretation of creating auxiliary problems from
raw text is inspired by Ando and Zhang \cite{ando2005highperformance,ando2005framework}.}

The key idea behind the unsupervised approaches is that one would like the
embedding vectors of ``similar'' words to have similar vectors. While word
similarity is hard to define and is usually very task-dependent, the current
approaches derive from the distributional hypothesis \cite{harris1954distributional},
stating that \emph{words are similar if they appear in similar contexts}. The different
methods all create supervised training instances in which the goal is to either
predict the word from its context, or predict the context from the word.

An important benefit of training word embeddings on large amounts of unannotated
data is that it provides vector representations for words that do not appear in
the supervised training set.  Ideally, the representations for these words will
be similar to those of related words that do appear in the training set,
allowing the model to generalize better on unseen events.  It is thus desired
that
the similarity between word vectors learned by the unsupervised algorithm
captures the same aspects of similarity that are useful for performing the
intended task of the network.
\ygcomment{Improve: discuss changes, why do we even expect this to
work?}

Common unsupervised word-embedding algorithms include \texttt{word2vec}
\footnote{While often treated as a single algorithm, \texttt{word2vec} is
actually a software package including various training objectives, optimization
methods and other hyperparameters. See \cite{rong2014word2vec,levy2015improving} for a discussion.}
\cite{mikolov2013distributed,mikolov2013efficient}, \texttt{GloVe} \cite{pennington2014glove} and the Collobert and Weston
\shortcite{collobert2008unified,collobert2011natural} embeddings algorithm.  These models are inspired by neural
networks and are based on stochastic gradient training.  However, they are
deeply connected to another family of algorithms which evolved in the NLP and IR
communities, and that are based on matrix factorization (see \cite{levy2014neural,levy2015improving}
for a discussion).

Arguably, the choice of auxiliary problem (what is being predicted, based
on what kind of context) affects the resulting vectors much more than the learning
method that is being used to train them.  We thus focus on the different
choices of auxiliary problems that are available, and only skim over the details
of the training methods.  Several software packages for deriving word vectors
are available, including word2vec\footnote{\url{https://code.google.com/p/word2vec/}} and Gensim\footnote{\url{https://radimrehurek.com/gensim/}}
implementing the word2vec models with word-windows based contexts,
word2vecf\footnote{\url{https://bitbucket.org/yoavgo/word2vecf}} which is a modified version of word2vec allowing
the use of arbitrary contexts, and GloVe\footnote{\url{http://nlp.stanford.edu/projects/glove/}} implementing the GloVe model.
Many pre-trained word vectors are also available for download on the web.

While beyond the scope of this tutorial, it is worth noting that the word
embeddings derived by unsupervised training algorithms have a wide range of
applications in NLP beyond using them for initializing the word-embeddings layer
of a neural-network model.

\subsection{Training Objectives}

Given a word $w$ and its context $c$, different algorithms formulate different
auxiliary tasks.  In all cases, each word is represented as a $d$-dimensional
vector which is initialized to a random value. Training the model to perform the
auxiliary tasks well will result in good word embeddings for relating the words
to the contexts, which in turn will result in the embedding vectors for similar
words to be similar to each other.

Language-modeling inspired approaches such as those taken by
\cite{mikolov2013distributed,mnih2013learning} as well as GloVe \cite{pennington2014glove}
use auxiliary tasks in which the goal
is to predict the word given its context. This is posed in a probabilistic
setup, trying to model the conditional probability $P(w|c)$. 

Other approaches reduce the problem to that of binary classification.
In addition to the set $D$ of observed word-context pairs, a set
$\bar{D}$ is created from random words and context pairings.
The binary classification problem is then: does the given $(w,c)$ pair come from
$D$ or not? The approaches differ in how the set $\bar{D}$ is constructed, what is the
structure of the classifier, and what is the objective being optimized.
Collobert and Weston \shortcite{collobert2008unified,collobert2011natural} take a margin-based binary ranking
approach, training a feed-forward neural network to score correct $(w,c)$ pairs
over incorrect ones.
Mikolov et al \shortcite{mikolov2013distributed,goldberg2014word2vec} take instead a probabilistic version, training a
log-bilinear model to predict the probability $P( (w,c) \in D|w,c)$ that the pair come from the
corpus rather than the random sample.

\subsection{The Choice of Contexts}

In most cases, the contexts of a word are taken to be other words that appear in
its surrounding, either in a short window around it, or within the same
sentence, paragraph or document. In some cases the text is automatically parsed
by a syntactic parser, and the contexts are derived from the syntactic
neighbourhood induced by the automatic parse trees. Sometimes, the definitions
of words and context change to include also parts of words, such as prefixes or
suffixes.

Neural word embeddings originated from the world of language modeling, in which
a network is trained to predict the next word based on a sequence of preceding
words \cite{bengio2003neural}. There, the text is used to create auxiliary tasks in
which the aim is to predict a word based on a context the $k$ previous words.
While training for the language modeling auxiliary prediction problems indeed
produce useful embeddings, this approach is needlessly restricted by the
constraints of the language modeling task, in which one is allowed to look only
at the previous words.
If we do not care about language modeling but only about the resulting
embeddings, we may do better by ignoring this constraint and taking the context
to be a symmetric window around the focus word.

\subsubsection{Window Approach} The most common approach is a sliding window
approach, in which auxiliary tasks are created by looking at a sequence of
$2k+1$ words. The middle word is callled the \emph{focus word} and the $k$ words
to each side are the \emph{contexts}.  Then, either a single task is created in
which the goal is to predict the focus word based on all of the context words
(represented either using CBOW \cite{mikolov2013efficient} or vector concatenation
\cite{collobert2008unified}), or $2k$ distinct tasks are created, each pairing the focus
word with a different context word. The $2k$ tasks approach, popularized by
\cite{mikolov2013distributed} is referred to as a \emph{skip-gram} model.  Skip-gram
based approaches are shown to be robust and efficient to train
\cite{mikolov2013distributed,pennington2014glove}, and often produce state of the art results. 

\paragraph{Effect of Window Size} The size of the sliding window has a strong
effect on the resulting vector similarities. Larger windows tend to produce more
topical similarities (i.e. ``dog'', ``bark'' and ``leash'' will be grouped
together, as well as ``walked'', ``run'' and ``walking''), while smaller windows
tend to produce more functional and syntactic similarities (i.e. ``Poodle'',
``Pitbull'', ``Rottweiler'', or ``walking'',``running'',``approaching'').

\paragraph{Positional Windows} When using the CBOW or skip-gram context
representations, all the different context words within the window are treated
equally. There is no distinction between context words that are close to the
focus words and those that are farther from it, and likewise there is no
distinction between context words that appear before the focus words to context
words that appear after it.  Such information can easily be factored in by using
\emph{positional contexts}: indicating for each context word also its relative
position to the focus words (i.e. instead of the context word being ``the'' it
becomes ``the:+2'', indicating the word appears two positions to the right of
the focus word). The use of positional context together with smaller windows
tend to produce similarities that are more syntactic, with a strong tendency of
grouping together words that share a part of speech, as well as being
functionally similar in terms of their semantics. Positional vectors were shown
by \cite{ling2015twotoo} to be more effective than window-based vectors when
used to initialize networks for part-of-speech tagging and syntactic dependency
parsing. 

\paragraph{Variants} Many variants on the window approach are possible. One may
lemmatize words before learning, apply text normalization, filter too short or
too long sentences, or remove capitalization (see, e.g., the pre-processing
steps described in \cite{dossantos2014deep}. One may sub-sample part of the corpus,
skipping with some probability the creation of tasks from windows that have too common or too rare
focus words. The window size may be dynamic, using a different window size at
each turn. One may weigh the different positions in the window differently,
focusing more on trying to predict correctly close word-context pairs than
further away ones.  Each of these choices will effect the resulting vectors.
Some of these hyperparameters (and others) are discussed in \cite{levy2015improving}.

\subsubsection{Sentences, Paragraphs or Documents} Using a skip-grams (or CBOW)
approach, one can consider the contexts of a word to be all the other words that
appear with it in the same sentence, paragraph or document. This is equivalent
to using very large window sizes, and is expected to result in word vectors that
capture topical similarity (words from the same topic, i.e. words that one would
expect to appear in the same document, are likely to receive similar vectors).

\subsubsection{Syntactic Window} Some work replace the linear context within a
sentence with a syntactic one \cite{levy2014dependencybased,bansal2014tailoring}.
The text is automatically parsed using a
dependency parser, and the context of a word is taken to be the words that are
in its proximity in the parse tree, together with the syntactic relation by
which they are connected. Such approaches produce highly \emph{functional}
similarities, grouping together words than can fill the same role in a sentence
(e.g. colors, names of schools, verbs of movement). The grouping is also
syntactic, grouping together words that share an inflection \cite{levy2014dependencybased}.

\subsubsection{Multilingual} Another option is using multilingual,
translation based contexts \cite{hermann2014multilingual,faruqui2014improving}.
For example, given a large amount of
sentence-aligned parallel text, one can run a bilingual alignment model such as
the IBM model 1 or model 2 (i.e. using the GIZA++ software), and then use the
produced alignments to derive word contexts. Here, the context of a word
instance are the foreign language words that are aligned to it.  Such alignments
tend to result in synonym words receiving similar vectors.  Some authors work instead on
the sentence alignment level, without relying on word alignments. An appealing method
is to mix a monolingual window-based approach with a multilingual approach,
creating both kinds of auxiliary tasks.  This is likely to produce vectors that
are similar to the window-based approach, but reducing the somewhat undesired
effect of the window-based approach in which antonyms (e.g. hot and cold, high
and low) tend to receive similar vectors \cite{faruqui2014improving}.

\subsubsection{Character-based and Sub-word Representations} 
An interesting line of work attempts to derive the vector representation of a
word from the characters that compose it.  Such approaches are likely to be
particularly useful for tasks which are syntactic in nature, as the character
patterns within words are strongly related to their syntactic function.  These
approaches also have the benefit of producing very small model sizes (only
one vector for each character in the alphabet together with a handful of small
matrices needs to be stored), and being able to provide an embedding vector for
every word that may be encountered.  
dos Santos and Gatti \shortcite{dossantos2014deep} and dos Santos and Zadrozny
\shortcite{santos2014learning} model the embedding of a word using a
convolutional network (see Section \ref{sec:convnet}) over the characters.
Ling et al \shortcite{ling2015finding} model the embedding of a word using the
concatenation of the final states of two RNN (LSTM) encoders (Section \ref{sec:rnn}), one
reading the characters from left to right, and the other from right to left.
Both produce very strong results for part-of-speech tagging. The work of
Ballesteros et al \shortcite{ballesteros2015improved} show that the two-LSTMs encoding of
\cite{ling2015finding} is beneficial also for representing words in dependency
parsing of morphologically rich languages.

Deriving representations of words from the representations of their characters
is motivated by the \emph{unknown words problem} -- what do you do when you
encounter a word for which you do not have an embedding vector?  Working on the
level of characters alleviates this problem to a large extent, as the vocabulary
of possible characters is much smaller than the vocabulary of possible words.
However, working on the character level is very challenging, as the relationship
between form (characters) and function (syntax, semantics) in language is quite
loose.  Restricting oneself to stay on the character level may be an
unnecessarily hard constraint.
Some researchers propose a middle-ground, in which a word is represented as a
combination of a vector for the word itself with vectors of sub-word units that
comprise it.  The sub-word embeddings then help in sharing information between
different words with similar forms, as well as allowing back-off to the subword
level when the word is not observed.  At the same time, the models are not
forced to rely solely on form when enough observations of the word are available.
Botha and Blunsom \shortcite{botha2014compositional} suggest to
model the embedding vector of a word as a sum of the word-specific vector if such vector
is available, with vectors
for the different morphological components that comprise it (the components are
derived using Morfessor \cite{creutz2007unsupervised}, an unsupervised morphological
segmentation method).  Gao et al \cite{gao2014modeling} suggest using as core
features not only the word form itself but also a unique feature (hence a unique
embedding vector) for each of the letter-trigrams in the word.

\clearpage
\section{Neural Network Training}
\label{sec:training}

Neural network training is done by trying to minimize a loss function over a
training set, using a gradient-based method.  Roughly speaking, all training
methods work by repeatedly computing an estimate of the error over the dataset, computing
the gradient with respect to the error, and then moving the parameters in the direction
of the gradient.  Models differ in how the error estimate is computed, and how
``moving in the direction of the gradient'' is defined.  We describe the
basic algorithm, \emph{stochastic gradient descent} (SGD),
and then briefly mention the other approaches with pointers for further reading.
Gradient calculation is central to the approach. Gradients can be efficiently
and automatically computed using reverse mode differentiation on a computation
graph -- a general
algorithmic framework for automatically computing the gradient of any network
and loss function.

\subsection{Stochastic Gradient Training}

The common approach for training neural networks is using the stochastic gradient
descent (SGD) algorithm \cite{bottou2012stochastic,lecun1998efficient} or a
variant of it.
SGD is a general optimization algorithm. It receives a function $f$
parameterized by $\theta$,
a loss function, and desired input and output pairs. It then attempts
to set the parameters $\theta$ such that the loss of $f$ with respect to the training examples is small.
\noindent The algorithm works as follows:

\begin{algorithm}[h]
    \caption{Online Stochastic Gradient Descent Training}
   \label{alg:online-sgd}
\begin{algorithmic}[1]
    \State \textbf{Input:} Function $f(\m{x}; \theta)$ parameterized with parameters
    $\theta$.
    \State \textbf{Input:} Training set of inputs $\m{x_1},\ldots,\m{x_n}$ and outputs $\m{y_1},\ldots,\m{y_n}$.
    \State \textbf{Input:} Loss function $L$.
    \While {\text{stopping criteria not met} }
        \State Sample a training example $\m{x_i},\m{y_i}$
        \State Compute the loss $L(f(\m{x_i}; \theta), \m{y_i})$ \label{line:loss1}
    \State $\m{\hat{g}} \gets \text{ gradients of }  L(f(\m{x_i}; \theta), \m{y_i}) 
                \text{ w.r.t } \theta $ \label{line:grad1} 
    \State $\theta \gets \theta + \eta_k \m{\hat{g}} $ \label{line:update1}
   \EndWhile
   \State \Return $\theta$
\end{algorithmic}
\end{algorithm}

The goal of the algorithm is to set the parameters $\theta$ so as to minimize the
total loss $\sum_{i=1}^{n}L(f(\m{x_i}; \theta), \m{y_i})$ over the training set.
It works by repeatedly sampling a training example and computing the gradient
of the error on the example with respect to the parameters $\theta$ (line
\ref{line:grad1}) -- the input and expected output are assumed to be
fixed, and the loss is treated as a function of the parameters $\theta$.
The parameters $\theta$ are then updated in the direction of the gradient, scaled by 
a learning rate $\eta_k$ (line \ref{line:update1}). For further discussion on setting the learning rate, see Section \ref{sec:learning-rate}.

Note that the error calculated in line \ref{line:loss1} is based on a single
training example, and is thus just a rough estimate of the corpus-wide loss that
we are aiming to minimize.  The noise in the loss computation may result in
inaccurate gradients.  
A common way of reducing this noise is to estimate the error and the gradients
based on a sample of $m$ examples. This gives rise to the \emph{minibatch SGD} algorithm:

\begin{algorithm}[h]
    \caption{Minibatch Stochastic Gradient Descent Training}
   \label{alg:minibatch-sgd}
\begin{algorithmic}[1]
    \State \textbf{Input:} Function $f(\m{x}; \theta)$ parameterized with parameters
    $\theta$.
    \State \textbf{Input:} Training set of inputs $\m{x_1},\ldots,\m{x_n}$ and outputs $\m{y_1},\ldots,\m{y_n}$.
    \State \textbf{Input:} Loss function $L$.
    \While {\text{stopping criteria not met} }
        \State Sample a minibatch of $m$ examples $\{(\m{x_1},\m{y_1}),\ldots,(\m{x_m},\m{y_m})\}$
        \State $\hat{\m{g}} \gets 0$ \label{line:loopstart}
        \For {$i = 1 \text{ to } m$}
        \State Compute the loss $L(f(\m{x_i}; \theta), \m{y_i})$
        \State $\m{\hat{g}} \gets \m{\hat{g}} + \text{ gradients of }
        \frac{1}{m} L(f(\m{x_i}; \theta), \m{y_i}) \text{ w.r.t } \theta $ \label{line:grad} 
    \EndFor \label{line:loopend}
    \State $\theta \gets \theta + \eta_k \m{\hat{g}} $
   \EndWhile
   \State \Return $\theta$
\end{algorithmic}
\end{algorithm}

In lines \ref{line:loopstart} -- \ref{line:loopend} the algorithm estimates the
gradient of the corpus loss based on the minibatch.
After the loop, $\m{\hat{g}}$ contains the gradient estimate, and the parameters
$\theta$ are updated toward $\m{\hat{g}}$.
The minibatch size can vary in size from $m=1$ to $m=n$.  Higher values provide
better estimates of the corpus-wide gradients, while smaller values allow more
updates and in turn faster convergence.
Besides the improved accuracy of the gradients estimation,
the minibatch algorithm provides opportunities for improved training efficiency. For modest
sizes of $m$, some computing architectures (i.e. GPUs) allow an efficient parallel implementation
of the computation in lines \ref{line:loopstart}--\ref{line:loopend}.
With a small enough learning rate, SGD is guaranteed to converge to a global optimum
if the function is convex. However, it can also be used to optimize non-convex
functions such as neural-network. While there are no longer guarantees of finding a global optimum, the algorithm
proved to be robust and performs well in practice.
\ygcomment{``such as neural networks'': Gabi found confusing.}

When training a neural network, the parameterized function $f$ is the neural
network, and the parameters $\theta$ are the layer-transfer matrices, bias terms, embedding
matrices and so on.  The gradient computation is a key step in the SGD
algorithm, as well as in all other neural network training algorithms.
The question is, then, how to compute the gradients of the network's error with
respect to the parameters.
Fortunately, there is an easy solution in the form of the \emph{backpropagation
algorithm} \cite{rumelhart1986learning,lecun1998gradient}. The backpropagation algorithm is a fancy name for
methodologically computing the derivatives of a complex expression using the chain-rule,
while caching intermediary results.
More generally, the backpropagation algorithm is a special case of the
reverse-mode automatic differentiation algorithm \cite[Section
7]{neidinger2010introduction}, \cite{baydin2015automatic,bengio2012practical}.
The following section
describes reverse mode automatic differentiation in the context of the
\emph{computation graph} abstraction.

\paragraph{Beyond SGD} While the SGD algorithm can and often does
produce good results, more advanced algorithms are also available.
The \emph{SGD+Momentum} \cite{polyak1964methods} and \emph{Nesterov Momentum} \cite{sutskever2013importance}
algorithms are variants of SGD in which previous gradients are accumulated and
affect the current update.  Adaptive learning rate algorithms including AdaGrad
\cite{duchi2011adaptive}, AdaDelta \cite{zeiler2012adadelta}, RMSProp
\cite{tieleman2012lecture} and Adam \cite{kingma2014adam} are designed
to select the learning rate for each minibatch, sometimes on a per-coordinate
basis, potentially alleviating the need of fiddling with learning rate
scheduling.  For details of these algorithms, see the original papers or
\cite[Sections 8.3, 8.4]{bengio2015deep}.  As many neural-network software
frameworks provide implementations of these algorithms, it is easy and sometimes
worthwhile to try out different variants.

\subsection{The Computation Graph Abstraction}

While one can compute the gradients of the various parameters of a network by
hand and implement them in code, this procedure is cumbersome and error prone.
For most purposes, it is preferable to use automatic tools for gradient
computation \cite{bengio2012practical}. The computation-graph abstraction allows
us to easily construct arbitrary networks, evaluate their predictions for given
inputs (forward pass), and compute gradients for their parameters with respect
to arbitrary scalar losses (backward pass).

A computation graph is a representation of an arbitrary mathematical computation as
a graph. It is a directed acyclic graph (DAG) in which nodes correspond to
mathematical operations or (bound) variables and edges correspond to the flow of
intermediary values between the nodes. The graph structure defines the order of
the computation in terms of the dependencies between the different components.
The graph is a DAG and not a tree, as the result of one operation can be the
input of several continuations. Consider for example a 
graph for the computation of $(a*b+1)*(a*b+2)$:
\begin{center}
\includegraphics[width=0.2\textwidth]{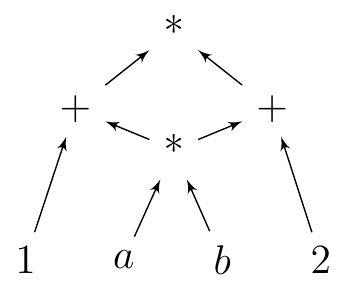}
\end{center}

\noindent The computation of $a*b$ is shared.
We restrict ourselves to the case where the computation graph is
connected.

\begin{figure}[ht]
\includegraphics[width=\textwidth]{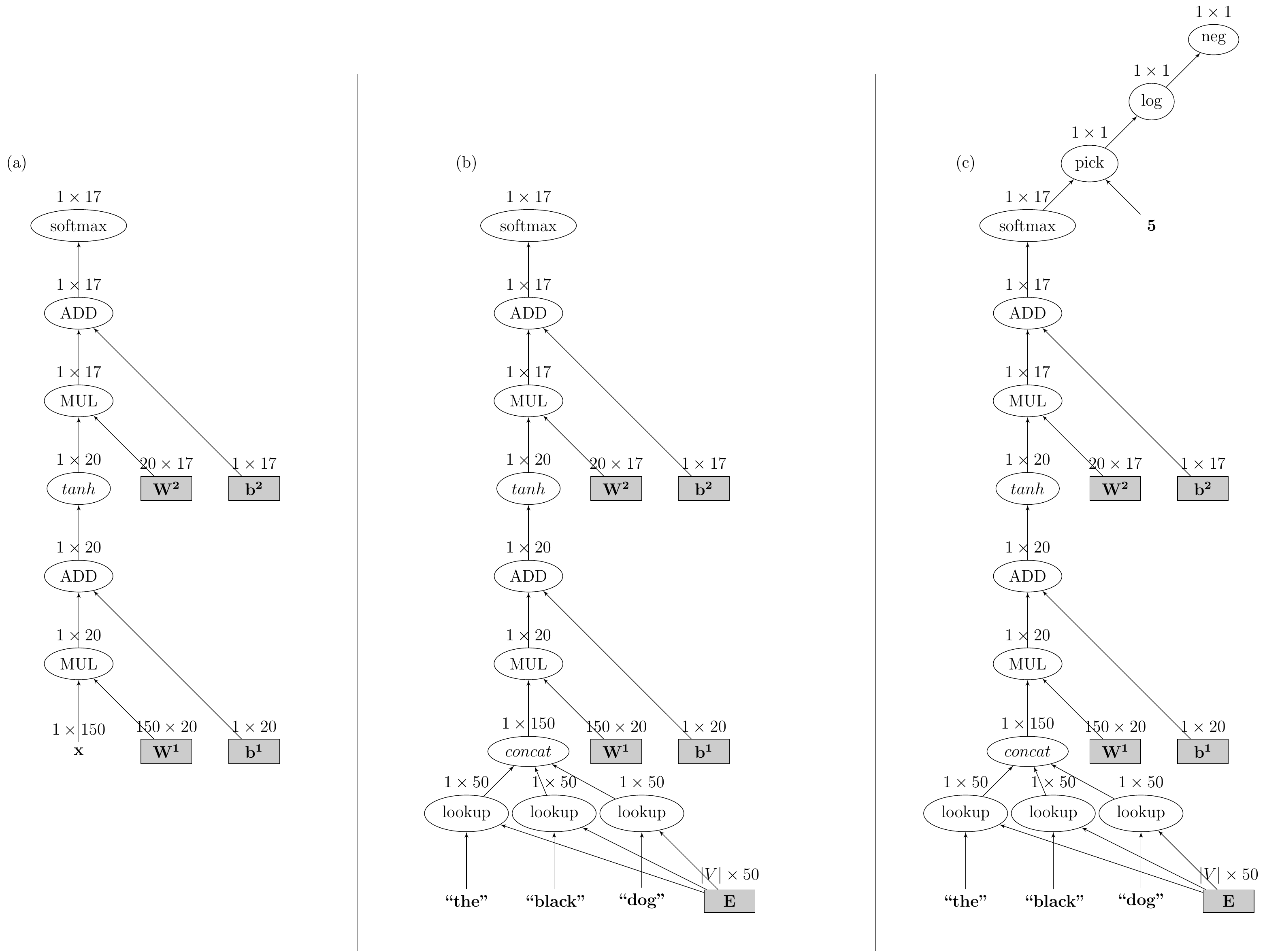}
\caption{\textbf{Computation Graph for MLP1.} (a) Graph with unbound input. (b) Graph with concrete input. (c) Graph with concrete input, expected output, and loss node.}
\label{fig:mlp-cg}
\end{figure}

Since a neural network is essentially a mathematical expression, it can be
represented as a computation graph.

For example, Figure \ref{fig:mlp-cg}a presents the computation graph for a 1-layer MLP with a softmax output transformation.
In our notation, oval nodes represent mathematical operations or functions, and shaded rectangle nodes
represent parameters (bound variables). Network inputs are treated as constants, and drawn without a surrounding node.
Input and parameter nodes have no incoming arcs, and output nodes
have no outgoing arcs.
The output of each node is a matrix, the dimensionality of which is indicated
above the node.

This graph is incomplete: without specifying the inputs, we cannot compute an
output. Figure \ref{fig:mlp-cg}b shows a complete graph for an MLP that takes
three words as inputs, and predicts the distribution over part-of-speech tags
for the third word.  This graph can be used for prediction, but not for
training, as the output is a vector (not a scalar) and the graph does not take
into account the correct answer or the loss term. Finally, the graph in
\ref{fig:mlp-cg}c shows the computation graph for a specific training example,
in which the inputs are the (embeddings of) the words ``the'', ``black'',
``dog'', and the expected output is ``NOUN'' (whose index is 5).

Once the graph is built, it is straightforward to run either a forward
computation (compute the result of the computation) or a backward computation
(computing the gradients), as we show below. Constructing the graphs may look
daunting, but is actually very easy using dedicated software libraries and APIs.

\paragraph{Forward Computation} The forward pass computes the outputs of the
nodes in the graph. Since each node's output depends only on itself and on its
incoming edges, it is trivial to compute the outputs of all nodes by traversing
the nodes in a topological order and computing the output of each node given the
already computed outputs of its predecessors.

More formally, in a graph of $N$ nodes, we associate each node with an index $i$ according to their topological
ordering. Let $f_i$ be the function computed by node $i$ (e.g.
\emph{multiplication}. \emph{addition}, \dots). Let $\pi(i)$ be the
parent nodes of node $i$, and $\pi^{-1}(i) = \{j\mid i\in\pi(j)\}$ the children nodes
of node $i$ (these are the arguments of $f_i$).
Denote by $v(i)$ the output of node $i$, that is, the application of $f_i$ to
the output values of its arguments $\pi^{-1}(i)$.    For variable and input nodes,
$f_i$ is a constant function and $\pi^{-1}(i)$ is empty. 
The Forward algorithm computes the values $v(i)$ for all $i \in [1, N]$.
\begin{algorithm}[h]
    \caption{Computation Graph Forward Pass}
    \label{alg:forward}
\begin{algorithmic}[1]
    \For {i = 1 to N}
    \State Let $a_1,\ldots,a_m = \pi^{-1}(i)$
    \State $v(i) \gets f_i(v(a_1),\ldots,v(a_m))$
    \EndFor
\end{algorithmic}
\end{algorithm}

\paragraph{Backward Computation (Derivatives, Backprop)} 
The backward pass begins by designating a node $N$ with scalar ($1 \times 1$) output as a loss-node, and running forward computation up to that node.
The backward computation will computes the gradients with respect to that node's value. 
Denote by $d(i)$ the quantity $\diffp{N}{i}$.  The backpropagation algorithm is
used to compute the values $d(i)$ for all nodes $i$.

\noindent The backward pass fills a table $d(i)$ as follows:
\begin{algorithm}[h]
    \caption{Computation Graph Backward Pass (Backpropagation)}
    \label{alg:forward}
\begin{algorithmic}[1]
    \State $d(N) \gets 1$
    \For {i = N-1 to 1}
        \State $d(i) \gets \sum_{j \in\pi(i)} d(j) \cdot \diffp{f_j}{i}$
    \EndFor
\end{algorithmic}
\end{algorithm}

\noindent The quantity $\diffp{f_j}{i}$ is the partial derivative of
$f_j(\pi^{-1}(j))$ w.r.t the
argument $i \in \pi^{-1}(j)$.  This value depends on the function $f_j$ and the
values $v(a_1),\ldots,v(a_m)$ (where $a_1,\ldots,a_m = \pi^{-1}(j)$) of its arguments, which were computed in the forward pass.

Thus, in order to define a new kind of node, one need to define two methods: one
for calculating the forward value $v(i)$ based on the nodes inputs, and the
another for calculating $\diffp{f_i}{x}$ for each $x \in \pi^{-1}(i)$.

For further information on automatic differentiation see
\cite[Section 7]{neidinger2010introduction}, \cite{baydin2015automatic}.
\ygcomment{\cite{siam-book@@,others?}}
For more in depth discussion of the backpropagation algorithm and computation
graphs (also called flow graphs) see
\cite[Section 6.4]{bengio2015deep}, \cite{lecun1998gradient,bengio2012practical}.
For a popular yet technical presentation,
see Chris Olah's description at
\url{http://colah.github.io/posts/2015-08-Backprop/}.

\paragraph{Software} 
Several software packages implement the computation-graph model, including
Theano\footnote{\url{http://deeplearning.net/software/theano/}},
Chainer\footnote{\url{http://chainer.org}},
penne\footnote{\url{https://bitbucket.org/ndnlp/penne}} and
CNN/pyCNN\footnote{\url{https://github.com/clab/cnn}}.
All these packages support all the essential
components (node types) for defining a wide range of neural network
architectures, covering the structures described in this tutorial and more.
Graph creation is made almost transparent by use of operator overloading. The
framework defines a type for representing graph nodes (commonly called
\emph{expression}s), methods for constructing nodes for inputs and parameters,
and a set of functions and mathematical operations that take expressions as
input and result in more complex expressions. For example, the python code for
creating the computation graph from Figure (\ref{fig:mlp-cg}c) using the pyCNN
framework is:

\begin{lstlisting}[language=python]
from pycnn import *
# model initialization.
model = Model()
model.add_parameters("W1", (20,150))
model.add_parameters("b1", 20)
model.add_parameters("W2", (17,20))
model.add_parameters("b2", 17)
model.add_lookup_parameters("words", (100, 50))

# Building the computation graph:
renew_cg() # create a new graph.
# Wrap the model parameters as graph-nodes.
W1 = parameter(model["W1"])
b1 = parameter(model["b1"])
W2 = parameter(model["W2"])
b2 = parameter(model["b2"])
def get_index(x): return 1
# Generate the embeddings layer.
vthe   = lookup(model["words"], get_index("the"))
vblack = lookup(model["words"], get_index("black"))
vdog   = lookup(model["words"], get_index("dog"))

# Connect the leaf nodes into a complete graph.
x = concatenate([vthe, vblack, vdog])
output = softmax(W2*(tanh(W1*x)+b1)+b2)
loss = -log(pick(output, 5))

loss_value = loss.forward()
loss.backward() # the gradient is computed 
                 # and stored in the corresponding
                 # parameters.
\end{lstlisting}

\noindent Most of the code involves various initializations: the first block
defines model parameters that are be shared between different computation
graphs (recall that each graph corresponds to a specific training example).
The second block turns the model parameters into the graph-node
(Expression) types.  The third block retrieves the Expressions for the
embeddings of the input words.  Finally, the fourth block is where the graph
is created. Note how transparent the graph creation is -- there is an almost a
one-to-one correspondence between creating the graph and describing it
mathematically.  The last block shows a forward and backward pass.  
The other software frameworks follow similar patterns.

Theano involves an optimizing compiler for computation graphs, which is both a
blessing and a curse. On the one hand, once compiled, large graphs can be run
efficiently on either the CPU or a GPU, making it ideal for large graphs with a
fixed structure, where only the inputs change between instances. However, the
compilation step itself can be costly, and it makes the interface a bit
cumbersome to work with.  In contrast, the other packages focus on building
large and dynamic computation graphs and executing them ``on the fly'' without a
compilation step. While the execution speed may suffer with respect to Theano's
optimized version, these packages are especially convenient when working with
the recurrent and recursive networks described in Sections \ref{sec:rnn}, \ref{sec:recnn} as well
as in structured prediction settings as described in Section \ref{sec:structured}.

\paragraph{Implementation Recipe}
Using the computation graph abstraction, the pseudo-code for a network training
algorithm is given in Algorithm \ref{alg:nn-train}.

\begin{algorithm}[h!]
    \caption{Neural Network Training with Computation Graph Abstraction
    (using minibatches of size 1) }
   \label{alg:nn-train}

\begin{algorithmic}[1]
   \State Define network parameters.
   \For {iteration = 1 to N}
   \For {Training example $\m{x_i}, \m{y_i}$ in dataset}
   \State loss\_node $\gets$ build\_computation\_graph($\m{x_i}$, $\m{y_i}$, parameters)
   \State loss\_node.forward()
   \State gradients $\gets$ loss\_node().backward()
   \State parameters $\gets$ update\_parameters(parameters,  gradients)
   \EndFor
   \EndFor
   \State \Return parameters.
\end{algorithmic}
\end{algorithm}

Here, build\_computation\_graph is a user-defined function that builds the
computation graph for the given input, output and network structure, returning a
single loss node. update\_parameters is an optimizer specific update rule.
The recipe specifies that a new graph is created for each training example. This
accommodates cases in which the network structure varies between training
example, such as recurrent and recursive neural networks, to be discussed in
Sections \ref{sec:rnn} -- \ref{sec:recnn}.  For networks with fixed
structures, such as an MLPs, it may be more efficient to create one base
computation graph and vary only the
inputs and expected outputs between examples.


\paragraph{Network Composition} As long as the network's output is a vector ($1
\times k$ matrix), it is trivial to compose networks by making the output of one
network the input of another, creating arbitrary networks.  The computation
graph abstractions makes this ability explicit: a node in the computation graph
can itself be a computation graph with a designated output node. 
One can then design arbitrarily deep and complex networks, and be able to easily evaluate and
train them thanks to automatic forward and gradient computation. This makes it
easy to define and train networks for structured outputs and multi-objective
training, as we discuss in Section \ref{sec:joint}, as well as complex recurrent and recursive
networks, as discussed in Sections \ref{sec:rnn}--\ref{sec:recnn}.

\subsection{Optimization Issues}
\label{sec:learning-rate}

Once the gradient computation is taken care of, the network is trained
using SGD or another gradient-based optimization
algorithm. The function being optimized is not
convex, and for a long time training of neural networks was considered a
``black art'' which can only be done by selected few.  Indeed, many parameters
affect the optimization process, and care has to be taken to tune these parameters.
While this tutorial is not intended as a comprehensive guide to successfully training
neural networks, we do list here a few of the prominent issues.  For further discussion
on optimization techniques and algorithms for neural networks, refer to
\cite[Chapter 8]{bengio2015deep}.  For some theoretical discussion and analysis,
refer to \cite{glorot2010understanding}. For various practical tips and
recommendations, see \cite{lecun1998efficient,bottou2012stochastic}.

\paragraph{Initialization} 
\label{sec:glorot-init}
The non-convexity of the loss function means the
optimization procedure may get stuck in a local minimum or a saddle point, and
that starting from different initial points (e.g. different random values for
the parameters) may result in different results. Thus, it is advised to run
several restarts of the training starting at different random initializations,
and choosing the best one based on a development set.\footnote{When debugging,
and for reproducibility of results, it is advised to used a fixed random seed.}
The amount of variance in the
results is different for different network formulations and datasets, and cannot
be predicted in advance.\ygcomment{discuss ensembles?}

The magnitude of the random values has an important effect on the
success of training.  An effective scheme due to Glorot and Bengio
\shortcite{glorot2010understanding}, called \emph{xavier initialization} after
Glorot's first name, suggests initializing a weight matrix $\m{W}\in
\mathbb{R}^{d_{in} \times d_{out}}$ as:
\begin{align*}
    \m{W} \sim U\left[ -\frac{\sqrt{6}}{\sqrt{d_{in} + d_{out}}}, +\frac{\sqrt{6}}{\sqrt{d_{in} + d_{out}}}  \right]
\end{align*}

\noindent where $U[a,b]$ is a uniformly sampled random value in the range $[a,b]$. 
This advice works well on many occasions, and is the preferred default
initialization method by many.  

Analysis by He et al \shortcite{he2015delving}
suggests that when using ReLU non-linearities, the weights should be
initialized by sampling from a zero-mean Gaussian distribution whose standard
deviation is $\sqrt\frac{2}{d_{in}}$.
This initialization was found by He et
al to work better than xavier initialization in an image classification task,
especially when deep networks were involved.
\ygcomment{bias and embeddings?}

\paragraph{Vanishing and Exploding Gradients} In deep networks, it is common for
the error gradients to either vanish (become exceedingly close to 0) or explode
(become exceedingly high) as they propagate back through the computation graph.
The problem becomes more severe in deeper networks, and especially so in
recursive and recurrent networks \cite{pascanu2012difficulty}. Dealing with the vanishing gradients problem is
still an open research question. Solutions include making the networks shallower,
step-wise training (first train the first layers based on some auxiliary output signal,
then fix them and train the upper layers of the complete network based on the
real task signal), or specialized architectures that are designed to assist in
gradient flow (e.g., the LSTM and GRU architectures for recurrent networks,
discussed in Section \ref{sec:rnn-arch}).
Dealing with the exploding gradients has a simple but very
effective solution: clipping the gradients if their norm exceeds a given
threshold.  Let $\m{\hat{g}}$ be the gradients of all parameters in the network,
and $\|\m{\hat{g}}\|$ be their $L_2$ norm. Pascanu et al
\shortcite{pascanu2012difficulty} suggest to set: $\m{\hat{g}} \gets
\frac{threshold}{\|\m{\hat{g}}\|}\m{\hat{g}}$ if $\|\m{\hat{g}}\| > threshold$.

\paragraph{Saturation and Dead Neurons}  Layers with $tanh$ and $sigmoid$ activations can
become saturated -- resulting in output values for that layer that are all close to one,
the upper-limit
of the activation function. Saturated neurons have very small gradients, and
should be avoided.
Layers with the ReLU activation cannot be saturated, but can ``die'' -- most or
all values are negative and thus clipped at zero for all inputs, resulting in a
gradient of zero for that layer.  
If your network does not train well, it is advisable to monitor the network for
saturated or dead layers. Saturated neurons are caused by
too large values entering the layer.  This may be controlled for by changing the
initialization, scaling the range of the input values, or changing the learning
rate.
Dead neurons are caused by all weights entering the layer being
negative (for example this can happen after a large gradient update). Reducing
the learning rate will help in this situation.
For saturated layers, another option is to normalize the values in the saturated
layer after the activation, i.e. instead of $g(\m{h})=tanh(\m{h})$ using
$g(\m{h})=\frac{tanh(\m{h})}{\|tanh(\m{h})\|}$.
Layer normalization is an effective measure for countering saturation, but is also expensive in
terms of gradient computation. 

\paragraph{Shuffling} The order in which the training examples are presented to
the network is important. The SGD formulation above specifies selecting a random
example in each turn. In practice, most implementations go over the training example
in order.
It is advised to shuffle the training examples before
each pass through the data.

\paragraph{Learning Rate} Selection of the learning rate is important. Too large
learning rates will prevent the network from converging on an effective
solution.  Too small learning rates will take very long time to converge.  As a
rule of thumb, one should experiment with a range of initial learning rates in
range $[0, 1]$, e.g. $0.001$, $0.01$, $0.1$, $1$.  Monitor the network's loss
over time, and decrease the learning rate once the network seem to be stuck in a
fixed region.  \emph{Learning rate scheduling} decrease the rate as a function
of the number of observed minibatches.
A common schedule is dividing the initial learning rate by the iteration
number.  L\'eon Bottou \shortcite{bottou2012stochastic} recommends using a
learning rate of the form $\eta_t = \eta_0(1+\eta_0\lambda t)^{-1}$ where
$\eta_0$ is the initial learning rate, $\eta_t$ is the learning rate to use on
the $t$th training example, and $\lambda$ is an additional hyperparameter.  He
further recommends determining a good value of $\eta_0$ based on a small sample of
the data prior to running on the entire dataset.

\paragraph{Minibatches} Parameter updates occur either every training
example (minibatches of size 1) or every $k$ training examples. 
Some problems benefit from training with larger minibatch sizes.
In terms of the computation graph abstraction, one can create
a computation graph for each of the $k$ training examples, and then connecting
the $k$ loss nodes under an averaging node, whose output will be the loss
of the minibatch. 
Large minibatched training can also be beneficial in terms of computation
efficiency on specialized computing architectures such as GPUs.  This is beyond
the scope of this tutorial.

\subsection{Regularization}
\label{sec:dropout}

Neural network models have many parameters, and overfitting can easily occur.
Overfitting can be alleviated to some extent by \emph{regularization}.
A common regularization method is $L_2$ regularization, placing a squared
penalty on parameters with large values by adding an additive $\frac{\lambda}{2}\|\theta\|^2$ term
to the objective function to be minimized, where $\theta$ is the set of model
parameters, $\|\cdot\|^2$ is the squared $L_2$ norm (sum of squares of the
values), and $\lambda$ is a hyperparameter controlling the amount of
regularization.

A recently proposed alternative regularization method is \emph{dropout}
\cite{hinton2012improving}.  The dropout method is designed to prevent the
network from learning to rely on specific weights.  It works by randomly
dropping (setting to 0) half of the neurons in the network (or in a specific
layer) in each training example.  
Work by Wager et al
\shortcite{wager2013dropout} establishes a strong connection between the dropout
method and $L_2$ regularization.  Gal and Gharamani \shortcite{gal2015dropout}
show that a multi-layer perceptron with dropout applied at every layer can be
interpreted as Bayesian model averaging.

The dropout technique is one of the key factors
contributing to very strong results of neural-network methods on image
classification tasks \cite{krizhevsky2012imagenet}, especially when combined
with ReLU activation units \cite{dahl2013improving}.  The dropout technique
is effective also in NLP applications of neural networks.  

\clearpage
\section{Cascading and Multi-task Learning}
\label{sec:joint}

The combination of online training methods with automatic gradient computations
using the computation graph abstraction allows for an easy implementation of model
cascading, parameter sharing and multi-task learning.

\paragraph{Model cascading} is a powerful technique in which large networks are built by
composing them out of smaller component networks.
For example, we may have a feed-forward network for predicting the part of
speech of a word based on its neighbouring words and/or the characters that
compose it.  In a pipeline approach, we would use this network for predicting
parts of speech, and then feed the predictions as input features to neural
network that does syntactic chunking or parsing.
Instead, we could think of the hidden layers of this network as an encoding
that captures the relevant information for predicting the part of speech.  In a
cascading approach, we take the hidden layers of this network and connect them (and
not the part of speech prediction themselves) as the inputs for the syntactic
network.  We now have a larger network that takes as input sequences of words and
characters, and outputs a syntactic structure.
The computation graph abstraction allows us to easily propagate the
error gradients from the syntactic task loss all the way back to the characters.

To combat the vanishing gradient problem of deep networks, as well as to make
better use of available training material, the individual component network's
parameters can be bootstrapped by training
them separately on a relevant task, before plugging them in to the larger network
for further tuning.  For example, the part-of-speech predicting network can be
trained to accurately predict parts-of-speech on a relatively large annotated
corpus, before plugging its hidden layer into the syntactic parsing network
for which less training data is available.  In case the training data provide
direct supervision for both tasks, we can make use of it during training by
creating a network with two outputs, one for each task, computing a separate
loss for each output, and then summing the losses into a single node from which
we backpropagate the error gradients.

Model cascading is very common when using convolutional, recursive and recurrent neural
networks, where, for example, a recurrent network is used to encode a sentence
into a fixed sized vector, which is then used as the input of another network.
The supervision signal of the recurrent network comes primarily from the upper
network that consumes the recurrent network's output as it inputs.

\paragraph{Multi-task learning} is used when we have related prediction tasks that do not
necessarily feed into one another, but we do believe that information that is
useful for one type of prediction can be useful also to some of the other tasks.
For example, chunking, named entity recognition (NER) and language modeling are
examples of synergistic tasks. Information for predicting chunk boundaries,
named-entity boundaries and the next word in the sentence all rely on some
shared underlying syntactic-semantic representation.  Instead of training a
separate network for each task, we can create a single network with several
outputs.  A common approach is to have a multi-layer feed-forward network, whose
final hidden layer (or a concatenation of all hidden layers) is then passed to
different output layers.
\ygcomment{figure!}
This way, most of the parameters of the network are
shared between the different tasks.  Useful information learned from one
task can then help to disambiguate other tasks.  
Again, the computation graph abstraction makes it very easy to construct such
networks and compute the gradients for them, by computing a separate loss for
each available supervision signal, and then summing the losses into a single
loss that is used for computing the gradients.  In case we have several corpora,
each with different kind of supervision signal (e.g. we have one corpus for NER
and another for chunking), the training procedure will shuffle all of the
available training example, performing gradient computation and updates with
respect to a different loss in every turn.  Multi-task learning in the context
of language-processing is introduced and discussed in \cite{collobert2011natural}.

\clearpage
\section{Structured Output Prediction}
\label{sec:structured}

Many problems in NLP involve structured outputs: cases where the desired output
is not a class label or distribution over class labels, but a structured object
such as a sequence, a tree or a graph. Canonical examples are sequence tagging
(e.g. part-of-speech tagging) sequence segmentation (chunking, NER), and syntactic parsing.
In this section, we discuss how feed-forward neural network models can be used for structured tasks.
In later sections we discuss specialized neural network models for dealing with
sequences (Section \ref{sec:rnn}) and trees (Section \ref{sec:recnn}).

\subsection{Greedy Structured Prediction}

The greedy approach to structured prediction is to decompose the structure
prediction problem into a sequence of local prediction problems and training
a classifier to perform each local decision.  At test time, the trained
classifier is used in a greedy manner. Examples of this approach are
left-to-right tagging models \cite{gimenez2004svmtool} and greedy transition-based parsing \cite{nivre2008algorithms}.
Such approaches are easily adapted to use neural networks
by simply replacing the local classifier from a linear classifier such as an SVM
or a logistic regression model to a neural network, as demonstrated in \cite{chen2014fast,lewis2014improved}.

The greedy approaches suffer from error propagation, where mistakes in early
decisions carry over and influence later decisions.  The overall higher accuracy
achievable with non-linear neural network classifiers helps in offsetting this
problem to some extent.  In addition,
training techniques were proposed for mitigating the error
propagation problem by either attempting to take easier predictions before
harder ones (the easy-first approach \cite{goldberg2010efficient}) or making
training conditions more similar to testing conditions by exposing the
training procedure to inputs that result from likely mistakes
\cite{daume09searn,goldberg2013training}.  These are effective also for
training greedy neural network models, as demonstrated by Ma et al
\cite{ma2014tagging} (easy-first tagger) and \cite{in-submission} (dynamic
oracle training for greedy dependency parsing).

\subsection{Search Based Structured Prediction}

The common approach to predicting natural language structures is search based.
For in-depth discussion of search-based structure prediction in NLP, see the book by Smith
\cite{smith2011linguistic}.
The techniques can easily be
adapted to use a neural-network.  In the neural-networks literature, such models
were discussed under the framework of \emph{energy based learning}
\cite[Section 7]{lecun2006tutorial}. They are presented here using setup and
terminology familiar to the NLP community.

Search-based structured prediction is formulated as a search problem over
possible structures:

\[ 
predict(x) = \argmax_{y \in \mathcal{Y}(x)} score(x,y)
\]

\noindent where $x$ is an input structure, $y$ is an output over $x$ (in a
typical example $x$ is a sentence and $y$ is a tag-assignment or a parse-tree
over the sentence), $\mathcal{Y}(x)$ is the set of all valid structures over
$x$, and we are looking for an output $y$ that will maximize the score of the $x,y$ pair.
 
The scoring function is defined as a linear model:
\[
score(x,y) = \Phi(x,y)\cdot\m{w}
\]
\noindent where $\Phi$ is a feature extraction function and $\m{w}$ is a weight vector.

In order to make the search for the optimal $y$ tractable, the structure $y$ is
decomposed into parts, and the feature function is defined in terms of the
parts, where $\phi(p)$ is a part-local feature extraction function:
\[
    \Phi(x,y) = \sum_{\mathclap{p \in parts(x,y)}} \phi(p)
\]

Each part is scored separately, and the structure score
is the sum of the component parts scores:

\begin{align*}
score(x,y) =& \m{w}\cdot\Phi(x,y) = \m{w}\cdot\sum_{p \in y} \phi(p) = \sum_{p\in y} \m{w}\cdot\phi(p) = \sum_{p \in y} score(p)
\end{align*}

\noindent where $p \in y$ is a shorthand for $p \in parts(x,y)$.
The decomposition of $y$ into parts is such that there exists an inference
algorithm that allows for efficient search for the best scoring structure given
the scores of the individual parts.

One can now trivially replace the linear scoring function over parts with a neural-network:

\[
score(x,y) = \sum_{p\in y} score(p) = \sum_{p \in y} NN(c(p))
\]

\noindent where $c(p)$ maps the part $p$ into a $d_{in}$ dimensional vector. 

In case of a one hidden-layer feed-forward network:

\[
score(x,y) = \sum_{p \in y} NN_{MLP1}(c(p)) = \sum_{p \in y} (g(c(p)\m{W^1} + \m{b^1}))\m{w}
\]

\noindent
$c(p) \in \mathbb{R}^{d_{in}}$,
$\m{W^1} \in \mathbb{R}^{d_{in} \times d_{1}}$,
$\m{b^1} \in \mathbb{R}^{d_1}$,
$\m{w} \in \mathbb{R}^{d_1}$.
A common objective in structured prediction is making the gold structure $y$
score higher than any other structure $y'$, leading to the following
(generalized perceptron) loss:

\[
\max_{y'} score(x,y') - score(x,y)
\]

In terms of implementation, this means: create a computation graph $CG_p$ for
each of the possible parts, and calculate its score. Then, run inference over
the scored parts to find the best scoring structure $y'$. Connect the output
nodes of the computation graphs corresponding to parts in the gold (predicted)
structure $y$ ($y'$) into a summing node $CG_y$ ($CG_y'$). Connect $CG_y$ and
$CG_y'$ using a ``minus'' node, $CG_l$, and compute the gradients. 

As argued in \cite[Section 5]{lecun2006tutorial}, the generalized perceptron loss may not be
a good loss function when training structured prediction neural networks as it
does not have a margin, and a margin-based hinge loss is preferred:

\[
max(0, m + score(x,y) - \max_{y' \neq y} score(x,y'))
\]

\noindent It is trivial to modify the implementation above to work with the hinge loss.

Note that in both cases we lose the nice properties of the linear model. In particular,
the model is no longer convex. This is to be expected, as even the simplest non-linear neural
network is already non-convex. Nonetheless, we could still use standard
neural-network optimization techniques to train the structured model.

Training and inference is slower, as we have to evaluate the neural network (and
take gradients) $|parts(x,y)|$ times.

Structured prediction is a vast field and is beyond the scope of this tutorial,
but loss functions, regularizers and methods described in, e.g.,
\cite{smith2011linguistic}, such as cost-augmented decoding, can be easily
applied or adapted to the neural-network framework.\footnote{One should keep in
mind that the resulting objectives are no longer convex, and so lack the formal
guarantees and bounds associated with convex optimization problems. Similarly,
the theory, learning bounds and guarantees associated with the algorithms do not
automatically transfer to the neural versions.}

\paragraph{Probabilistic objective (CRF)}

In a probabilistic framework (``CRF''), we treat each of the parts scores as a
\emph{clique potential} (see \cite{smith2011linguistic}) and define the score of each structure $y$ to be:
\[
score_{CRF}(x,y) = P(y|x) = \frac{\sum_{p \in y} e^{score(p)}}{\sum_{y' \in \mathcal{Y}(x)}\sum_{p \in y'}e^{score(p)}} = \frac{\sum_{p \in y} e^{NN(c(p))}}{\sum_{y' \in \mathcal{Y}(x)}\sum_{p \in y'}e^{NN(c(p))}} 
\]

The scoring function defines a conditional distribution $P(y|x)$, and we wish to
set the parameters of the network such that corpus conditional log likelihood
$\sum_{(x_i,y_i) \in training}\log P(y_i|x_i)$ is maximized.

The loss for a given training example $(x,y)$ is then: $-\log score_{CRF}(x,y)$.
Taking the gradient with respect to the loss is as involved as building the
associated computation graph. The tricky part is the denominator (the
\emph{partition function}) which requires summing over the potentially
exponentially many structures in $\mathcal{Y}$. However, for some problems, a
dynamic programming algorithm exists for efficiently solving the summation in
polynomial time. When such an algorithm exists, it can be adapted to also create
a polynomial-size computation graph.

When an efficient enough algorithm for computing the partition function is not
available, approximate methods can be used. For example, one may use beam search
for inference, and for the partition function sum over the structures remaining
in the beam instead of over the exponentially large $\mathcal{Y}(x)$.  

A hinge based approached was used by Pei et al \shortcite{pei2015effective} for 
arc-factored dependency parsing, and the probabilistic approach by Durrett and
Klein \cite{durrett2015neural} for a CRF constituency parser.
The approximate beam-based partition function was effectively used by 
Zhou et al \shortcite{zhou2015neural} in a transition based parser.

\paragraph{Reranking} When searching over all possible structures is
intractable, inefficient or hard to integrate into a model, reranking methods
are often used. In the reranking framework \cite{charniak2005coarsetofine,collins2005discriminative}
a base model is used to produce a list of the $k$-best scoring structures.
A more complex model is then trained to score the candidates in the $k$-best
list such that the best structure with respect to the gold one is scored
highest.  As the search is now performed over $k$ items rather than over an
exponential space, the complex model can condition on
(extract features from) arbitrary aspects of the scored structure.
Reranking methods are natural candidates for structured prediction using
neural-network models, as they allow the modeler to focus on the feature
extraction and network structure, while removing the need to integrate the
neural network scoring into a decoder.  Indeed, reranking methods are often used
for experimenting with neural models that are not straightforward to integrate
into a decoder, such as convolutional, recurrent and recursive networks, which
will be discussed in later sections. Works using the reranking approach include 
\cite{socher2013parsing,auli2013joint,le2014insideoutside,zhu2015reranking}

\paragraph{MEMM and hybrid approaches}\ygcomment{elaborate on MEMM?}
Other formulations are, of course, also possible. For example, an MEMM
\cite{mccallum2000maximum} can be
trivially adapted to the neural network world by replacing the logistic
regression (``Maximum Entropy'') component with an MLP.

Hybrid approaches between neural networks and linear models are also explored.
In particular, Weiss et al \cite{weiss2015structured} report strong results for
transition-based dependency parsing in a two-stage model.  In the first stage, 
a static feed-forward neural
network (MLP2) is trained to perform well on 
each of the individual decisions of the structured problem in isolation.
In the second stage, the neural network model is held fixed, and the
different layers (output as well as hidden layer vectors) for each input are
then concatenated and used as the input features of a linear structured
perceptron model \cite{collins2002discriminative} that is trained to perform
beam-search for the best resulting
structure. While it is not clear that such training regime is more effective than
training a single structured-prediction neural network, the use of two simpler,
isolated models allowed the researchers to perform a much more extensive hyper-parameter
search (e.g. tuning layer sizes, activation functions, learning rates and so on)
for each model than is feasible with more complicated networks.

\clearpage
\section{Convolutional Layers}
\label{sec:convnet}


Sometimes we are interested in making predictions based on ordered sets of items
(e.g. the sequence of words in a sentence, the sequence of sentences in a
document and so on).  Consider for example predicting the sentiment (positive,
negative or neutral) of a sentence.  Some of the sentence words are very
informative of the sentiment, other words are less informative, and to a good
approximation, an informative clue is informative regardless of its position in
the sentence.  We would like to feed all of the sentence words into a learner,
and let the training process figure out the important clues.  One possible
solution is feeding a CBOW representation into a fully connected
network such as an MLP.  However, a downside of the CBOW approach is that it
ignores the ordering information completely, assigning the sentences ``it was
not good, it was actually quite bad'' and ``it was not bad, it was actually
quite good'' the exact same representation.  While the global position of the
indicators ``not good'' and ``not bad'' does not matter for the classification
task, the local ordering of the words (that the word ``not'' appears right
before the word ``bad'') is very important.  A naive approach would suggest
embedding word-pairs (bi-grams) rather than words, and building a CBOW over the
embedded bigrams.  While such architecture could be effective, it will result in
huge embedding matrices, will not scale for longer n-grams, and will suffer from
data sparsity problems as it does not share statistical strength between
different n-grams (the embedding of ``quite good'' and ``very good'' are
completely independent of one another, so if the learner saw only one of them
during training, it will not be able to deduce anything about the other based on
its component words).  The convolution-and-pooling (also called convolutional
neural networks, or CNNs) architecture is an elegant and robust solution to the
this modeling problem.  A convolutional neural network is designed to identify
indicative local predictors in a large structure, and combine them to produce a
fixed size vector representation of the structure, capturing these local aspects
that are most informative for the prediction task at hand.

Convolution-and-pooling architectures \cite{lecun1995convolutional} evolved in the neural
networks vision community, where
they showed great success as object detectors -- recognizing an object from a
predefined category (``cat'', ``bicycles'') regardless of its position in the
image \cite{krizhevsky2012imagenet}.  When applied to images, the architecture is using
2-dimensional (grid) convolutions.  When applied to text, NLP we are mainly
concerned with 1-d (sequence) convolutions.  Convolutional networks were introduced
to the NLP community in the pioneering work of Collobert, Weston and Colleagues \shortcite{collobert2011natural}
who used them for semantic-role labeling, and later by Kalchbrenner et al \shortcite{kalchbrenner2014convolutional}
and Kim \cite{kim2014convolutional} who used them for sentiment and question-type classification.

\subsection{Basic Convolution + Pooling}
The main idea behind a convolution and pooling architecture for language tasks
is to apply a non-linear (learned) function over each instantiation of a $k$-word sliding
window over the
sentence. This
function (also called ``filter'') transforms a window of $k$ words into a
$d$ dimensional vector that
captures important properties of the words in the window (each dimension is
sometimes referred to in the literature as a ``channel''). Then, a ``pooling''
operation is used combine the vectors resulting from the different windows into
a single $d$-dimensional vector, by taking the max or the average value observed in each of the
$d$ channels over the different windows.  The intention is to focus on the
most important ``features'' in the sentence, regardless of their location.
The $d$-dimensional vector is then fed further into a network that is used
for prediction. The gradients that are propagated back from the network's loss
during the training process are used to tune the parameters of the filter
function to highlight the aspects of the data that are important for the task
the network is trained for.  Intuitively, when the sliding window is run over a sequence,
the filter function learns to identify informative k-grams.

More formally, consider a sequence of words $\m{x} = x_1,\ldots,x_n$, each with their
corresponding $d_{emb}$ dimensional word embedding $v(x_i)$.  A $1$d convolution
layer\footnote{$1$d here refers to a convolution operating over 1-dimensional
inputs such as sequences, as opposed to 2d convolutions which are
applied to images.} of width $k$ works by moving a sliding window of size $k$ over the
sentence, and applying the same ``filter'' to each window in the sequence
$(v(x_i);v(x_{i+1}),\ldots;v(x_{i+k-1})$. 
The filter function is usually a linear transformation followed by a non-linear activation function.

Let the concatenated vector of the $i$th window be $\m{w_i} =
v(x_i);v(x_{i+1});v(x_{i+k-1})$, $\m{w_i} \in \mathbb{R}^{kd_{emb}}$. 
Depending on whether we pad the sentence with $k-1$ words to each side, we may
get either $m=n-k+1$ (\emph{narrow convolution}) or $m=n+k+1$ windows (\emph{wide
convolution}) \cite{kalchbrenner2014convolutional}.  
The result of the convolution layer is $m$ vectors $\m{p_1},\ldots,\m{p_m}$, $\m{p_i}\in \mathbb{R}^{d_{conv}}$ where:
\begin{align*}
\m{p_i}  = g(\m{w_i}\m{W}+\m{b})
\end{align*}.
\ygcomment{TODO: dimensions, terminology:channels;maps }

$g$ is a non-linear activation function that is applied element-wise, $\m{W} \in
\mathbb{R}^{k\cdot d_{emb} \times d_{conv}}$ and $\m{b} \in
\mathbb{R}^{d_{conv}}$ are parameters of the network.  Each $\m{p_i}$ is a
$d_{conv}$ dimensional vector, encoding the information in $\m{w_i}$. Ideally,
each dimension captures a different kind of indicative information.  The $m$
vectors are then combined using a \emph{max pooling layer}, resulting in a
single $d_{conv}$ dimensional vector $\m{c}$.

\[
c_j = \max_{1 < i \leq m} \m{p_i}[j]
\]  

\noindent $\m{p_i}[j]$ denotes the $j$th component of $\m{p_i}$.
The effect of the max-pooling operation is to get the most salient information across
window positions. Ideally, each dimension will ``specialize'' in a particular
sort of predictors, and max operation will pick on the most important
predictor of each type.

Figure \ref{fig:cnn} provides an illustration of the process.
\begin{figure}[t]
\begin{center}
\includegraphics[width=0.9\textwidth]{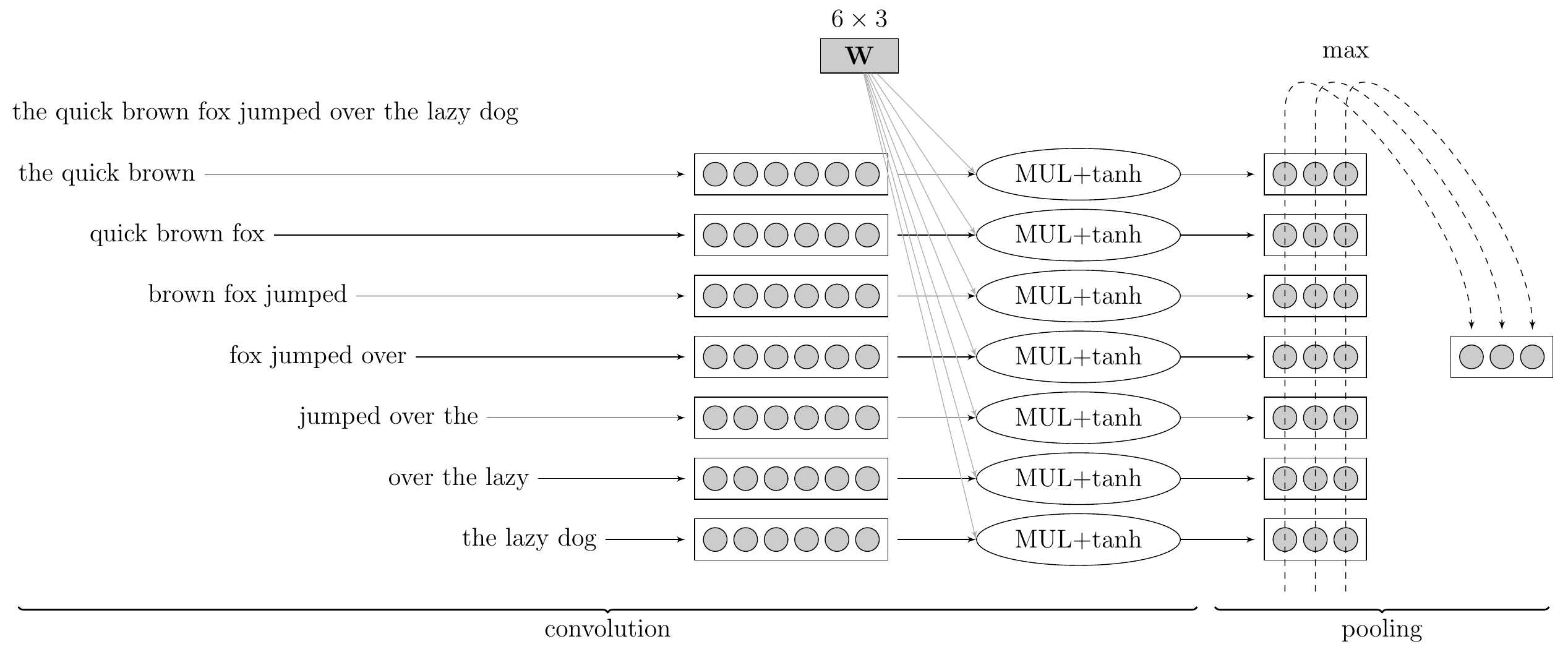}
\end{center}
\caption{$1$d convolution+pooling over the sentence ``the quick brown fox
jumped over the lazy dog''. This is a narrow convolution (no padding is added to
the sentence) with a window size of 3.  Each word is translated to a $2$-dim
embedding vector (not shown). The embedding vectors are then concatenated,
resulting in $6$-dim window representations.
Each of the seven windows is transfered through a $6\times 3$ filter (linear transformation followed by element-wise tanh),
resulting in seven $3$-dimensional filtered representations. Then, a max-pooling
operation is applied, taking the max over each dimension, resulting in a final
$3$-dimensional pooled vector.}
\label{fig:cnn}
\end{figure}

The resulting vector $\m{c}$ is a representation of the sentence in which each dimension
reflects the most salient information with respect to some prediction task.
$\m{c}$ is then fed into a downstream network layers, perhaps in parallel to
other vectors, culminating in an output layer which is used for prediction.  The
training procedure of the network calculates the loss with respect to the
prediction task, and the error gradients are propagated all the way back through
the pooling and convolution layers, as well as the embedding layers.
\footnote{Besides being useful for prediction, a by-product of the training
procedure is a set of parameters $\m{W}$, $\m{B}$ and embeddings $v()$ that can
be used in a convolution and pooling architecture to encode arbitrary length
sentences into fixed-size vectors, such that sentences that share the same kind
of predictive information will be close to each other.}

While max-pooling is the most common pooling operation in text applications,
other pooling operations are also possible, the second most common operation
being \emph{average pooling}, taking the average value of each index instead of
the max.

\subsection{Dynamic, Hierarchical and k-max Pooling} 

Rather than performing a single pooling operation over the entire sequence, we
may want to retain some positional information based on our domain understanding
of the prediction problem at hand.  To this end, we can split the vectors
$\m{p_i}$ into $\ell$ distinct groups, apply the pooling separately on each group,
and then concatenate the $\ell$ resulting $d_{conv}$ vectors $
\m{c_1},\ldots,\m{c_\ell}$.  The division of the $\m{p_i}$s into groups is performed
based on domain knowledge. For example, we may conjecture that words appearing
early in the sentence are more indicative than words appearing late. We
can then split the sequence into $\ell$ equally sized regions, applying a separate
max-pooling to each region.  For example, Johnson and Zhang
\cite{johnson2014effective} found
that when classifying documents into topics, it is useful to have 20
average-pooling regions, clearly separating the initial sentences (where
the topic is usually introduced) from later ones, while for a sentiment
classification task a single max-pooling operation over the entire sentence was
optimal (suggesting that one or two very strong signals are enough to determine
the sentiment, regardless of the position in the sentence).

Similarly, in a relation extraction kind of task we may be given two words and
asked to determine the relation between them. We could argue that the words
before the first word, the words after the second word, and the words between
them provide three different kinds of information \cite{chen2015event}.  We can thus
split the $\m{p_i}$ vectors accordingly, pooling separately the windows
resulting from each group.

Another variation is performing \emph{hierarchical pooling}, in which we have a
succession of convolution and pooling layers, where each stage applies a
convolution to a sequence, pools every $k$ neighboring vectors, performs a
convolution on the resulting pooled sequence, applies another convolution and so
on. This architecture allows sensitivity to increasingly larger structures. 
\ygcomment{Figure?}

Finally, \cite{kalchbrenner2014convolutional} introduced a \emph{k-max pooling}
operation, in which the top $k$ values in each dimension are retained instead of
only the best one, while preserving the order in which they appeared in the
text.
For example a, consider the following matrix:

\[
\begin{bmatrix}
1 & 2 & 3 \\
9 & 6 & 5 \\
2 & 3 & 1 \\
7 & 8 & 1 \\
3 & 4 & 1 \\
\end{bmatrix}
\]
\noindent A 1-max pooling over the column vectors will result in
$
\begin{bmatrix}9 & 8 & 5\\
\end{bmatrix}
$, while a 2-max pooling will result in
    the following matrix:
$
\begin{bmatrix}
9 & 6 & 3 \\
7 & 8 & 5 \\
\end{bmatrix}
$ whose rows will then be concatenated to $
\begin{bmatrix}
    9 & 6 & 3 & 7 & 8 & 5\\
\end{bmatrix}
$

The k-max pooling operation makes it possible
to pool the $k$ most active indicators that may be
a number of positions apart; it preserves the order
of the features, but is insensitive to their specific
positions. It can also discern more finely the number
of times the feature is highly activated \cite{kalchbrenner2014convolutional}.

\subsection{Variations}

Rather than a single convolutional layer, several convolutional layers may be
applied in parallel. For example, we may have four different convolutional
layers, each with a different window size in the range 2--5, capturing n-gram
sequences of varying lengths.  
The result of each convolutional layer will then
be pooled, and the resulting vectors concatenated and fed to further processing
\cite{kim2014convolutional}.

The convolutional architecture need not be restricted into the linear ordering
of a sentence.  For example, Ma et al \shortcite{ma2015dependencybased}  generalize the
convolution operation to work over syntactic dependency trees.  There, each
window is around a node in the syntactic tree, and the pooling is performed over the different nodes.
Similarly, Liu et al \shortcite{liu2015dependencybased} apply a convolutional
architecture on top of dependency paths extracted from dependency trees.
Le and Zuidema \shortcite{le2015forest} propose to perform max pooling 
over vectors representing the different derivations leading to the same chart
item in a chart parser.\ygcomment{elaborate on le and zuidema?}

\clearpage
\section{Recurrent Neural Networks -- Modeling Sequences and Stacks}
\label{sec:rnn}

When dealing with language data, it is very common to work with sequences, such
as words (sequences of letters), sentences (sequences of words) and documents.
We saw how feed-forward networks can accommodate arbitrary feature functions over
sequences through the use of vector concatenation and vector addition (CBOW).
In particular, the CBOW representations allows to encode arbitrary length
sequences as fixed sized vectors.
However, the
CBOW representation is quite limited, and forces one to disregard the order of
features.  The convolutional networks also allow encoding a sequence into a
fixed size vector. While representations derived from convolutional networks
are an improvement above the CBOW representation as they offer some sensitivity
to word order, their order sensitivity is restricted to mostly local patterns,
and disregards the order of patterns that are far apart in the sequence.

Recurrent neural networks (RNNs) \cite{elman1990finding} allow representing
arbitrarily sized structured
inputs in a fixed-size vector, while paying attention to the structured
properties of the input. 

\subsection{The RNN Abstraction}
We use $\m{x_{i:j}}$ to denote the sequence of vectors $\m{x_i},\ldots,\m{x_j}$.
The RNN abstraction takes as input an ordered list of input vectors
$\m{x_1},...,\m{x_n}$ together with an initial \emph{state vector} $\m{s_0}$,
and returns an ordered list of state vectors $\m{s_1},...,\m{s_n}$, as well as
an ordered list of \emph{output vectors} $\m{y_1},...,\m{y_n}$. An output vector
$\m{y_i}$ is a function of the corresponding state vector $\m{s_i}$. 
The input vectors $\m{x_i}$ are presented to the RNN in a sequential fashion, and
the state vector $\m{s_i}$ and output vector $\m{y_i}$ represent the state of the RNN after observing the
inputs $\m{x_{1:i}}$.  The output vector $\m{y_i}$ is then used for further
prediction. For example, a model for predicting the conditional
probability of an event $e$ given the sequence $\m{m_{1:i}}$ can be defined as
$p(e=j|\m{x_{1:i}}) = softmax(\m{y_i}\m{W} + \m{b})[j]$.  The RNN model provides
a framework for conditioning on the entire history $\m{x_1},\ldots,\m{x_i}$
without resorting to the Markov assumption which is traditionally used for modeling
sequences.  Indeed, RNN-based language models result in very good perplexity
scores when compared to n-gram based models.

Mathematically, we have a recursively defined function $R$ that takes as input a
state vector $\m{s_i}$ and an input vector $\m{x_{i+1}}$, and results in a new
state vector $\m{s_{i+1}}$. An additional function $O$ is used to map a state
vector $\m{s_i}$ to an output vector $\m{y_i}$.
When constructing an RNN, much like when constructing a feed-forward network,
one has to specify the dimension of the inputs $\m{x_i}$ as well as the
dimensions of the outputs $\m{y_i}$.  The dimensions of the states $\m{s_i}$ are
a function of the output dimension.\footnote{While RNN architectures in which the state dimension
is independent of the output dimension are possible, the current popular
architectures, including the Simple RNN, the LSTM and the GRU do not follow this
flexibility.}

\begin{align*}
    RNN(\m{s_0},\m{x_{1:n}}) =& \m{s_{1:n}}, \; \m{y_{1:n}} \\
\m{s_i} &= R(\m{s_{i-1}}, \m{x_i}) \\
\m{y_i} &= O(\m{s_i}) \\
\end{align*}
\[
\m{x_i} \in \mathbb{R}^{d_{in}}, \;\;
\m{y_i} \in \mathbb{R}^{d_{out}}, \;\;
\m{s_i} \in \mathbb{R}^{f(d_{out})}
\]

The functions $R$ and $O$ are the same across the sequence positions, but the
RNN keeps track of the states of computation through the state vector that is
kept and being passed between invocations of $R$.

Graphically, the RNN has been traditionally presented as in Figure
\ref{fig:rnn-rec}.

\begin{figure}[h!]
\begin{center}
\includegraphics[width=0.3\textwidth]{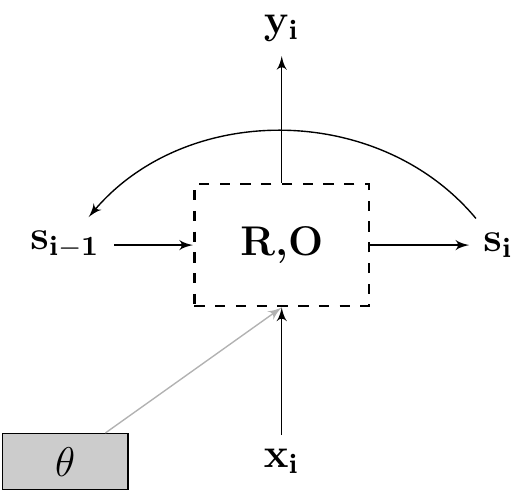}
\end{center}
\caption{Graphical representation of an RNN (recursive).}
\label{fig:rnn-rec}
\end{figure}

\noindent This presentation follows the recursive definition, and is correct for
arbitrary long sequences. However, for a finite sized input sequence (and all
input sequences we deal with are finite) one can \emph{unroll} the recursion,
resulting in the structure in Figure \ref{fig:unrolled-rnn}.

\begin{figure}[h!]
\begin{center}
\includegraphics[width=0.8\textwidth]{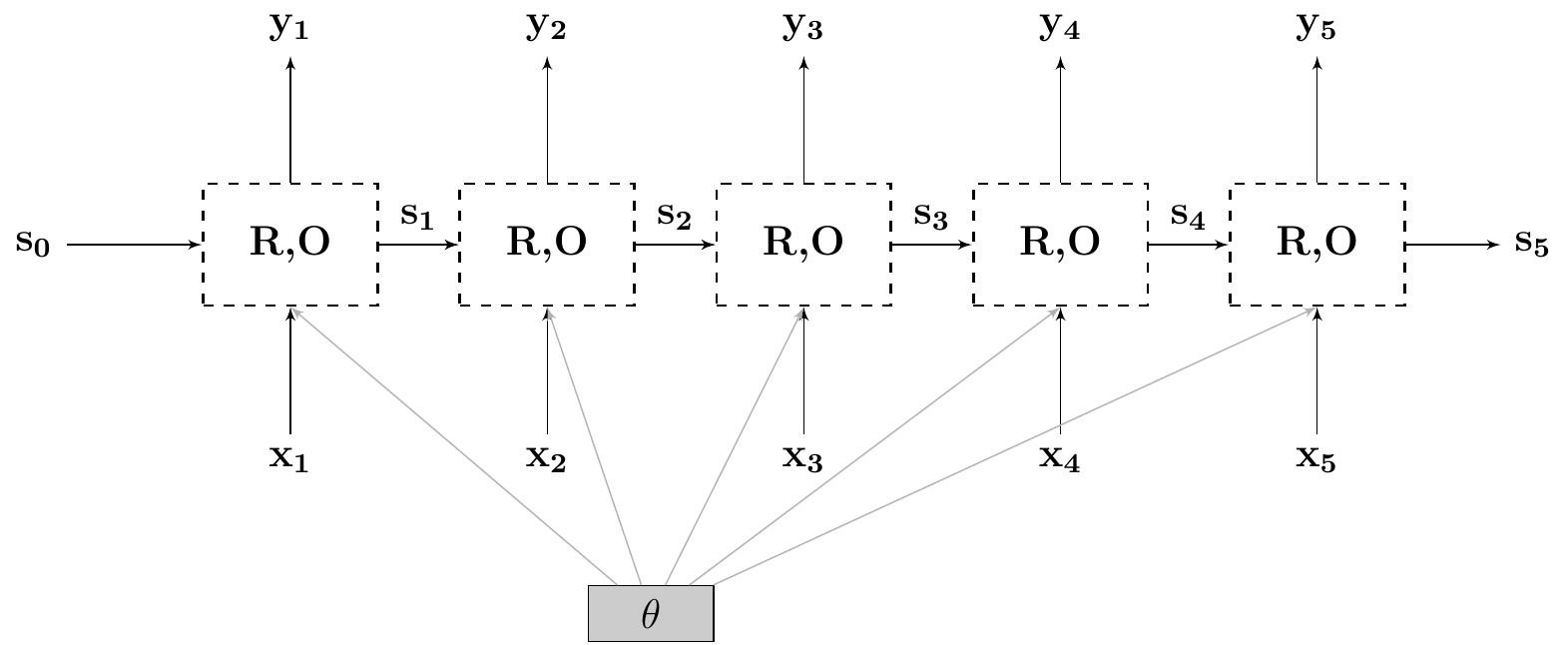}
\caption{Graphical representation of an RNN (unrolled).}
\label{fig:unrolled-rnn}
\end{center}
\end{figure}

\noindent While not usually shown in the visualization, we include here the parameters $\theta$
in order to highlight the fact that the same parameters are shared across all time steps.
Different instantiations of $R$ and $O$ will result in different network
structures, and will exhibit different properties in terms of their running
times and their ability to be trained effectively using gradient-based methods.
However, they all adhere to the same abstract interface. We will provide details
of concrete instantiations of $R$ and $O$ -- the Simple RNN, the LSTM and the
GRU -- in Section \ref{sec:rnn-arch}. 
Before that, let's consider modeling with the RNN abstraction.

First, we note that the value of $\m{s_i}$ is based on the entire input $\m{x_1},...,\m{x_i}$. 
For example, by expanding the recursion for $i=4$ we get:


\begin{align*}
    \m{s_{4}}   =& R(\m{s_3},\m{x_4}) \\
                =& R(\overbrace{R(\m{s_2},\m{x_3})}^{\m{s_3}},\m{x_4}) \\
                =& R(R(\overbrace{R(\m{s_1},\m{x_2})}^{\m{s_2}},\m{x_3}),\m{x_4}) \\
                =&
                R(R(R(\overbrace{R(\m{s_0},\m{x_1})}^{\m{s_1}},\m{x_2}),\m{x_3}),\m{x_4})
\end{align*}


Thus, $\m{s_n}$ (as well as $\m{y_n}$) could be thought of as \emph{encoding} the entire input
sequence.\footnote{Note that, unless $R$ is specifically designed against this,
it is likely that the later elements of the input sequence have stronger effect
on $\m{s_n}$ than earlier ones.}
Is the encoding useful? This depends on our definition of usefulness. The job of
the network training is to set the parameters of $R$ and $O$ such that the state
conveys useful information for the task we are tying to solve.

\subsection{RNN Training}

Viewed as in Figure \ref{fig:unrolled-rnn} it is easy to see that an unrolled RNN is just a
very deep neural network (or rather, a very large \emph{computation graph} with
somewhat complex nodes), in which the same parameters are shared across many
parts of the computation. To train an RNN network, then, all we need to do is to
create the unrolled computation graph for a given input sequence, add a loss
node to the unrolled graph, and then use the backward (backpropagation)
algorithm to compute the gradients with respect to that loss. This procedure is
referred to in the RNN literature as \emph{backpropagation through time}, or
BPTT \cite{werbos1990backpropagation}.\footnote{Variants of the BPTT algorithm include unrolling the RNN
only for a fixed number of input symbols at each time: first unroll the RNN for
inputs $\m{x_{1:k}}$, resulting in $\m{s_{1:k}}$. Compute a loss, and
backpropagate the error through the network ($k$ steps back). Then, unroll the
inputs $\m{x_{k+1:2k}}$, this time using $\m{s_{k}}$ as the initial state, and
again backpropagate the error for $k$ steps, and so on. This strategy is based
on the observations that for the Simple-RNN variant, the gradients after $k$
steps tend to vanish (for large enough $k$), and so omitting them is negligible.
This procedure allows training of arbitrarily long sequences. For RNN variants
such as the LSTM or the GRU that are designed specifically to mitigate the
vanishing gradients problem, this fixed size unrolling is less motivated,
yet it is still being used, for example when doing language modeling over a book
without breaking it into sentences.} There are various ways in which the
supervision signal can be applied.

\paragraph{Acceptor} One option is to base the supervision signal only on the
final output vector, $\m{y_n}$. Viewed this way, the RNN is an
\emph{acceptor}.
We observe the final state, and then decide on an outcome.\footnote{The terminology is borrowed from Finite-State
Acceptors. However, the RNN has a potentially infinite number of states, making
it necessary to rely on a function other than a lookup table for mapping states to
decisions.}
For example, consider training an RNN to read the characters of a word one by
one and then use the final state to predict the part-of-speech of that word
(this is inspired by \cite{ling2015finding}), an RNN that reads in a sentence and,
based on the final state decides if it conveys positive or negative sentiment
(this is inspired by \cite{wang2015predicting}) or an RNN that reads in a sequence of words
and decides whether it is a valid noun-phrase.
The loss in such cases is defined in terms of a function of $\m{y_n} = O(\m{s_n})$, and the
error gradients will backpropagate through the rest of the
sequence (see Figure \ref{fig:acceptor-rnn}).\footnote{This kind of supervision signal may be hard to train for long
sequences, especially so with the Simple-RNN, because of the vanishing gradients problem.
It is also a generally hard learning task, as we do not tell the process on which parts of the input to focus.}
The loss can take any familiar form -- cross entropy, hinge, margin, etc.

\begin{figure}[ht]
    \begin{center}
    \includegraphics[width=0.6\textwidth]{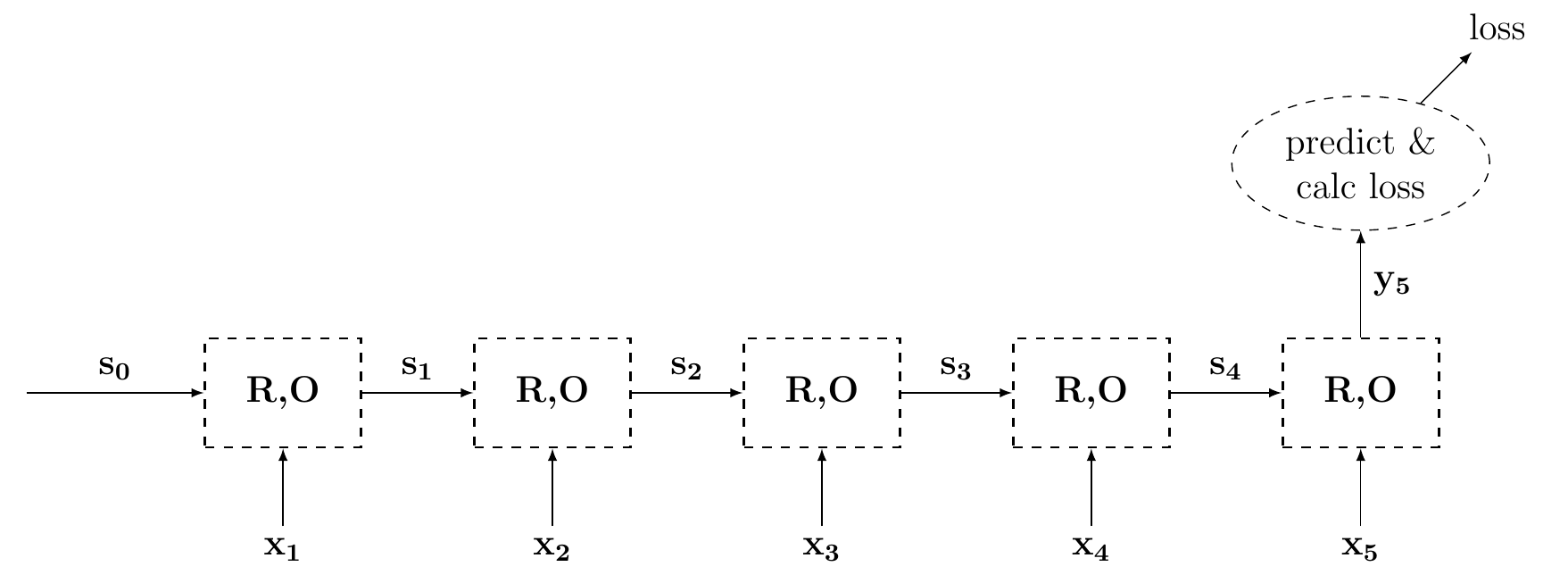} 
    \end{center}
    \caption{Acceptor RNN Training Graph.}
    \label{fig:acceptor-rnn}
\end{figure}

\paragraph{Encoder} Similar to the acceptor case, an encoder supervision uses only the final output
vector, $\m{y_n}$. However, unlike the acceptor, where a prediction is made solely on the basis of
the final vector, here the final vector is treated as an encoding of the information in the sequence,
and is used as additional information together with other signals.  For example, an extractive document
summarization system may first run over the document with an RNN, resulting in a
vector $\m{y_n}$ summarizing the entire document.  Then, $\m{y_n}$ will be used
together with with other features in order to select the sentences to be included in the summarization.

\paragraph{Transducer} Another option is to treat the RNN as a transducer,
producing an output for each input it reads in. Modeled this way, we can compute
a local loss signal $L_{local}(\m{\hat{y_i}},\m{y_i})$ for each of the outputs
$\m{\hat{y_i}}$ based on a true label $\m{y_i}$. The loss for unrolled sequence
will then be: $L(\m{\hat{y_{1:n}}},\m{y_{1:n}}) =
\sum_{i=1}^{n}{L_{local}(\m{\hat{y_i}},\m{y_i})}$, or using another combination rather
than a sum such as an average or a weighted average (see Figure
\ref{fig:transducer-rnn}). One example for such a
transducer is a sequence tagger, in which we take $\m{x_{i:n}}$ to be feature
representations for the $n$ words of a sentence, and $\m{y_i}$ as an input for
predicting the tag
assignment of word $i$ based on words $1$:$i$. 
A CCG super-tagger based on such an 
architecture provides state-of-the art CCG super-tagging results
\cite{xu2015ccg}.

\begin{figure}[ht]
    \begin{center}
    \includegraphics[width=0.6\textwidth]{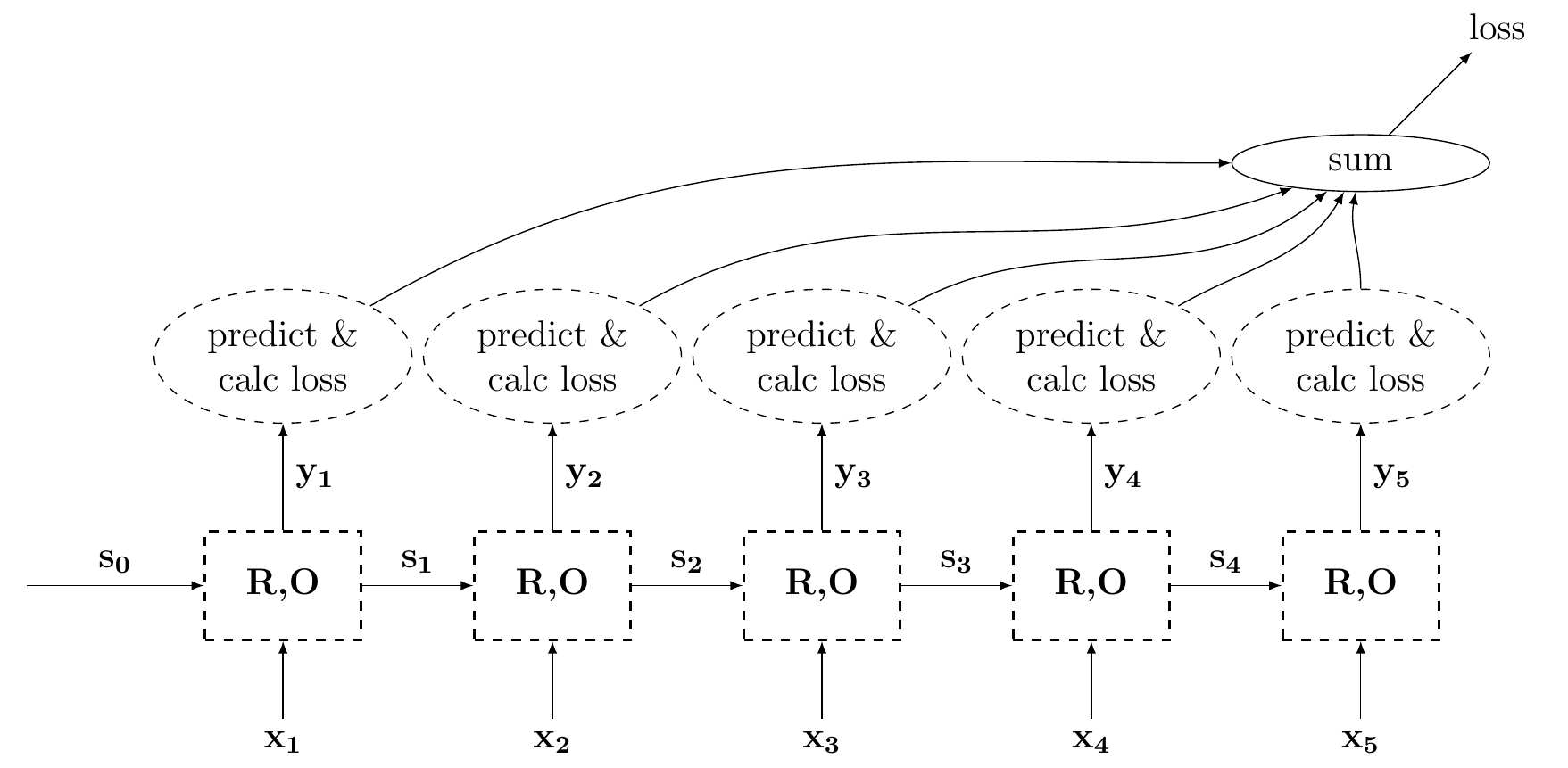} 
    \end{center}
    \caption{Transducer RNN Training Graph.}
    \label{fig:transducer-rnn}
\end{figure}

A very natural use-case of the transduction setup is for
language modeling, in which the sequence of words $\m{x_{1:i}}$ is used to
predict a distribution over the $i+1$th word.  RNN based language models are shown to
provide much better perplexities than traditional language models
\cite{mikolov2010recurrent,sundermeyer2012lstm,mikolov2012statistical}.

Using RNNs as transducers allows us to relax the Markov assumption that is
traditionally taken in language models and HMM taggers, and condition on the
entire prediction history.  The power of the ability to condition on arbitrarily long
histories is demonstrated in generative character-level RNN models, in which
a text is generated character by character, each character conditioning on the
previous ones \cite{sutskever2011generating}.  The generated texts show sensitivity to properties that are not
captured by n-gram language models, including line lengths and nested
parenthesis balancing.  For a good demonstration and analysis of the properties of
RNN-based character level
language models, see \cite{karpathy2015visualizing}.

\paragraph{Encoder - Decoder} Finally, an important special case of the encoder scenario
is the Encoder-Decoder framework \cite{cho2014properties,sutskever2014sequence}.
The RNN is used to encode the sequence
into a vector representation $\m{y_n}$, and this vector representation is then used
as auxiliary input to another RNN that is used as a decoder. For example, in a machine-translation
setup the first RNN encodes the source sentence
into a vector representation $\m{y_n}$, and then this state vector is fed into a separate
(decoder) RNN that is trained to predict (using a transducer-like language modeling objective)
the words of the target language sentence based on the previously predicted
words as well as $\m{y_n}$.
The supervision happens only for the decoder RNN,
but the gradients are propagated all the way back to the encoder RNN (see Figure
\ref{fig:encdec-rnn}).

\begin{figure}[h!]
    \begin{center}
    \includegraphics[width=0.8\textwidth]{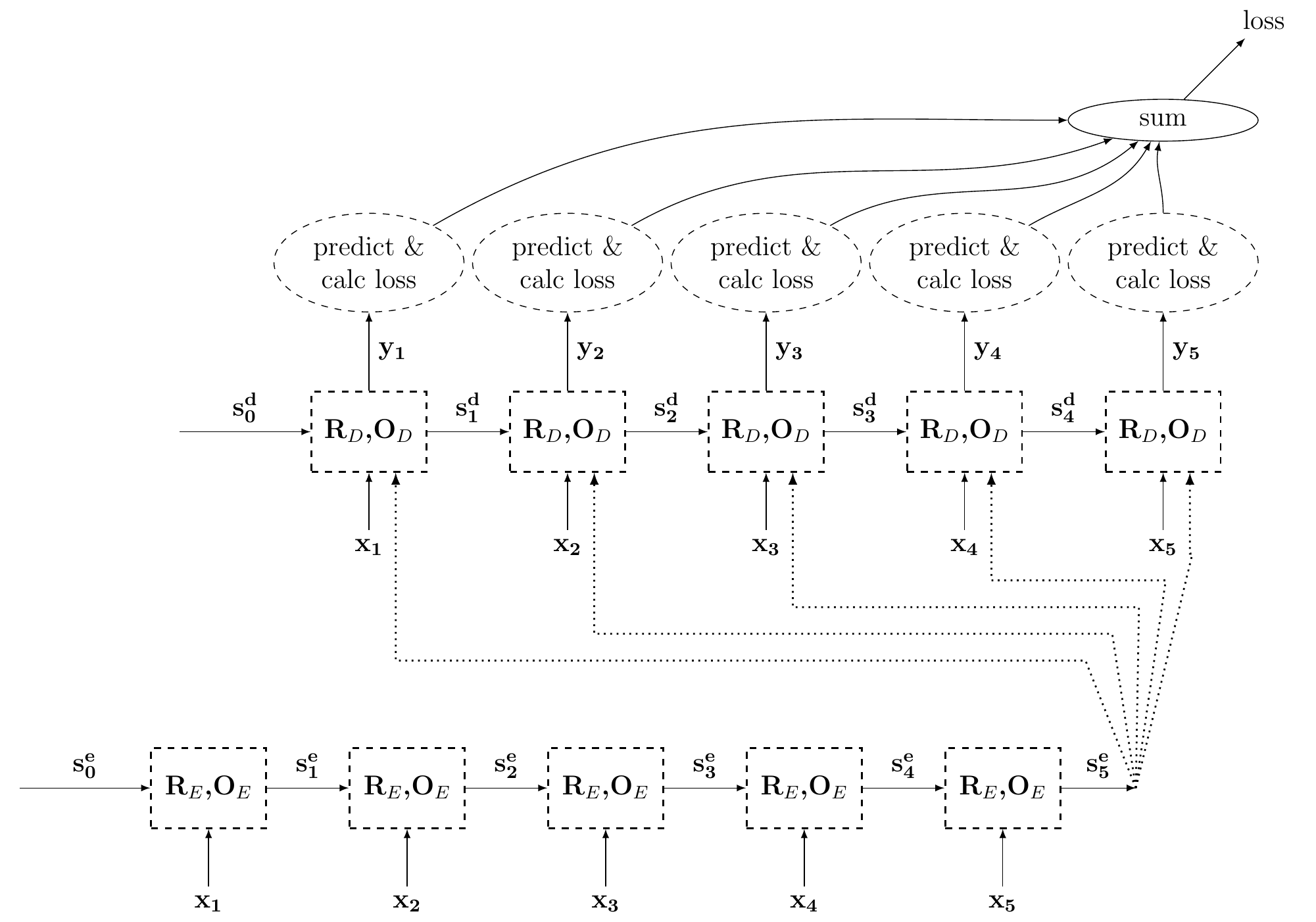} 
    \end{center}
    \caption{Encoder-Decoder RNN Training Graph.}
    \label{fig:encdec-rnn}
\end{figure}

Such an
approach was shown to be surprisingly effective for Machine Translation
\cite{sutskever2014sequence} using LSTM RNNs.  In order for this technique to work,
Sutskever et al found it effective to input the source sentence in reverse,
such that $\m{x_n}$ corresponds to the first word of the sentence. In this way,
it is easier for the second RNN to establish the relation between the first word
of the source sentence to the first word of the target sentence.
Another use-case of the encoder-decoder framework is for sequence transduction.
Here, in order to generate tags $t_1,\ldots,t_n$, an encoder RNN is first used
to encode the sentence $\m{x_{1:n}}$ into fixed sized vector. This vector is then fed as the
initial state vector of another (transducer) RNN, which is used together with $\m{x_{1:n}}$
to predict the label $t_i$ at each position $i$.  This approach was used in
\cite{filippova2015sentence} to model sentence compression by deletion.
\ygcomment{@@improve paragraph}

\subsection{Multi-layer (stacked) RNNs} RNNs can be stacked in layers, forming a
grid \cite{hihi1996hierarchical}.  Consider $k$ RNNs, $RNN_1,\ldots,RNN_k$,
where the $j$th RNN has states $\m{s_{1:n}^j}$ and outputs $\m{y_{1:n}^j}$. 
The input for the first RNN are $\m{x_{1:n}}$, while the input of the
$j$th RNN ($j \geq 2$) are the outputs of the RNN below it, $\m{y_{1:n}^{j-1}}$.
The output of the entire formation is the output of the last RNN,
$\m{y_{1:n}^k}$.  
Such layered architectures are often called \emph{deep RNNs}. A visual
representation of a 3-layer RNN is given in Figure \ref{fig:stacked-rnn}.

\begin{figure}[h!t]
    \begin{center}
    \includegraphics[width=0.6\textwidth]{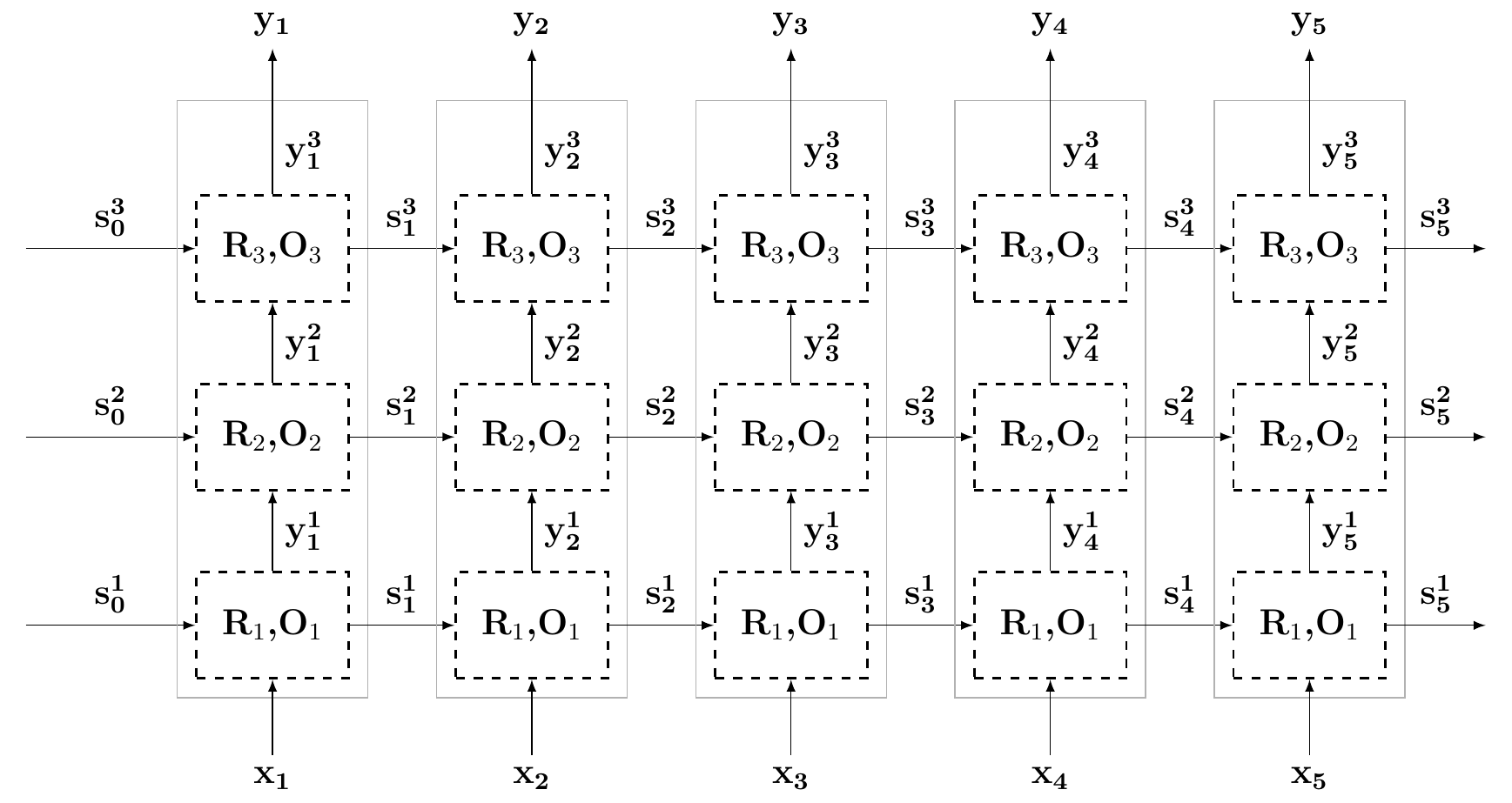} 
    \end{center}
    \caption{A 3-layer (``deep'') RNN architecture.}
    \label{fig:stacked-rnn}
\end{figure}

While it is not theoretically clear what is
the additional power gained by the deeper architecture, it was observed
empirically that deep RNNs work better than shallower ones on some tasks.
In particular, Sutskever et al \shortcite{sutskever2014sequence} report that a
4-layers deep architecture was crucial in achieving good machine-translation
performance in an encoder-decoder framework. Irsoy and Cardie
\shortcite{irsoy2014opinion} also report improved results from moving from a
one-layer BI-RNN to an architecture with several layers.
Many other works report result
using layered RNN architectures, but do not explicitly compare to
1-layer RNNs.

\subsection{BI-RNN}
A useful elaboration of an RNN is a \emph{bidirectional-RNN} (BI-RNN)
\cite{schuster1997bidirectional,graves2008supervised}.\footnote{When used with a
specific RNN architecture such as an LSTM, the model is called BI-LSTM.}
Consider the task of sequence tagging over a sentence $x_1,\ldots,x_n$.
An RNN allows us to compute a function
of the $i$th word $x_i$ based on the past -- the words $x_{1:i}$ up to and including it.
However, the following words $x_{i:n}$ may also be useful for prediction, as is
evident by the common sliding-window approach in which the focus word is
categorized based on a window of $k$ words surrounding it.
Much like the RNN relaxes the Markov assumption and allows looking
arbitrarily back into the past, the BI-RNN relaxes the fixed
window size assumption, allowing to look arbitrarily far at both the past and
the future.

Consider an input sequence $\m{x_{1:n}}$.
The BI-RNN works by maintaining two separate states, $\m{s^f_i}$ and $\m{s^b_i}$ for each input position $i$.
The \emph{forward state} $\m{s^f_i}$ is based on $\m{x_1},\m{x_2},\ldots,\m{x_i}$, while the
\emph{backward state} $\m{s^b_i}$ is based on $\m{x_n}, \m{x_{n-1}}, \ldots,\m{x_i}$.
The forward and backward states are generated by two different RNNs. The first
RNN ($R^f$, $O^f$)
is fed the input sequence $\m{x_{1:n}}$ as is, while the second RNN ($R^b$, $O^b$) is fed the
input sequence in reverse.
The state representation $\m{s_i}$ is then composed of both the forward and
backward states.

The output at position $i$ is based on the concatenation of the two output
vectors $\m{y_i} = [\m{y^f_i};\m{y^b_i}] = [O^f(\m{s^f_i});O^b(\m{s^b_i})]$,
taking into account both the past and the future. 
The vector $\m{y_i}$ can then be used directly for prediction, or fed as part of the
input to a more complex network.
While the two RNNs are run independently of each other, the error gradients at
position $i$ will flow both forward and backward through the two RNNs. A
visual
representation of the BI-RNN architecture is given in Figure \ref{fig:bi-rnn}.

\begin{figure}[ht]
    \begin{center}
    \includegraphics[width=0.8\textwidth]{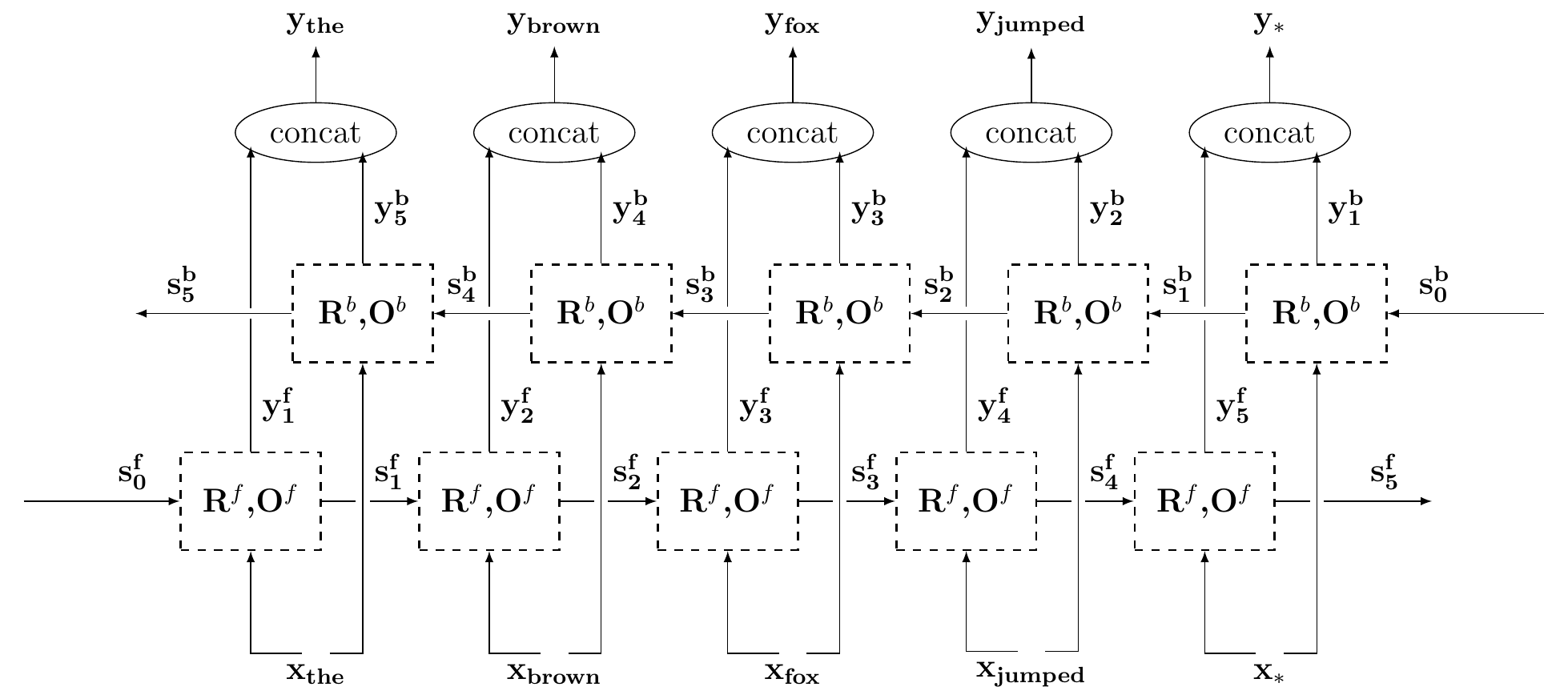} 
    \end{center}
    \caption{BI-RNN over the sentence ``the brown fox jumped .''.}
    \label{fig:bi-rnn}
\end{figure}

The use of BI-RNNs for sequence tagging was introduced to the NLP community by
Irsoy and Cardie \shortcite{irsoy2014opinion}.

\subsection{RNNs for Representing Stacks}

Some algorithms in language processing, including those for transition-based
parsing \cite{nivre2008algorithms}, require performing feature extraction over a
stack.  Instead of being confined to looking at the $k$ top-most elements of the
stack, the RNN framework can be used to provide a fixed-sized vector encoding of
the entire stack.

The main intuition is that a stack is essentially a sequence, and so the stack
state can be represented by taking the stack elements and feeding them in order
into an RNN, resulting in a final encoding of the entire stack.  In order to do
this computation efficiently (without performing an $O(n)$ stack encoding
operation each time the stack changes), the RNN state is maintained together
with the stack state.  If the stack was push-only, this would be trivial:
whenever a new element $x$ is pushed into the stack, the corresponding vector
$\m{x}$ will be used together with the RNN state $\m{s_i}$ in order to obtain a
new state $\m{s_{i+1}}$.  Dealing with pop operation is more challenging, but
can be solved by using the persistent-stack data-structure
\cite{okasaki1999purely,goldberg2013efficient}.  Persistent, or immutable,
data-structures keep old versions of themselves intact when modified.
The persistent stack construction represents a stack as a pointer
to the head of a linked list.  An empty stack is the empty list.
The push operation appends an element to the list, returning the new head.
The pop operation then returns the parent of the head, but keeping the original
list intact. From the point of view of someone who held a pointer to the
previous head, the stack did not change.  A subsequent push operation will add a
new child to the same node.  Applying this procedure throughout the lifetime of
the stack results in a tree, where the root is an empty stack and each path from
a node to the root represents an intermediary stack state. Figure
\ref{fig:immutable-stack} provides an example of such a tree.
The same process can be applied in the computation graph construction, creating
an RNN with a tree structure instead of a chain structure.  Backpropagating the
error from a given node will then affect all the elements that participated in
the stack when the node was created, in order.  Figure \ref{fig:stack-rnn} shows
the computation graph for the stack-RNN corresponding to the last state in
Figure \ref{fig:immutable-stack}.
This modeling approach was proposed independently by Dyer et al and Watanabe et
al \cite{dyer2015transitionbased,watanabe2015transitionbased} for
transition-based dependency parsing.  

\begin{figure}[ht]
    \begin{center}
    \includegraphics[width=0.9\textwidth]{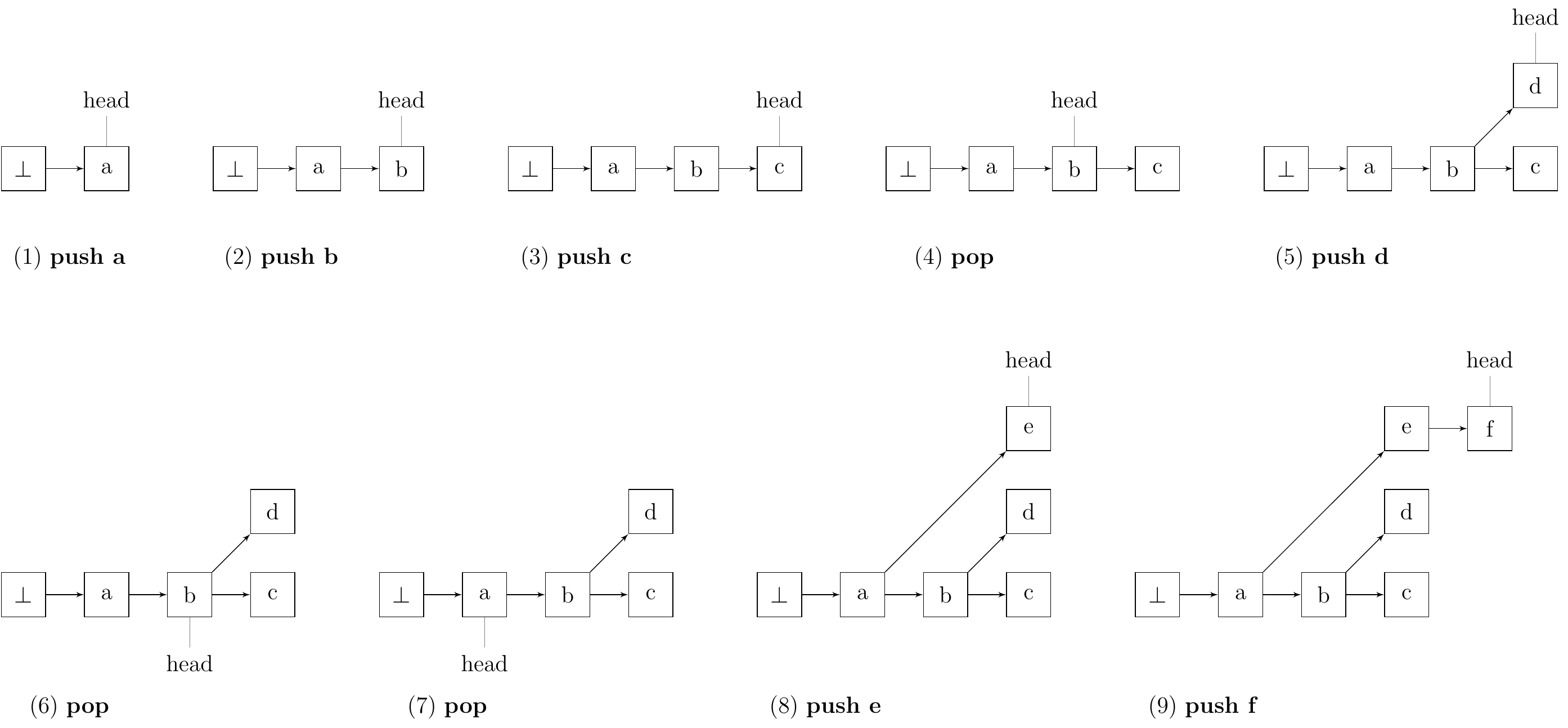}
    \end{center}
    \caption{An immutable stack construction for the sequence of operations
    \emph{push a; push b; push c; pop; push d; pop; pop; push e; push f}.}
    \label{fig:immutable-stack}
\end{figure}

\begin{figure}[ht]
    \begin{center}
    \includegraphics[width=0.7\textwidth]{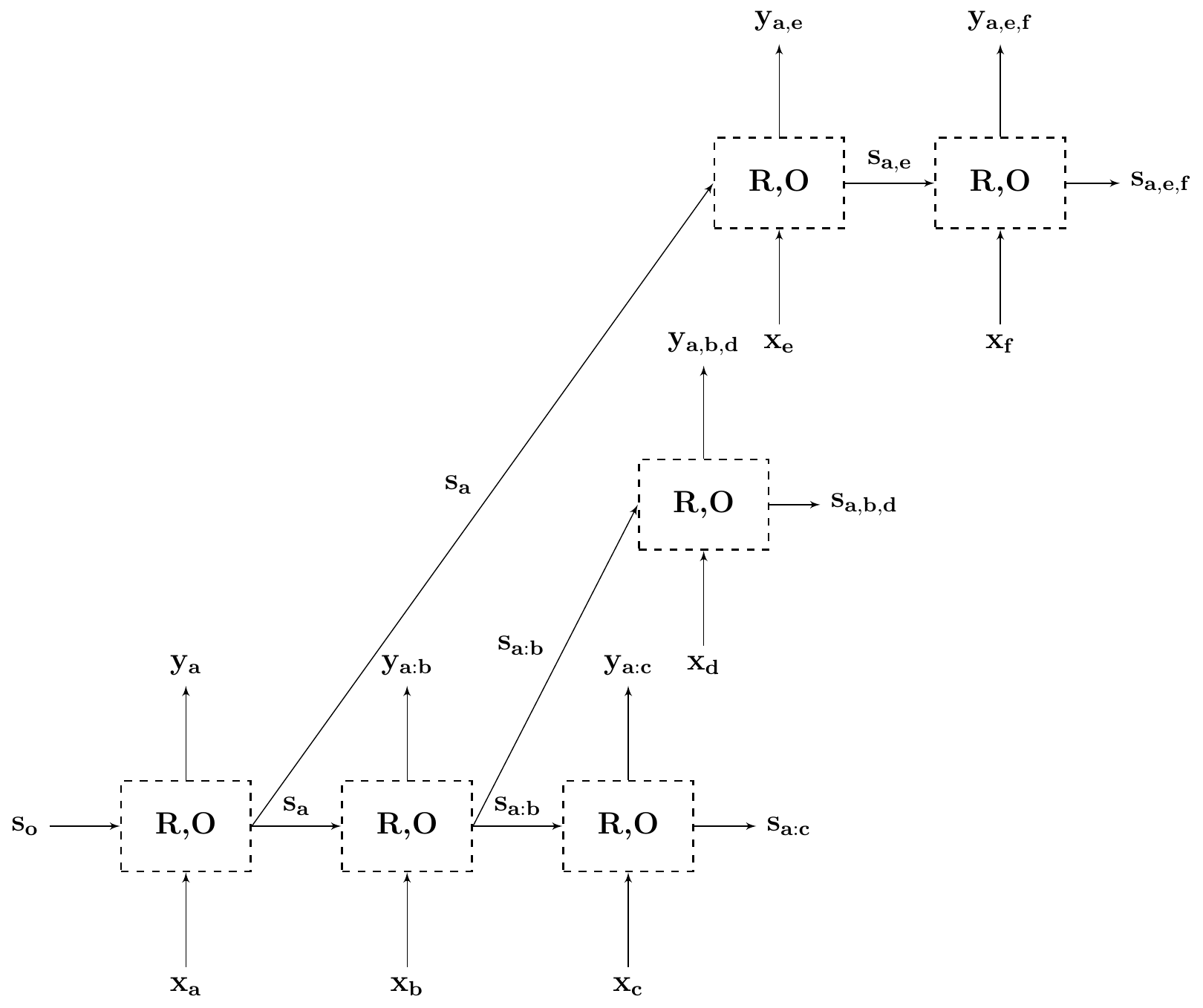}
    \end{center}
    \caption{The stack-RNN corresponding to the final state in Figure
    \ref{fig:immutable-stack}.}
    \label{fig:stack-rnn}
\end{figure}

\clearpage
\section{Concrete RNN Architectures}
\label{sec:rnn-arch}

We now turn to present three different instantiations of the abstract $RNN$
architecture discussed in the previous section, providing concrete definitions
of the functions $R$ and $O$.  These are the \emph{Simple RNN} (SRNN), the
\emph{Long Short-Term Memory} (LSTM) and the \emph{Gated Recurrent Unit} (GRU). 

\subsection{Simple RNN} 

The simplest RNN formulation, known as an Elman Network or Simple-RNN (S-RNN),
was proposed by Elman \shortcite{elman1990finding} and explored for use in language
modeling by Mikolov \shortcite{mikolov2012statistical}.  The S-RNN takes the following form:

\begin{align*}
    \m{s_i} =& R_{SRNN}(\m{s_{i-1}}, \m{x_i} ) = g(\m{x_i}\m{W^x} + \m{s_{i-1}}\m{W^s} + \m{b}) \\
    \m{y_i} =& O_{SRNN}(\m{s_{i}})             = \m{s_{i}} 
\end{align*}
\[
\m{s_i}, \m{y_i} \in \mathbb{R}^{d_s}, \,\;
\m{x_i} \in \mathbb{R}^{d_x}, \,\;
\m{W^x} \in \mathbb{R}^{d_x \times d_s}, \,\;
\m{W^s} \in \mathbb{R}^{d_s \times d_s}, \,\;
\m{b}   \in \mathbb{R}^{d_s}
\]

That is, the state at position $i$ is a linear combination of the input at
position $i$ and the previous state, passed through a non-linear activation
(commonly $\tanh$ or ReLU).
The output at position $i$ is the same as
the hidden state in that position.\footnote{Some authors treat the output at
position $i$ as a more complicated function of the state.  In our presentation,
such further transformation of the output are not considered part of the RNN,
but as separate computations that are applied to the RNNs output.
The distinction between the state and the output are needed for the LSTM
architecture, in which not all of the state is observed outside of the RNN.}

In spite of its simplicity, the Simple RNN provides strong results for sequence
tagging \cite{xu2015ccg} as well as language modeling.
For comprehensive discussion on using Simple RNNs for language modeling, see
the PhD thesis by Mikolov \shortcite{mikolov2012statistical}.

\subsection{LSTM}

The S-RNN is hard to train effectively because of the vanishing gradients problem.
Error signals (gradients) in later steps in the sequence diminish in quickly in the back-propagation
process, and do not reach earlier input signals, making it hard for the S-RNN to capture
long-range dependencies.  The Long Short-Term Memory (LSTM) architecture \cite{hochreiter1997long}
was designed to solve the vanishing gradients problem.  The main idea behind the LSTM is
to introduce as part of the state representation also ``memory cells'' (a vector) that can preserve
gradients across time.  Access to the memory cells is controlled by \emph{gating components} -- smooth
mathematical functions that simulate logical gates.  At each input state, a gate is used to decide how
much of the new input should be written to the memory cell, and how much of the current content of the memory
cell should be forgotten.
Concretely, a gate $\m{g} \in [0,1]^n$ is a vector of values in the range $[0,1]$ that is multiplied
component-wise with another vector $\m{v} \in \mathbb{R}^n$, and the result is then added to another vector.
The values of $\m{g}$ are designed to be close to either $0$ or $1$, i.e. by using a sigmoid function.
Indices in $\m{v}$ corresponding to near-one values in $\m{g}$ are allowed
to pass, while those corresponding to near-zero values are blocked.

Mathematically, the LSTM architecture is defined as:\footnote{There are many variants on the LSTM architecture presented here.  
For example, forget gates were not part of the original proposal in \cite{hochreiter1997long}, but are shown
to be an important part of the architecture. Other variants include peephole connections and gate-tying.
For an overview and comprehensive empirical comparison of various LSTM architectures see \cite{greff2015lstm}.}

\begin{align*}
    \m{s_j} = R_{LSTM}(\m{s_{j-1}}, \m{x_j}) =& [\m{c_j};\m{h_j}] \\
    \m{c_j} =& \m{c_{j-1}} \odot \m{f}  + \m{g}\odot\m{i} \\
    \m{h_j} =& \tanh(\m{c_j}) \odot \m{o}            \\
    \m{i} =& \sigma(\m{x_j}\m{W^{xi}} + \m{h_{j-1}}\m{W^{hi}}) \\
    \m{f} =& \sigma(\m{x_j}\m{W^{xf}} + \m{h_{j-1}}\m{W^{hf}}) \\
    \m{o} =& \sigma(\m{x_j}\m{W^{xo}} + \m{h_{j-1}}\m{W^{ho}}) \\
    \m{g} =& \tanh(\m{x_j}\m{W^{xg}} + \m{h_{j-1}}\m{W^{hg}}) \\
    \\
    \m{y_j} = O_{LSTM}(\m{s_j}) =& \m{h_j} 
\end{align*}

\[
\m{s_j} \in \mathbb{R}^{2\cdot d_h}, \,\;
\m{x_i} \in \mathbb{R}^{d_x}, \,\;
\m{c_j},\m{h_j},\m{i},\m{f},\m{o},\m{g} \in \mathbb{R}^{d_h}, \,\;
\m{W^{x\circ}} \in \mathbb{R}^{d_x \times d_h}, \,\;
\m{W^{h\circ}} \in \mathbb{R}^{d_h \times d_h}, \,\;
\]

The symbol $\odot$ is used to denote component-wise product.
The state at time $j$ is composed of two vectors, $\m{c_j}$ and $\m{h_j}$, where $\m{c_j}$ is the memory component and $\m{h_j}$ is
the output, or state, component.
There are three gates, $\m{i}$, $\m{f}$ and $\m{o}$, controlling for \textbf{i}nput, \textbf{f}orget and \textbf{o}utput.
The gate values are computed based on linear combinations of the current input $\m{x_j}$ and the previous state $\m{h_{j-1}}$, passed
through a sigmoid activation function.  An update candidate $\m{g}$ is computed
as a linear combination of $\m{x_j}$ and $\m{h_{j-1}}$, passed through a $\tanh$
activation function.  The memory $\m{c_j}$ is then updated: the forget gate controls how much of the previous memory to keep ($\m{c_{j-1}}\odot\m{f}$), 
and the input gate controls how much of the proposed update to keep ($\m{g}\odot\m{i}$).  Finally, the value of $\m{h_j}$ (which is also the output $\m{y_j}$) is
determined based on the content of the memory $\m{c_j}$, passed through a $\tanh$ non-linearity and controlled by the output gate.
The gating mechanisms allow for gradients related to the memory part $\m{c_j}$ to stay high across very long time ranges.
\ygcomment{Explain better?}

For further discussion on the LSTM architecture see the PhD thesis by Alex
Graves \shortcite{graves2008supervised}, as well as
Chris Olah's description.\footnote{\url{http://colah.github.io/posts/2015-08-Understanding-LSTMs/}}  For an analysis of the
behavior of an LSTM when used as a character-level language model, see
\cite{karpathy2015visualizing}.

LSTMs are currently the most successful type of RNN architecture, and they are
responsible for many state-of-the-art sequence modeling results.
The main competitor of the LSTM-RNN is the GRU, to be discussed next.

\paragraph{Practical Considerations}  When training LSTM networks, Jozefowicz et al
\shortcite{jozefowicz2015empirical} strongly recommend to always initialize the bias term
of the forget gate to be close to one.  When applying dropout to an RNN with an
LSTM, Zaremba et al \shortcite{zaremba2014recurrent} found out that it is
crucial to apply dropout only on the non-recurrent connection, i.e. only to
apply it between layers and not between sequence positions.

\subsection{GRU} 
The LSTM architecture is very effective, but also quite complicated.  The
complexity of the system makes it hard to analyze,
and also computationally expensive to work with.
The gated recurrent unit (GRU) was recently introduced by Cho et al
\shortcite{cho2014learning} as an alternative to the LSTM.  It was subsequently
shown by Chung et al \shortcite{chung2014empirical} to perform comparably to the
LSTM on several (non textual) datasets.

Like the LSTM, the GRU is also based on a gating mechanism, but with
substantially fewer gates and without a separate memory component.

\begin{align*}
    \m{s_j} = R_{GRU}(\m{s_{j-1}}, \m{x_j}) =& (\m{1} - \m{z})\odot \m{s_{j-1}} + \m{z}\odot\m{h} \\
    \m{z} =& \sigma(\m{x_j}\m{W^{xz}} + \m{h_{j-1}}\m{W^{hz}}) \\
    \m{r} =& \sigma(\m{x_j}\m{W^{xr}} + \m{h_{j-1}}\m{W^{hr}}) \\
    \m{h} =& \tanh(\m{x_j}\m{W^{xh}} + (\m{h_{j-1}}\odot\m{r})\m{W^{hg}}) \\
    \\
    \m{y_j} = O_{LSTM}(\m{s_j}) =& \m{s_j} 
\end{align*}

\[
\m{s_j} \in \mathbb{R}^{d_h}, \,\;
\m{x_i} \in \mathbb{R}^{d_x}, \,\;
\m{z},\m{r},\m{h} \in \mathbb{R}^{d_h}, \,\;
\m{W^{x\circ}} \in \mathbb{R}^{d_x \times d_h}, \,\;
\m{W^{h\circ}} \in \mathbb{R}^{d_h \times d_h}, \,\;
\]

\noindent One gate ($\m{r}$) is used to control access to the previous state $\m{s_{j-1}}$ and compute a proposed update $\m{h}$. The updated state
$\m{s_j}$ (which also serves as the output $\m{y_j}$) is then determined based on an interpolation of the previous state $\m{s_{j-1}}$ and the proposal $\m{h}$,
where the proportions of the interpolation are controlled using the gate $\m{z}$.

The GRU was shown to be effective in language modeling and machine
translation.
However, the jury between the GRU, the LSTM and possible alternative
RNN architectures is still out, and the subject is actively researched.
For an empirical exploration of the GRU and the LSTM architectures, see
\cite{jozefowicz2015empirical}.

\subsection{Other Variants}
The gated architectures of the LSTM and the GRU help in alleviating the
vanishing gradients problem of the Simple RNN, and allow these RNNs to capture
dependencies that span long time ranges.
Some researchers explore simpler architectures than the LSTM and the GRU for
achieving similar benefits.

Mikolov et al \shortcite{mikolov2014learning} observed that the matrix
multiplication $\m{s_{i-1}}\m{W^{s}}$ coupled with the nonlinearity $g$ in the update rule $R$
of the Simple RNN causes the state vector $\m{s_i}$ to undergo large changes
at each time step, prohibiting it from remembering information over long time
periods. They propose to split the state vector $\m{s_i}$ into a slow changing component
$\m{c_i}$ (``context units'') and a fast changing component $\m{h_i}$.\footnote{We
depart from the notation in \cite{mikolov2014learning} and reuse the
symbols used in the LSTM description.}
The slow changing component $\m{c_i}$ is updated according to a linear interpolation
of the input and the previous component:
$\m{c_i} =
(1-\alpha)\m{x_i}\m{W^{x1}} + \alpha\m{c_{i-1}}$, where $\alpha \in (0,1)$. This
update allows $\m{c_i}$ to accumulate the previous inputs.  The fast changing
component $\m{h_i}$ is updated similarly to the Simple RNN update rule, 
but changed to take $\m{c_i}$ into account as well:\footnote{The update rule
diverges from the S-RNN update rule also by fixing the non-linearity to be
a sigmoid function, and by not using a bias term. However, these changes are
not discussed as central to the proposal.}
$\m{h_i} = \sigma(\m{x_i}\m{W^{x2}} + \m{h_{i-1}}\m{W^h} + \m{c_i}\m{W^c})$.
Finally, the output $\m{y_i}$ is the concatenation of the slow and the fast
changing parts of the state: $\m{y_i} = [\m{c_i};\m{h_i}]$.  Mikolov et al
demonstrate that this architecture provides competitive perplexities to
the much more complex LSTM on language modeling tasks.

The approach of Mikolov et al can be interpreted as constraining the block of the
matrix $\m{W^s}$ in the S-RNN corresponding to $\m{c_i}$ to be a multiply of
the identity matrix (see Mikolov et al \shortcite{mikolov2014learning} for the
details). Le et al \cite{le2015simple} propose an even simpler approach: set
the activation function of the S-RNN to a ReLU, and initialize the biases
$\m{b}$ as zeroes and the matrix
$\m{W^s}$ as the identify matrix.
This causes an untrained RNN to copy the previous state to the current state, add
the effect of the current input $\m{x_i}$ and set the negative values to zero.
After setting this initial bias towards state copying, 
the training procedure allows $\m{W^s}$ to change freely.
Le et al demonstrate that this simple modification makes the S-RNN comparable
to an LSTM with the same number of parameters on several tasks, including
language modeling.

\clearpage
\section{Modeling Trees -- Recursive Neural Networks}
\label{sec:recnn}

The RNN is very useful for modeling sequences. In language
processing, it is often natural and desirable to work with tree structures.
The trees can be syntactic trees, discourse trees, or even trees representing
the sentiment expressed by various parts of a sentence \cite{socher2013recursive}.
We may
want to predict values based on specific tree nodes, predict values based on the
root nodes, or assign a quality score to a complete tree or part of a tree.
In other cases, we may not care about the tree structure directly but rather
reason about spans in the sentence.  In such cases, the tree is merely used as a
backbone structure which help guide the encoding process of the sequence into a
fixed size vector.

The \emph{recursive neural network} (RecNN) abstraction  \cite{pollack1990recursive},
popularized in NLP by
Richard Socher and colleagues
\cite{socher2010learning,socher2011parsing,socher2013parsing,socher2014recursive} is a generalization
of the RNN from sequences to (binary) trees.\footnote{While presented in terms
of binary parse trees, the concepts easily transfer to general recursively-defined data
structures, with the major technical challenge is the definition of an effective
form for $R$, the combination function.}

Much like the RNN encodes each sentence prefix as a state vector, the RecNN encodes
each tree-node as a state vector in $\mathbb{R}^d$.  We can then use these
state vectors either to predict values of the corresponding nodes, assign
quality values to each node, or as a semantic representation of the spans rooted
at the nodes.

The main intuition behind the recursive neural networks is that each subtree is
represented as a $d$ dimensional vector, and the representation of a node $p$
with children $c_1$ and $c_2$ is a function of the representation of the nodes:
$vec(p) = f(vec(c_1), vec(c_2))$, where $f$ is a composition function taking two
$d$-dimensional vectors and returning a single $d$-dimensional vector.  Much
like the RNN state $\m{s_i}$ is used to encode the entire sequence $\m{x_1:i}$,
the RecNN state associated with a tree node $p$ encodes the entire subtree
rooted at $p$.  See Figure \ref{fig:rec-nn} for an illustration.

\begin{figure}[h!t]
    \begin{center}
    \includegraphics[width=0.4\textwidth]{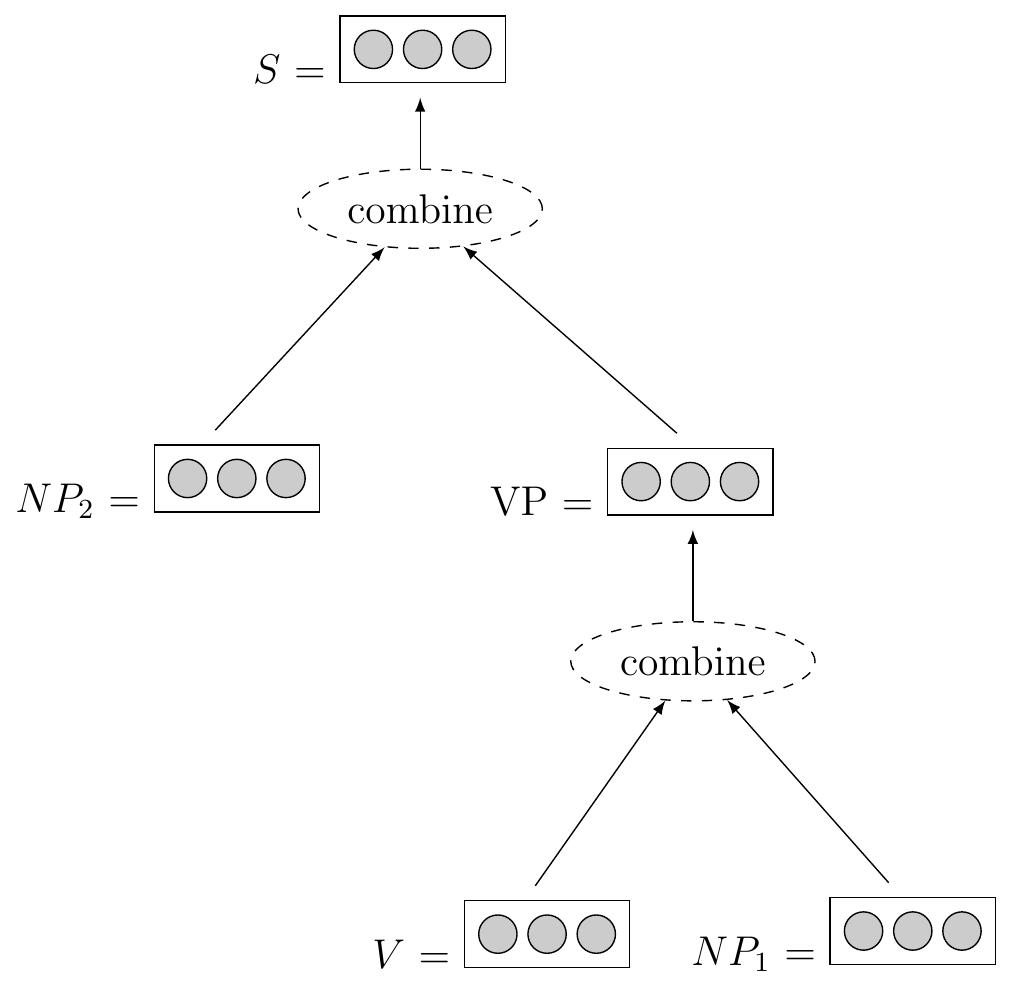}
    \end{center}
    \caption{Illustration of a recursive neural network. The representations of
    V and NP$_1$ are combined to form the representation of VP.  The
    representations of VP and NP$_2$ are then combined to form the
    representation of S.}
    \label{fig:rec-nn}
\end{figure}

\subsection{Formal Definition}
Consider a binary parse tree $\mathcal{T}$ over an $n$-word sentence. As a reminder,
an ordered, unlabeled tree over a string $x_1,\ldots,x_n$ can be represented as a unique
set of triplets $(i,k,j)$, s.t. $i \leq k \leq j$. Each such triplet indicates
that a node spanning words $x_{i:j}$ is parent of the nodes spanning $x_{i:k}$
and $x_{k+1:j}$.  Triplets of the form $(i,i,i)$ correspond to terminal symbols
at the tree leaves (the words $x_i$).  Moving from the unlabeled case to the
labeled one, we can represent a tree as a set of 6-tuples $(A \rightarrow
B,C,i,k,j)$, whereas $i$, $k$ and $j$ indicate the spans as before, and $A$, $B$
and $C$ are the node labels of of the nodes spanning $x_{i:j}$, $x_{i:k}$ and
$x_{k+1:j}$ respectively.  Here, leaf nodes have the form $(A \rightarrow
A,A,i,i,i)$, where $A$ is a pre-terminal symbol.  We refer to such tuples as
\emph{production rules}.
For an example, consider the syntactic tree for the sentence ``the boy saw her
duck''.
\begin{center}
    \Tree [.S [.NP [.Det the ] [.Noun boy ] ] [.VP [.Verb saw ] [.NP [.Det her ]
    [.Noun duck ] ] ] ]
\end{center}
Its corresponding unlabeled and labeled representations are :

\begin{center}
\begin{tabular}{c|c|l}
    Unlabeled & Labeled & Corresponding Span \\
    \hline
    (1,1,1)   & (Det, Det, Det, 1, 1, 1)       &  $x_{1:1}$ the    \\
    (2,2,2)   & (Noun, Noun, Noun, 2, 2, 2)    &  $x_{2:2}$ boy \\
    (3,3,3)   & (Verb, Verb, Verb, 3, 3, 3)    &  saw   \\
    (4,4,4)   & (Det, Det, Det, 4, 4, 4)       &  her  \\
    (5,5,5)   & (Noun, Noun, Noun, 5, 5, 5)        &  duck  \\
    (4,4,5)   & (NP, Det, Noun, 4, 4, 5)        &  her duck \\
    (3,3,5)   & (VP, Verb, NP, 3, 3, 5)        &  saw her duck \\
    (1,1,2)   & (NP, Det, Noun, 1, 1, 2)        &  the boy \\
    (1,2,5)   & (S, NP, VP, 1, 2, 5)        &  the boy saw her duck \\
\end{tabular}
\end{center}

The set of production rules above can be uniquely converted to a set tree nodes
$q^A_{i:j}$ (indicating a node with symbol $A$ over the span $x_{i:j}$) 
by simply ignoring
the elements $(B,C,k)$ in each production rule.  We are now in position to define
the Recursive Neural Network.

A Recursive Neural Network (RecNN) is a function that takes as input a parse
tree over an $n$-word sentence $x_1,\ldots,x_n$.  Each of the sentence's
words is represented as a $d$-dimensional vector $\m{x_i}$, and the tree is represented as
a set $\mathcal{T}$ of production rules $(A \rightarrow B,C,i,j,k)$.  Denote the nodes of
$\mathcal{T}$ by $q^A_{i:j}$.
The RecNN returns as output a corresponding set of
$\emph{inside state vectors}$ $\m{s^A_{i:j}}$, where each inside state vector
$\m{s^A_{i:j}} \in \mathbb{R}^d$ represents the
corresponding tree node $q^A_{i:j}$, and encodes the entire structure rooted at that node.
Like the sequence RNN, the tree shaped RecNN is defined recursively using a
function $R$, where the inside vector of a given node is defined as a function of the inside vectors of its
direct children.\footnote{Le and Zuidema \shortcite{le2014insideoutside} extend the RecNN
definition such that each node has, in addition to its inside state vector, also an 
\emph{outside state vector} representing the entire structure around the
subtree rooted at that node.  Their formulation is based on the recursive
computation of the classic inside-outside algorithm, and can be thought of as
the BI-RNN counterpart of the tree RecNN.  For details, see
\cite{le2014insideoutside}.}
Formally:

\begin{align*}
    RecNN(x_1,\ldots,x_n,\mathcal{T}) =& \{ \m{s^A_{i:j}} \in \mathbb{R}^d \,\mid\, q^A_{i:j} \in \mathcal{T} \} \\
    \m{s^A_{i:i}} =& v(x_i) \\
    \m{s^A_{i:j}} =& R(A, B, C, \m{s^B_{i:k}}, \m{s^C_{k+1:j}})  \;\;\;\;\;\; q^B_{i:k}\in \mathcal{T}, \;\; q^C_{k+1:j} \in \mathcal{T} 
\end{align*}

The function $R$ usually takes the form of a simple linear
transformation, which may or may not be followed by a non-linear activation
function $g$:

\[
R(A, B, C, \m{s^B_{i:k}}, \m{s^C_{k+1:j}}) = g([\m{s^B_{i:k}} ; \m{s^C_{k+1:j}}]\m{W})
\]

\noindent This formulation of $R$ ignores the tree labels, using the same matrix $\m{W}
\in \mathbb{R}^{2d\times d}$ for all combinations.  This may be a useful
formulation in case the node labels do not exist (e.g. when the tree does not
represent a syntactic structure with clearly defined labels) or when they are
unreliable.  However, if the labels are available, it is generally useful to
include them in the composition function.  One approach would be to introduce
\emph{label embeddings} $v(A)$ mapping each non-terminal symbol to a $d_{nt}$
dimensional vector, and change $R$ to include the embedded symbols in the
combination function:

\[
R(A, B, C, \m{s^B_{i:k}}, \m{s^C_{k+1:j}}) = g([\m{s^B_{i:k}} ; \m{s^C_{k+1:j}} ; v(A); v(B)]\m{W})
\]

\noindent (here, $\m{W} \in \mathbb{R}^{2d+2d_{nt} \times d}$). Such approach is
taken by \cite{qian2015learning}.
An alternative approach, due to \cite{socher2013parsing} is to untie the weights
according to the non-terminals, using a different composition matrix for each
$B,C$ pair of symbols:\footnote{While not explored in the literature, a trivial
extension would condition the transformation matrix also on $A$.}

\[
R(A, B, C, \m{s^B_{i:k}}, \m{s^C_{k+1:j}}) = g([\m{s^B_{i:k}} ; \m{s^C_{k+1:j}}]\m{W^{BC}})
\]

\noindent This formulation is useful when the number of non-terminal symbols (or
the number of possible symbol combinations) is relatively small, as is usually
the case with phrase-structure parse trees.  A similar model was also used by
\cite{hashimoto2013simple} to encode subtrees in semantic-relation
classification task.

\subsection{Extensions and Variations}
As all of the definitions of $R$ above suffer from the vanishing gradients
problem of the Simple RNN, several authors sought to replace it with functions
inspired by the Long Short-Term Memory (LSTM) gated architecture, resulting in
Tree-shaped LSTMs \cite{tai2015improved,zhu2015long}.
The question of optimal tree representation is still very much an open research
question, and the vast space of possible combination functions $R$ is yet to be
explored.
Other proposed variants on tree-structured RNNs includes a
\emph{recursive matrix-vector model} \cite{socher2012semantic}
and \emph{recursive neural tensor network} \cite{socher2013recursive}.
In the first variant, each word is
represented as a combination of a vector and a matrix, where the vector defines
the word's static semantic content as before, while the matrix acts as a learned
``operator'' for the word, allowing more subtle semantic compositions than the
addition and weighted averaging implied by the concatenation followed by linear
transformation function.  In the second variant, words are associated with
vectors as usual, but the composition
function becomes more expressive by basing it on tensor instead of matrix operations.

\subsection{Training Recursive Neural Networks}

The training procedure for a recursive neural network follows the same recipe as
training other forms of networks: define a loss, spell out the computation graph, compute
gradients using backpropagation\footnote{Before the introduction of the
computation graph abstraction, the specific backpropagation procedure for
computing the gradients in a RecNN as defined above was referred to as the
Back-propagation trough Structure (BPTS) algorithm \cite{goller1996learning}.}, and train the parameters using SGD.

With regard to the loss function, similar to the sequence RNN one can associate
a loss either with the root of the tree, with any given node, or with a set of
nodes, in which case the individual node's losses are combined, usually by
summation. The loss function is based on the labeled training data which
associates a label or other quantity with different tree nodes.

Additionally, one can treat the RecNN as an Encoder, whereas the inside-vector
associated with a node is taken to be an encoding of the tree rooted at that
node.  The encoding can potentially be sensitive to arbitrary properties of
the structure. The vector is then passed as input to another network.

For further discussion on recursive neural networks and
their use in natural language tasks, refer to the PhD thesis of Richard Socher
\shortcite{socher2014recursive}.

\clearpage
\section{Conclusions} 
Neural networks are powerful learners, providing opportunities
ranging from non-linear classification to non-Markovian modeling of sequences
and trees.
We hope that this exposition
help NLP researchers to incorporate neural network models in
their work and take advantage of their power.

\ignore{
\section{Acknowledgments*}
Anders Sogaard
Chris Dyer,
Miguel Ballesteros,
Matthew Honnibal,
Delip Rao,
Joakim Nivre,

Omer Levy,
Gabriel Satanovsky
Micah Shlain
Oren Melamud
Vered Schwartz
}

\bibliographystyle{theapa}
\bibliography{main}

\begin{thebibliography}{}

\bibitem[\protect\BCAY{Adel, {Vu},\ \BBA\ {Schultz}}{Adel
  et~al.}{2013}]{adel2013combination}
Adel, H., {Vu}, N.~T., \BBA\ {Schultz}, T. \BBOP2013\BBCP.
\newblock \BBOQ Combination of {Recurrent} {Neural} {Networks} and {Factored}
  {Language} {Models} for {Code}-{Switching} {Language} {Modeling}\BBCQ\
\newblock In {\Bem Proceedings of the 51st {Annual} {Meeting} of the
  {Association} for {Computational} {Linguistics} ({Volume} 2: {Short}
  {Papers})}, \BPGS\ 206--211, Sofia, {Bulgaria}. Association for
  {Computational} {Linguistics}.

\bibitem[\protect\BCAY{Ando\ \BBA\ {Zhang}}{Ando\ \BBA\
  {Zhang}}{2005a}]{ando2005highperformance}
Ando, R.\BBACOMMA\  \BBA\ {Zhang}, T. \BBOP2005a\BBCP.
\newblock \BBOQ {A} {High}-{Performance} {Semi}-{Supervised} {Learning}
  {Method} for {Text} {Chunking}\BBCQ\
\newblock In {\Bem Proceedings of the 43rd {Annual} {Meeting} of the
  {Association} for {Computational} {Linguistics} ({ACL}'05)}, \BPGS\ 1--9, Ann
  {Arbor}, {Michigan}. Association for {Computational} {Linguistics}.

\bibitem[\protect\BCAY{Ando\ \BBA\ {Zhang}}{Ando\ \BBA\
  {Zhang}}{2005b}]{ando2005framework}
Ando, R.~K.\BBACOMMA\  \BBA\ {Zhang}, T. \BBOP2005b\BBCP.
\newblock \BBOQ {A} framework for learning predictive structures from multiple
  tasks and unlabeled data\BBCQ\
\newblock {\Bem The {Journal} of {Machine} {Learning} {Research}}, {\Bem 6},
  1817--1853.

\bibitem[\protect\BCAY{Auli, {Galley}, {Quirk},\ \BBA\ {Zweig}}{Auli
  et~al.}{2013}]{auli2013joint}
Auli, M., {Galley}, M., {Quirk}, C., \BBA\ {Zweig}, G. \BBOP2013\BBCP.
\newblock \BBOQ Joint {Language} and {Translation} {Modeling} with {Recurrent}
  {Neural} {Networks}\BBCQ\
\newblock In {\Bem Proceedings of the 2013 {Conference} on {Empirical}
  {Methods} in {Natural} {Language} {Processing}}, \BPGS\ 1044--1054, Seattle,
  {Washington}, {USA}. Association for {Computational} {Linguistics}.

\bibitem[\protect\BCAY{Auli\ \BBA\ {Gao}}{Auli\ \BBA\
  {Gao}}{2014}]{auli2014decoder}
Auli, M.\BBACOMMA\  \BBA\ {Gao}, J. \BBOP2014\BBCP.
\newblock \BBOQ Decoder {Integration} and {Expected} {BLEU} {Training} for
  {Recurrent} {Neural} {Network} {Language} {Models}\BBCQ\
\newblock In {\Bem Proceedings of the 52nd {Annual} {Meeting} of the
  {Association} for {Computational} {Linguistics} ({Volume} 2: {Short}
  {Papers})}, \BPGS\ 136--142, Baltimore, {Maryland}. Association for
  {Computational} {Linguistics}.

\bibitem[\protect\BCAY{Ballesteros, {Dyer},\ \BBA\ {Smith}}{Ballesteros
  et~al.}{2015}]{ballesteros2015improved}
Ballesteros, M., {Dyer}, C., \BBA\ {Smith}, N.~A. \BBOP2015\BBCP.
\newblock \BBOQ Improved {Transition}-based {Parsing} by {Modeling}
  {Characters} instead of {Words} with {LSTMs}\BBCQ\
\newblock In {\Bem Proceedings of the 2015 {Conference} on {Empirical}
  {Methods} in {Natural} {Language} {Processing}}, \BPGS\ 349--359, Lisbon,
  {Portugal}. Association for {Computational} {Linguistics}.

\bibitem[\protect\BCAY{Bansal, {Gimpel},\ \BBA\ {Livescu}}{Bansal
  et~al.}{2014}]{bansal2014tailoring}
Bansal, M., {Gimpel}, K., \BBA\ {Livescu}, K. \BBOP2014\BBCP.
\newblock \BBOQ Tailoring {Continuous} {Word} {Representations} for
  {Dependency} {Parsing}\BBCQ\
\newblock In {\Bem Proceedings of the 52nd {Annual} {Meeting} of the
  {Association} for {Computational} {Linguistics} ({Volume} 2: {Short}
  {Papers})}, \BPGS\ 809--815, Baltimore, {Maryland}. Association for
  {Computational} {Linguistics}.

\bibitem[\protect\BCAY{Baydin, {Pearlmutter}, {Radul},\ \BBA\ {Siskind}}{Baydin
  et~al.}{2015}]{baydin2015automatic}
Baydin, A.~G., {Pearlmutter}, B.~A., {Radul}, A.~A., \BBA\ {Siskind}, J.~M.
  \BBOP2015\BBCP.
\newblock \BBOQ Automatic differentiation in machine learning: a survey\BBCQ\
\newblock {\Bem {arXiv}:1502.05767 {[}cs]}.

\bibitem[\protect\BCAY{Bengio}{Bengio}{2012}]{bengio2012practical}
Bengio, Y. \BBOP2012\BBCP.
\newblock \BBOQ Practical recommendations for gradient-based training of deep
  architectures\BBCQ\
\newblock {\Bem {arXiv}:1206.5533 {[}cs]}.

\bibitem[\protect\BCAY{Bengio, {Ducharme}, {Vincent},\ \BBA\ {Janvin}}{Bengio
  et~al.}{2003}]{bengio2003neural}
Bengio, Y., {Ducharme}, R., {Vincent}, P., \BBA\ {Janvin}, C. \BBOP2003\BBCP.
\newblock \BBOQ {A} {Neural} {Probabilistic} {Language} {Model}\BBCQ\
\newblock {\Bem {J}. {Mach}. {Learn}. {Res}.}, {\Bem 3}, 1137--1155.

\bibitem[\protect\BCAY{Bengio, {Goodfellow},\ \BBA\ {Courville}}{Bengio
  et~al.}{2015}]{bengio2015deep}
Bengio, Y., {Goodfellow}, I.~J., \BBA\ {Courville}, A. \BBOP2015\BBCP.
\newblock \BBOQ Deep {Learning}\BBCQ\
\newblock Book in preparation for MIT Press.

\bibitem[\protect\BCAY{Bitvai\ \BBA\ {Cohn}}{Bitvai\ \BBA\
  {Cohn}}{2015}]{bitvai2015nonlinear}
Bitvai, Z.\BBACOMMA\  \BBA\ {Cohn}, T. \BBOP2015\BBCP.
\newblock \BBOQ Non-{Linear} {Text} {Regression} with a {Deep} {Convolutional}
  {Neural} {Network}\BBCQ\
\newblock In {\Bem Proceedings of the 53rd {Annual} {Meeting} of the
  {Association} for {Computational} {Linguistics} and the 7th {International}
  {Joint} {Conference} on {Natural} {Language} {Processing} ({Volume} 2:
  {Short} {Papers})}, \BPGS\ 180--185, Beijing, {China}. Association for
  {Computational} {Linguistics}.

\bibitem[\protect\BCAY{Botha\ \BBA\ {Blunsom}}{Botha\ \BBA\
  {Blunsom}}{2014}]{botha2014compositional}
Botha, J.~A.\BBACOMMA\  \BBA\ {Blunsom}, P. \BBOP2014\BBCP.
\newblock \BBOQ Compositional {Morphology} for {Word} {Representations} and
  {Language} {Modelling}\BBCQ\
\newblock In {\Bem Proceedings of the 31st {International} {Conference} on
  {Machine} {Learning} ({ICML})}, Beijing, {China}.
\newblock *Award for best application paper*.

\bibitem[\protect\BCAY{Bottou}{Bottou}{2012}]{bottou2012stochastic}
Bottou, L. \BBOP2012\BBCP.
\newblock \BBOQ Stochastic gradient descent tricks\BBCQ\
\newblock In {\Bem Neural {Networks}: {Tricks} of the {Trade}}, \BPGS\
  421--436. Springer.

\bibitem[\protect\BCAY{Charniak\ \BBA\ {Johnson}}{Charniak\ \BBA\
  {Johnson}}{2005}]{charniak2005coarsetofine}
Charniak, E.\BBACOMMA\  \BBA\ {Johnson}, M. \BBOP2005\BBCP.
\newblock \BBOQ Coarse-to-{Fine} n-{Best} {Parsing} and {MaxEnt}
  {Discriminative} {Reranking}\BBCQ\
\newblock In {\Bem Proceedings of the 43rd {Annual} {Meeting} of the
  {Association} for {Computational} {Linguistics} ({ACL}'05)}, \BPGS\ 173--180,
  Ann {Arbor}, {Michigan}. Association for {Computational} {Linguistics}.

\bibitem[\protect\BCAY{Chen\ \BBA\ {Manning}}{Chen\ \BBA\
  {Manning}}{2014}]{chen2014fast}
Chen, D.\BBACOMMA\  \BBA\ {Manning}, C. \BBOP2014\BBCP.
\newblock \BBOQ {A} {Fast} and {Accurate} {Dependency} {Parser} using {Neural}
  {Networks}\BBCQ\
\newblock In {\Bem Proceedings of the 2014 {Conference} on {Empirical}
  {Methods} in {Natural} {Language} {Processing} ({EMNLP})}, \BPGS\ 740--750,
  Doha, {Qatar}. Association for {Computational} {Linguistics}.

\bibitem[\protect\BCAY{Chen, {Xu}, {Liu}, {Zeng},\ \BBA\ {Zhao}}{Chen
  et~al.}{2015}]{chen2015event}
Chen, Y., {Xu}, L., {Liu}, K., {Zeng}, D., \BBA\ {Zhao}, J. \BBOP2015\BBCP.
\newblock \BBOQ Event {Extraction} via {Dynamic} {Multi}-{Pooling}
  {Convolutional} {Neural} {Networks}\BBCQ\
\newblock In {\Bem Proceedings of the 53rd {Annual} {Meeting} of the
  {Association} for {Computational} {Linguistics} and the 7th {International}
  {Joint} {Conference} on {Natural} {Language} {Processing} ({Volume} 1: {Long}
  {Papers})}, \BPGS\ 167--176, Beijing, {China}. Association for
  {Computational} {Linguistics}.

\bibitem[\protect\BCAY{Cho, {van Merrienboer}, {Bahdanau},\ \BBA\ {Bengio}}{Cho
  et~al.}{2014a}]{cho2014properties}
Cho, K., {van Merrienboer}, B., {Bahdanau}, D., \BBA\ {Bengio}, Y.
  \BBOP2014a\BBCP.
\newblock \BBOQ On the {Properties} of {Neural} {Machine} {Translation}:
  {Encoder}{\textendash}{Decoder} {Approaches}\BBCQ\
\newblock In {\Bem Proceedings of {SSST}-8, {Eighth} {Workshop} on {Syntax},
  {Semantics} and {Structure} in {Statistical} {Translation}}, \BPGS\ 103--111,
  Doha, {Qatar}. Association for {Computational} {Linguistics}.

\bibitem[\protect\BCAY{Cho, {van Merrienboer}, {Gulcehre}, {Bahdanau},
  {Bougares}, {Schwenk},\ \BBA\ {Bengio}}{Cho et~al.}{2014b}]{cho2014learning}
Cho, K., {van Merrienboer}, B., {Gulcehre}, C., {Bahdanau}, D., {Bougares}, F.,
  {Schwenk}, H., \BBA\ {Bengio}, Y. \BBOP2014b\BBCP.
\newblock \BBOQ Learning {Phrase} {Representations} using {RNN}
  {Encoder}{\textendash}{Decoder} for {Statistical} {Machine}
  {Translation}\BBCQ\
\newblock In {\Bem Proceedings of the 2014 {Conference} on {Empirical}
  {Methods} in {Natural} {Language} {Processing} ({EMNLP})}, \BPGS\ 1724--1734,
  Doha, {Qatar}. Association for {Computational} {Linguistics}.

\bibitem[\protect\BCAY{Chrupala}{Chrupala}{2014}]{chrupala2014normalizing}
Chrupala, G. \BBOP2014\BBCP.
\newblock \BBOQ Normalizing tweets with edit scripts and recurrent neural
  embeddings\BBCQ\
\newblock In {\Bem Proceedings of the 52nd {Annual} {Meeting} of the
  {Association} for {Computational} {Linguistics} ({Volume} 2: {Short}
  {Papers})}, \BPGS\ 680--686, Baltimore, {Maryland}. Association for
  {Computational} {Linguistics}.

\bibitem[\protect\BCAY{Chung, {Gulcehre}, {Cho},\ \BBA\ {Bengio}}{Chung
  et~al.}{2014}]{chung2014empirical}
Chung, J., {Gulcehre}, C., {Cho}, K., \BBA\ {Bengio}, Y. \BBOP2014\BBCP.
\newblock \BBOQ Empirical {Evaluation} of {Gated} {Recurrent} {Neural}
  {Networks} on {Sequence} {Modeling}\BBCQ\
\newblock {\Bem {arXiv}:1412.3555 {[}cs]}.

\bibitem[\protect\BCAY{Collins}{Collins}{2002}]{collins2002discriminative}
Collins, M. \BBOP2002\BBCP.
\newblock \BBOQ Discriminative {Training} {Methods} for {Hidden} {Markov}
  {Models}: {Theory} and {Experiments} with {Perceptron} {Algorithms}\BBCQ\
\newblock In {\Bem Proceedings of the 2002 {Conference} on {Empirical}
  {Methods} in {Natural} {Language} {Processing}}, \BPGS\ 1--8. Association for
  {Computational} {Linguistics}.

\bibitem[\protect\BCAY{Collins\ \BBA\ {Koo}}{Collins\ \BBA\
  {Koo}}{2005}]{collins2005discriminative}
Collins, M.\BBACOMMA\  \BBA\ {Koo}, T. \BBOP2005\BBCP.
\newblock \BBOQ Discriminative {Reranking} for {Natural} {Language}
  {Parsing}\BBCQ\
\newblock {\Bem Computational {Linguistics}}, {\Bem 31\/}(1), 25--70.

\bibitem[\protect\BCAY{Collobert\ \BBA\ {Weston}}{Collobert\ \BBA\
  {Weston}}{2008}]{collobert2008unified}
Collobert, R.\BBACOMMA\  \BBA\ {Weston}, J. \BBOP2008\BBCP.
\newblock \BBOQ {A} unified architecture for natural language processing:
  {Deep} neural networks with multitask learning\BBCQ\
\newblock In {\Bem Proceedings of the 25th international conference on
  {Machine} learning}, \BPGS\ 160--167. {ACM}.

\bibitem[\protect\BCAY{Collobert, {Weston}, {Bottou}, {Karlen}, {Kavukcuoglu},\
  \BBA\ {Kuksa}}{Collobert et~al.}{2011}]{collobert2011natural}
Collobert, R., {Weston}, J., {Bottou}, L., {Karlen}, M., {Kavukcuoglu}, K.,
  \BBA\ {Kuksa}, P. \BBOP2011\BBCP.
\newblock \BBOQ Natural language processing (almost) from scratch\BBCQ\
\newblock {\Bem The {Journal} of {Machine} {Learning} {Research}}, {\Bem 12},
  2493--2537.

\bibitem[\protect\BCAY{Crammer\ \BBA\ {Singer}}{Crammer\ \BBA\
  {Singer}}{2002}]{crammer2002algorithmic}
Crammer, K.\BBACOMMA\  \BBA\ {Singer}, Y. \BBOP2002\BBCP.
\newblock \BBOQ On the algorithmic implementation of multiclass kernel-based
  vector machines\BBCQ\
\newblock {\Bem The {Journal} of {Machine} {Learning} {Research}}, {\Bem 2},
  265--292.

\bibitem[\protect\BCAY{Creutz\ \BBA\ {Lagus}}{Creutz\ \BBA\
  {Lagus}}{2007}]{creutz2007unsupervised}
Creutz, M.\BBACOMMA\  \BBA\ {Lagus}, K. \BBOP2007\BBCP.
\newblock \BBOQ Unsupervised {Models} for {Morpheme} {Segmentation} and
  {Morphology} {Learning}\BBCQ\
\newblock {\Bem {ACM} {Trans}. {Speech} {Lang}. {Process}.}, {\Bem 4\/}(1),
  3:1--3:34.

\bibitem[\protect\BCAY{Cybenko}{Cybenko}{1989}]{cybenko1989approximation}
Cybenko, G. \BBOP1989\BBCP.
\newblock \BBOQ Approximation by superpositions of a sigmoidal function\BBCQ\
\newblock {\Bem Mathematics of {Control}, {Signals} and {Systems}}, {\Bem
  2\/}(4), 303--314.

\bibitem[\protect\BCAY{Dahl, {Sainath},\ \BBA\ {Hinton}}{Dahl
  et~al.}{2013}]{dahl2013improving}
Dahl, G., {Sainath}, T., \BBA\ {Hinton}, G. \BBOP2013\BBCP.
\newblock \BBOQ Improving deep neural networks for {LVCSR} using rectified
  linear units and dropout\BBCQ\
\newblock In {\Bem 2013 {IEEE} {International} {Conference} on {Acoustics},
  {Speech} and {Signal} {Processing} ({ICASSP})}, \BPGS\ 8609--8613.

\bibitem[\protect\BCAY{{de Gispert}, {Iglesias},\ \BBA\ {Byrne}}{{de Gispert}
  et~al.}{2015}]{degispert2015fast}
{de Gispert}, A., {Iglesias}, G., \BBA\ {Byrne}, B. \BBOP2015\BBCP.
\newblock \BBOQ Fast and {Accurate} {Preordering} for {SMT} using {Neural}
  {Networks}\BBCQ\
\newblock In {\Bem Proceedings of the 2015 {Conference} of the {North}
  {American} {Chapter} of the {Association} for {Computational} {Linguistics}:
  {Human} {Language} {Technologies}}, \BPGS\ 1012--1017, Denver, {Colorado}.
  Association for {Computational} {Linguistics}.

\bibitem[\protect\BCAY{Dong, {Wei}, {Tan}, {Tang}, {Zhou},\ \BBA\ {Xu}}{Dong
  et~al.}{2014}]{dong2014adaptive}
Dong, L., {Wei}, F., {Tan}, C., {Tang}, D., {Zhou}, M., \BBA\ {Xu}, K.
  \BBOP2014\BBCP.
\newblock \BBOQ Adaptive {Recursive} {Neural} {Network} for {Target}-dependent
  {Twitter} {Sentiment} {Classification}\BBCQ\
\newblock In {\Bem Proceedings of the 52nd {Annual} {Meeting} of the
  {Association} for {Computational} {Linguistics} ({Volume} 2: {Short}
  {Papers})}, \BPGS\ 49--54, Baltimore, {Maryland}. Association for
  {Computational} {Linguistics}.

\bibitem[\protect\BCAY{Dong, {Wei}, {Zhou},\ \BBA\ {Xu}}{Dong
  et~al.}{2015}]{dong2015question}
Dong, L., {Wei}, F., {Zhou}, M., \BBA\ {Xu}, K. \BBOP2015\BBCP.
\newblock \BBOQ Question {Answering} over {Freebase} with {Multi}-{Column}
  {Convolutional} {Neural} {Networks}\BBCQ\
\newblock In {\Bem Proceedings of the 53rd {Annual} {Meeting} of the
  {Association} for {Computational} {Linguistics} and the 7th {International}
  {Joint} {Conference} on {Natural} {Language} {Processing} ({Volume} 1: {Long}
  {Papers})}, \BPGS\ 260--269, Beijing, {China}. Association for
  {Computational} {Linguistics}.

\bibitem[\protect\BCAY{dos {Santos}\ \BBA\ {Gatti}}{dos {Santos}\ \BBA\
  {Gatti}}{2014}]{dossantos2014deep}
dos {Santos}, C.\BBACOMMA\  \BBA\ {Gatti}, M. \BBOP2014\BBCP.
\newblock \BBOQ Deep {Convolutional} {Neural} {Networks} for {Sentiment}
  {Analysis} of {Short} {Texts}\BBCQ\
\newblock In {\Bem Proceedings of {COLING} 2014, the 25th {International}
  {Conference} on {Computational} {Linguistics}: {Technical} {Papers}}, \BPGS\
  69--78, Dublin, {Ireland}. Dublin {City} {University} and {Association} for
  {Computational} {Linguistics}.

\bibitem[\protect\BCAY{dos {Santos}, {Xiang},\ \BBA\ {Zhou}}{dos {Santos}
  et~al.}{2015}]{dossantos2015classifying}
dos {Santos}, C., {Xiang}, B., \BBA\ {Zhou}, B. \BBOP2015\BBCP.
\newblock \BBOQ Classifying {Relations} by {Ranking} with {Convolutional}
  {Neural} {Networks}\BBCQ\
\newblock In {\Bem Proceedings of the 53rd {Annual} {Meeting} of the
  {Association} for {Computational} {Linguistics} and the 7th {International}
  {Joint} {Conference} on {Natural} {Language} {Processing} ({Volume} 1: {Long}
  {Papers})}, \BPGS\ 626--634, Beijing, {China}. Association for
  {Computational} {Linguistics}.

\bibitem[\protect\BCAY{Duchi, {Hazan},\ \BBA\ {Singer}}{Duchi
  et~al.}{2011}]{duchi2011adaptive}
Duchi, J., {Hazan}, E., \BBA\ {Singer}, Y. \BBOP2011\BBCP.
\newblock \BBOQ Adaptive subgradient methods for online learning and stochastic
  optimization\BBCQ\
\newblock {\Bem The {Journal} of {Machine} {Learning} {Research}}, {\Bem 12},
  2121--2159.

\bibitem[\protect\BCAY{Duh, {Neubig}, {Sudoh},\ \BBA\ {Tsukada}}{Duh
  et~al.}{2013}]{duh2013adaptation}
Duh, K., {Neubig}, G., {Sudoh}, K., \BBA\ {Tsukada}, H. \BBOP2013\BBCP.
\newblock \BBOQ Adaptation {Data} {Selection} using {Neural} {Language}
  {Models}: {Experiments} in {Machine} {Translation}\BBCQ\
\newblock In {\Bem Proceedings of the 51st {Annual} {Meeting} of the
  {Association} for {Computational} {Linguistics} ({Volume} 2: {Short}
  {Papers})}, \BPGS\ 678--683, Sofia, {Bulgaria}. Association for
  {Computational} {Linguistics}.

\bibitem[\protect\BCAY{Durrett\ \BBA\ {Klein}}{Durrett\ \BBA\
  {Klein}}{2015}]{durrett2015neural}
Durrett, G.\BBACOMMA\  \BBA\ {Klein}, D. \BBOP2015\BBCP.
\newblock \BBOQ Neural {CRF} {Parsing}\BBCQ\
\newblock In {\Bem Proceedings of the 53rd {Annual} {Meeting} of the
  {Association} for {Computational} {Linguistics} and the 7th {International}
  {Joint} {Conference} on {Natural} {Language} {Processing} ({Volume} 1: {Long}
  {Papers})}, \BPGS\ 302--312, Beijing, {China}. Association for
  {Computational} {Linguistics}.

\bibitem[\protect\BCAY{Dyer, {Ballesteros}, {Ling}, {Matthews},\ \BBA\
  {Smith}}{Dyer et~al.}{2015}]{dyer2015transitionbased}
Dyer, C., {Ballesteros}, M., {Ling}, W., {Matthews}, A., \BBA\ {Smith}, N.~A.
  \BBOP2015\BBCP.
\newblock \BBOQ Transition-{Based} {Dependency} {Parsing} with {Stack} {Long}
  {Short}-{Term} {Memory}\BBCQ\
\newblock In {\Bem Proceedings of the 53rd {Annual} {Meeting} of the
  {Association} for {Computational} {Linguistics} and the 7th {International}
  {Joint} {Conference} on {Natural} {Language} {Processing} ({Volume} 1: {Long}
  {Papers})}, \BPGS\ 334--343, Beijing, {China}. Association for
  {Computational} {Linguistics}.

\bibitem[\protect\BCAY{Elman}{Elman}{1990}]{elman1990finding}
Elman, J.~L. \BBOP1990\BBCP.
\newblock \BBOQ Finding {Structure} in {Time}\BBCQ\
\newblock {\Bem Cognitive {Science}}, {\Bem 14\/}(2), 179--211.

\bibitem[\protect\BCAY{Faruqui\ \BBA\ {Dyer}}{Faruqui\ \BBA\
  {Dyer}}{2014}]{faruqui2014improving}
Faruqui, M.\BBACOMMA\  \BBA\ {Dyer}, C. \BBOP2014\BBCP.
\newblock \BBOQ Improving {Vector} {Space} {Word} {Representations} {Using}
  {Multilingual} {Correlation}\BBCQ\
\newblock In {\Bem Proceedings of the 14th {Conference} of the {European}
  {Chapter} of the {Association} for {Computational} {Linguistics}}, \BPGS\
  462--471, Gothenburg, {Sweden}. Association for {Computational}
  {Linguistics}.

\bibitem[\protect\BCAY{Filippova, {Alfonseca}, {Colmenares}, {Kaiser},\ \BBA\
  {Vinyals}}{Filippova et~al.}{2015}]{filippova2015sentence}
Filippova, K., {Alfonseca}, E., {Colmenares}, C.~A., {Kaiser}, L., \BBA\
  {Vinyals}, O. \BBOP2015\BBCP.
\newblock \BBOQ Sentence {Compression} by {Deletion} with {LSTMs}\BBCQ\
\newblock In {\Bem Proceedings of the 2015 {Conference} on {Empirical}
  {Methods} in {Natural} {Language} {Processing}}, \BPGS\ 360--368, Lisbon,
  {Portugal}. Association for {Computational} {Linguistics}.

\bibitem[\protect\BCAY{Gal\ \BBA\ {Ghahramani}}{Gal\ \BBA\
  {Ghahramani}}{2015}]{gal2015dropout}
Gal, Y.\BBACOMMA\  \BBA\ {Ghahramani}, Z. \BBOP2015\BBCP.
\newblock \BBOQ Dropout as a {Bayesian} {Approximation}: {Representing} {Model}
  {Uncertainty} in {Deep} {Learning}\BBCQ\
\newblock {\Bem {arXiv}:1506.02142 {[}cs, stat]}.

\bibitem[\protect\BCAY{Gao, {Pantel}, {Gamon}, {He},\ \BBA\ {Deng}}{Gao
  et~al.}{2014}]{gao2014modeling}
Gao, J., {Pantel}, P., {Gamon}, M., {He}, X., \BBA\ {Deng}, L. \BBOP2014\BBCP.
\newblock \BBOQ Modeling {Interestingness} with {Deep} {Neural}
  {Networks}\BBCQ\
\newblock In {\Bem Proceedings of the 2014 {Conference} on {Empirical}
  {Methods} in {Natural} {Language} {Processing} ({EMNLP})}, \BPGS\ 2--13,
  Doha, {Qatar}. Association for {Computational} {Linguistics}.

\bibitem[\protect\BCAY{Gim{\'e}nez\ \BBA\ {M}{\`a}rquez}{Gim{\'e}nez\ \BBA\
  {M}{\`a}rquez}{2004}]{gimenez2004svmtool}
Gim{\'e}nez, J.\BBACOMMA\  \BBA\ {M}{\`a}rquez, L. \BBOP2004\BBCP.
\newblock \BBOQ {SVMTool}: {A} general {POS} tagger generator based on
  {Support} {Vector} {Machines}\BBCQ\
\newblock In {\Bem Proceedings of the 4th {LREC}}, Lisbon, {Portugal}.

\bibitem[\protect\BCAY{Glorot\ \BBA\ {Bengio}}{Glorot\ \BBA\
  {Bengio}}{2010}]{glorot2010understanding}
Glorot, X.\BBACOMMA\  \BBA\ {Bengio}, Y. \BBOP2010\BBCP.
\newblock \BBOQ Understanding the difficulty of training deep feedforward
  neural networks\BBCQ\
\newblock In {\Bem International conference on artificial intelligence and
  statistics}, \BPGS\ 249--256.

\bibitem[\protect\BCAY{Glorot, {Bordes},\ \BBA\ {Bengio}}{Glorot
  et~al.}{2011}]{glorot2011deep}
Glorot, X., {Bordes}, A., \BBA\ {Bengio}, Y. \BBOP2011\BBCP.
\newblock \BBOQ Deep sparse rectifier neural networks\BBCQ\
\newblock In {\Bem International {Conference} on {Artificial} {Intelligence}
  and {Statistics}}, \BPGS\ 315--323.

\bibitem[\protect\BCAY{Goldberg\ \BBA\ {Elhadad}}{Goldberg\ \BBA\
  {Elhadad}}{2010}]{goldberg2010efficient}
Goldberg, Y.\BBACOMMA\  \BBA\ {Elhadad}, M. \BBOP2010\BBCP.
\newblock \BBOQ An {Efficient} {Algorithm} for {Easy}-{First}
  {Non}-{Directional} {Dependency} {Parsing}\BBCQ\
\newblock In {\Bem Human {Language} {Technologies}: {The} 2010 {Annual}
  {Conference} of the {North} {American} {Chapter} of the {Association} for
  {Computational} {Linguistics}}, \BPGS\ 742--750, Los {Angeles}, {California}.
  Association for {Computational} {Linguistics}.

\bibitem[\protect\BCAY{Goldberg\ \BBA\ {Levy}}{Goldberg\ \BBA\
  {Levy}}{2014}]{goldberg2014word2vec}
Goldberg, Y.\BBACOMMA\  \BBA\ {Levy}, O. \BBOP2014\BBCP.
\newblock \BBOQ word2vec {Explained}: deriving {Mikolov} et al.'s
  negative-sampling word-embedding method\BBCQ\
\newblock {\Bem {arXiv}:1402.3722 {[}cs, stat]}.

\bibitem[\protect\BCAY{Goldberg\ \BBA\ {Nivre}}{Goldberg\ \BBA\
  {Nivre}}{2013}]{goldberg2013training}
Goldberg, Y.\BBACOMMA\  \BBA\ {Nivre}, J. \BBOP2013\BBCP.
\newblock \BBOQ Training {Deterministic} {Parsers} with {Non}-{Deterministic}
  {Oracles}\BBCQ\
\newblock {\Bem Transactions of the {Association} for {Computational}
  {Linguistics}}, {\Bem 1\/}(0), 403--414.

\bibitem[\protect\BCAY{Goldberg, {Zhao},\ \BBA\ {Huang}}{Goldberg
  et~al.}{2013}]{goldberg2013efficient}
Goldberg, Y., {Zhao}, K., \BBA\ {Huang}, L. \BBOP2013\BBCP.
\newblock \BBOQ Efficient {Implementation} of {Beam}-{Search} {Incremental}
  {Parsers}\BBCQ\
\newblock In {\Bem Proceedings of the 51st {Annual} {Meeting} of the
  {Association} for {Computational} {Linguistics} ({Volume} 2: {Short}
  {Papers})}, \BPGS\ 628--633, Sofia, {Bulgaria}. Association for
  {Computational} {Linguistics}.

\bibitem[\protect\BCAY{Goller\ \BBA\ {K}{\"u}chler}{Goller\ \BBA\
  {K}{\"u}chler}{1996}]{goller1996learning}
Goller, C.\BBACOMMA\  \BBA\ {K}{\"u}chler, A. \BBOP1996\BBCP.
\newblock \BBOQ Learning {Task}-{Dependent} {Distributed} {Representations} by
  {Backpropagation} {Through} {Structure}\BBCQ\
\newblock In {\Bem In {Proc}. of the {ICNN}-96}, \BPGS\ 347--352. {IEEE}.

\bibitem[\protect\BCAY{Graves}{Graves}{2008}]{graves2008supervised}
Graves, A. \BBOP2008\BBCP.
\newblock {\Bem Supervised sequence labelling with recurrent neural networks}.
\newblock Ph.D.\ thesis, Technische {Universit}{\"a}t {M}{\"u}nchen.

\bibitem[\protect\BCAY{Greff, {Srivastava}, {Koutn}{\'\i}k, {Steunebrink},\
  \BBA\ {Schmidhuber}}{Greff et~al.}{2015}]{greff2015lstm}
Greff, K., {Srivastava}, R.~K., {Koutn}{\'\i}k, J., {Steunebrink}, B.~R., \BBA\
  {Schmidhuber}, J. \BBOP2015\BBCP.
\newblock \BBOQ {LSTM}: {A} {Search} {Space} {Odyssey}\BBCQ\
\newblock {\Bem {arXiv}:1503.04069 {[}cs]}.

\bibitem[\protect\BCAY{{Hal Daum{\'e} III}, {Langford},\ \BBA\ {Marcu}}{{Hal
  Daum{\'e} III} et~al.}{2009}]{daume09searn}
{Hal Daum{\'e} III}, {Langford}, J., \BBA\ {Marcu}, D. \BBOP2009\BBCP.
\newblock \BBOQ Search-based {Structured} {Prediction}\BBCQ\
\newblock {\Bem Machine {Learning} {Journal} ({MLJ})}.

\bibitem[\protect\BCAY{Harris}{Harris}{1954}]{harris1954distributional}
Harris, Z. \BBOP1954\BBCP.
\newblock \BBOQ Distributional {Structure}\BBCQ\
\newblock {\Bem Word}, {\Bem 10\/}(23), 146--162.

\bibitem[\protect\BCAY{Hashimoto, {Miwa}, {Tsuruoka},\ \BBA\
  {Chikayama}}{Hashimoto et~al.}{2013}]{hashimoto2013simple}
Hashimoto, K., {Miwa}, M., {Tsuruoka}, Y., \BBA\ {Chikayama}, T.
  \BBOP2013\BBCP.
\newblock \BBOQ Simple {Customization} of {Recursive} {Neural} {Networks} for
  {Semantic} {Relation} {Classification}\BBCQ\
\newblock In {\Bem Proceedings of the 2013 {Conference} on {Empirical}
  {Methods} in {Natural} {Language} {Processing}}, \BPGS\ 1372--1376, Seattle,
  {Washington}, {USA}. Association for {Computational} {Linguistics}.

\bibitem[\protect\BCAY{He, {Zhang}, {Ren},\ \BBA\ {Sun}}{He
  et~al.}{2015}]{he2015delving}
He, K., {Zhang}, X., {Ren}, S., \BBA\ {Sun}, J. \BBOP2015\BBCP.
\newblock \BBOQ Delving {Deep} into {Rectifiers}: {Surpassing} {Human}-{Level}
  {Performance} on {ImageNet} {Classification}\BBCQ\
\newblock {\Bem {arXiv}:1502.01852 {[}cs]}.

\bibitem[\protect\BCAY{Henderson, {Thomson},\ \BBA\ {Young}}{Henderson
  et~al.}{2013}]{henderson2013deep}
Henderson, M., {Thomson}, B., \BBA\ {Young}, S. \BBOP2013\BBCP.
\newblock \BBOQ Deep {Neural} {Network} {Approach} for the {Dialog} {State}
  {Tracking} {Challenge}\BBCQ\
\newblock In {\Bem Proceedings of the {SIGDIAL} 2013 {Conference}}, \BPGS\
  467--471, Metz, {France}. Association for {Computational} {Linguistics}.

\bibitem[\protect\BCAY{Hermann\ \BBA\ {Blunsom}}{Hermann\ \BBA\
  {Blunsom}}{2013}]{hermann2013role}
Hermann, K.~M.\BBACOMMA\  \BBA\ {Blunsom}, P. \BBOP2013\BBCP.
\newblock \BBOQ The {Role} of {Syntax} in {Vector} {Space} {Models} of
  {Compositional} {Semantics}\BBCQ\
\newblock In {\Bem Proceedings of the 51st {Annual} {Meeting} of the
  {Association} for {Computational} {Linguistics} ({Volume} 1: {Long}
  {Papers})}, \BPGS\ 894--904, Sofia, {Bulgaria}. Association for
  {Computational} {Linguistics}.

\bibitem[\protect\BCAY{Hermann\ \BBA\ {Blunsom}}{Hermann\ \BBA\
  {Blunsom}}{2014}]{hermann2014multilingual}
Hermann, K.~M.\BBACOMMA\  \BBA\ {Blunsom}, P. \BBOP2014\BBCP.
\newblock \BBOQ Multilingual {Models} for {Compositional} {Distributed}
  {Semantics}\BBCQ\
\newblock In {\Bem Proceedings of the 52nd {Annual} {Meeting} of the
  {Association} for {Computational} {Linguistics} ({Volume} 1: {Long}
  {Papers})}, \BPGS\ 58--68, Baltimore, {Maryland}. Association for
  {Computational} {Linguistics}.

\bibitem[\protect\BCAY{Hihi\ \BBA\ {Bengio}}{Hihi\ \BBA\
  {Bengio}}{1996}]{hihi1996hierarchical}
Hihi, S.~E.\BBACOMMA\  \BBA\ {Bengio}, Y. \BBOP1996\BBCP.
\newblock \BBOQ Hierarchical {Recurrent} {Neural} {Networks} for {Long}-{Term}
  {Dependencies}\BBCQ\
\newblock In Touretzky, D.~S., {Mozer}, M.~C., \BBA\ {Hasselmo}, M.~E.\BEDS,
  {\Bem Advances in {Neural} {Information} {Processing} {Systems} 8}, \BPGS\
  493--499. {MIT} {Press}.

\bibitem[\protect\BCAY{Hinton, {Srivastava}, {Krizhevsky}, {Sutskever},\ \BBA\
  {Salakhutdinov}}{Hinton et~al.}{2012}]{hinton2012improving}
Hinton, G.~E., {Srivastava}, N., {Krizhevsky}, A., {Sutskever}, I., \BBA\
  {Salakhutdinov}, R.~R. \BBOP2012\BBCP.
\newblock \BBOQ Improving neural networks by preventing co-adaptation of
  feature detectors\BBCQ\
\newblock {\Bem {arXiv}:1207.0580 {[}cs]}.

\bibitem[\protect\BCAY{Hochreiter\ \BBA\ {Schmidhuber}}{Hochreiter\ \BBA\
  {Schmidhuber}}{1997}]{hochreiter1997long}
Hochreiter, S.\BBACOMMA\  \BBA\ {Schmidhuber}, J. \BBOP1997\BBCP.
\newblock \BBOQ Long short-term memory\BBCQ\
\newblock {\Bem Neural computation}, {\Bem 9\/}(8), 1735--1780.

\bibitem[\protect\BCAY{Hornik, {Stinchcombe},\ \BBA\ {White}}{Hornik
  et~al.}{1989}]{hornik1989multilayer}
Hornik, K., {Stinchcombe}, M., \BBA\ {White}, H. \BBOP1989\BBCP.
\newblock \BBOQ Multilayer feedforward networks are universal
  approximators\BBCQ\
\newblock {\Bem Neural {Networks}}, {\Bem 2\/}(5), 359--366.

\bibitem[\protect\BCAY{Irsoy\ \BBA\ {Cardie}}{Irsoy\ \BBA\
  {Cardie}}{2014}]{irsoy2014opinion}
Irsoy, O.\BBACOMMA\  \BBA\ {Cardie}, C. \BBOP2014\BBCP.
\newblock \BBOQ Opinion {Mining} with {Deep} {Recurrent} {Neural}
  {Networks}\BBCQ\
\newblock In {\Bem Proceedings of the 2014 {Conference} on {Empirical}
  {Methods} in {Natural} {Language} {Processing} ({EMNLP})}, \BPGS\ 720--728,
  Doha, {Qatar}. Association for {Computational} {Linguistics}.

\bibitem[\protect\BCAY{Iyyer, {Boyd}-{Graber}, {Claudino}, {Socher},\ \BBA\
  {Daum}{\'e} {III}}{Iyyer et~al.}{2014a}]{iyyer2014neural}
Iyyer, M., {Boyd}-{Graber}, J., {Claudino}, L., {Socher}, R., \BBA\ {Daum}{\'e}
  {III}, H. \BBOP2014a\BBCP.
\newblock \BBOQ {A} {Neural} {Network} for {Factoid} {Question} {Answering}
  over {Paragraphs}\BBCQ\
\newblock In {\Bem Proceedings of the 2014 {Conference} on {Empirical}
  {Methods} in {Natural} {Language} {Processing} ({EMNLP})}, \BPGS\ 633--644,
  Doha, {Qatar}. Association for {Computational} {Linguistics}.

\bibitem[\protect\BCAY{Iyyer, {Enns}, {Boyd}-{Graber},\ \BBA\ {Resnik}}{Iyyer
  et~al.}{2014b}]{iyyer2014political}
Iyyer, M., {Enns}, P., {Boyd}-{Graber}, J., \BBA\ {Resnik}, P. \BBOP2014b\BBCP.
\newblock \BBOQ Political {Ideology} {Detection} {Using} {Recursive} {Neural}
  {Networks}\BBCQ\
\newblock In {\Bem Proceedings of the 52nd {Annual} {Meeting} of the
  {Association} for {Computational} {Linguistics} ({Volume} 1: {Long}
  {Papers})}, \BPGS\ 1113--1122, Baltimore, {Maryland}. Association for
  {Computational} {Linguistics}.

\bibitem[\protect\BCAY{Iyyer, {Manjunatha}, {Boyd}-{Graber},\ \BBA\ {Daum}{\'e}
  {III}}{Iyyer et~al.}{2015}]{iyyer2015deep}
Iyyer, M., {Manjunatha}, V., {Boyd}-{Graber}, J., \BBA\ {Daum}{\'e} {III}, H.
  \BBOP2015\BBCP.
\newblock \BBOQ Deep {Unordered} {Composition} {Rivals} {Syntactic} {Methods}
  for {Text} {Classification}\BBCQ\
\newblock In {\Bem Proceedings of the 53rd {Annual} {Meeting} of the
  {Association} for {Computational} {Linguistics} and the 7th {International}
  {Joint} {Conference} on {Natural} {Language} {Processing} ({Volume} 1: {Long}
  {Papers})}, \BPGS\ 1681--1691, Beijing, {China}. Association for
  {Computational} {Linguistics}.

\bibitem[\protect\BCAY{Johnson\ \BBA\ {Zhang}}{Johnson\ \BBA\
  {Zhang}}{2014}]{johnson2014effective}
Johnson, R.\BBACOMMA\  \BBA\ {Zhang}, T. \BBOP2014\BBCP.
\newblock \BBOQ Effective {Use} of {Word} {Order} for {Text} {Categorization}
  with {Convolutional} {Neural} {Networks}\BBCQ\
\newblock {\Bem {arXiv}:1412.1058 {[}cs, stat]}.

\bibitem[\protect\BCAY{Johnson\ \BBA\ {Zhang}}{Johnson\ \BBA\
  {Zhang}}{2015}]{johnson2015effective}
Johnson, R.\BBACOMMA\  \BBA\ {Zhang}, T. \BBOP2015\BBCP.
\newblock \BBOQ Effective {Use} of {Word} {Order} for {Text} {Categorization}
  with {Convolutional} {Neural} {Networks}\BBCQ\
\newblock In {\Bem Proceedings of the 2015 {Conference} of the {North}
  {American} {Chapter} of the {Association} for {Computational} {Linguistics}:
  {Human} {Language} {Technologies}}, \BPGS\ 103--112, Denver, {Colorado}.
  Association for {Computational} {Linguistics}.

\bibitem[\protect\BCAY{Jozefowicz, {Zaremba},\ \BBA\ {Sutskever}}{Jozefowicz
  et~al.}{2015}]{jozefowicz2015empirical}
Jozefowicz, R., {Zaremba}, W., \BBA\ {Sutskever}, I. \BBOP2015\BBCP.
\newblock \BBOQ An {Empirical} {Exploration} of {Recurrent} {Network}
  {Architectures}\BBCQ\
\newblock In {\Bem Proceedings of the 32nd {International} {Conference} on
  {Machine} {Learning} ({ICML}-15)}, \BPGS\ 2342--2350.

\bibitem[\protect\BCAY{Kalchbrenner, {Grefenstette},\ \BBA\
  {Blunsom}}{Kalchbrenner et~al.}{2014}]{kalchbrenner2014convolutional}
Kalchbrenner, N., {Grefenstette}, E., \BBA\ {Blunsom}, P. \BBOP2014\BBCP.
\newblock \BBOQ {A} {Convolutional} {Neural} {Network} for {Modelling}
  {Sentences}\BBCQ\
\newblock In {\Bem Proceedings of the 52nd {Annual} {Meeting} of the
  {Association} for {Computational} {Linguistics} ({Volume} 1: {Long}
  {Papers})}, \BPGS\ 655--665, Baltimore, {Maryland}. Association for
  {Computational} {Linguistics}.

\bibitem[\protect\BCAY{Karpathy, {Johnson},\ \BBA\ {Li}}{Karpathy
  et~al.}{2015}]{karpathy2015visualizing}
Karpathy, A., {Johnson}, J., \BBA\ {Li}, F.-F. \BBOP2015\BBCP.
\newblock \BBOQ Visualizing and {Understanding} {Recurrent} {Networks}\BBCQ\
\newblock {\Bem {arXiv}:1506.02078 {[}cs]}.

\bibitem[\protect\BCAY{Kim}{Kim}{2014}]{kim2014convolutional}
Kim, Y. \BBOP2014\BBCP.
\newblock \BBOQ Convolutional {Neural} {Networks} for {Sentence}
  {Classification}\BBCQ\
\newblock In {\Bem Proceedings of the 2014 {Conference} on {Empirical}
  {Methods} in {Natural} {Language} {Processing} ({EMNLP})}, \BPGS\ 1746--1751,
  Doha, {Qatar}. Association for {Computational} {Linguistics}.

\bibitem[\protect\BCAY{Kingma\ \BBA\ {Ba}}{Kingma\ \BBA\
  {Ba}}{2014}]{kingma2014adam}
Kingma, D.\BBACOMMA\  \BBA\ {Ba}, J. \BBOP2014\BBCP.
\newblock \BBOQ Adam: {A} {Method} for {Stochastic} {Optimization}\BBCQ\
\newblock {\Bem {arXiv}:1412.6980 {[}cs]}.

\bibitem[\protect\BCAY{Krizhevsky, {Sutskever},\ \BBA\ {Hinton}}{Krizhevsky
  et~al.}{2012}]{krizhevsky2012imagenet}
Krizhevsky, A., {Sutskever}, I., \BBA\ {Hinton}, G.~E. \BBOP2012\BBCP.
\newblock \BBOQ {ImageNet} {Classification} with {Deep} {Convolutional}
  {Neural} {Networks}\BBCQ\
\newblock In Pereira, F., {Burges}, C. J.~C., {Bottou}, L., \BBA\ {Weinberger},
  K.~Q.\BEDS, {\Bem Advances in {Neural} {Information} {Processing} {Systems}
  25}, \BPGS\ 1097--1105. Curran {Associates}, {Inc}.

\bibitem[\protect\BCAY{Kudo\ \BBA\ {Matsumoto}}{Kudo\ \BBA\
  {Matsumoto}}{2003}]{kudo2003fast}
Kudo, T.\BBACOMMA\  \BBA\ {Matsumoto}, Y. \BBOP2003\BBCP.
\newblock \BBOQ Fast {Methods} for {Kernel}-based {Text} {Analysis}\BBCQ\
\newblock In {\Bem Proceedings of the 41st {Annual} {Meeting} on {Association}
  for {Computational} {Linguistics} - {Volume} 1}, {ACL} '03, \BPGS\ 24--31,
  Stroudsburg, {PA}, {USA}. Association for {Computational} {Linguistics}.

\bibitem[\protect\BCAY{Le\ \BBA\ {Zuidema}}{Le\ \BBA\
  {Zuidema}}{2014}]{le2014insideoutside}
Le, P.\BBACOMMA\  \BBA\ {Zuidema}, W. \BBOP2014\BBCP.
\newblock \BBOQ The {Inside}-{Outside} {Recursive} {Neural} {Network} model for
  {Dependency} {Parsing}\BBCQ\
\newblock In {\Bem Proceedings of the 2014 {Conference} on {Empirical}
  {Methods} in {Natural} {Language} {Processing} ({EMNLP})}, \BPGS\ 729--739,
  Doha, {Qatar}. Association for {Computational} {Linguistics}.

\bibitem[\protect\BCAY{Le\ \BBA\ {Zuidema}}{Le\ \BBA\
  {Zuidema}}{2015}]{le2015forest}
Le, P.\BBACOMMA\  \BBA\ {Zuidema}, W. \BBOP2015\BBCP.
\newblock \BBOQ The {Forest} {Convolutional} {Network}: {Compositional}
  {Distributional} {Semantics} with a {Neural} {Chart} and without
  {Binarization}\BBCQ\
\newblock In {\Bem Proceedings of the 2015 {Conference} on {Empirical}
  {Methods} in {Natural} {Language} {Processing}}, \BPGS\ 1155--1164, Lisbon,
  {Portugal}. Association for {Computational} {Linguistics}.

\bibitem[\protect\BCAY{Le, {Jaitly},\ \BBA\ {Hinton}}{Le
  et~al.}{2015}]{le2015simple}
Le, Q.~V., {Jaitly}, N., \BBA\ {Hinton}, G.~E. \BBOP2015\BBCP.
\newblock \BBOQ {A} {Simple} {Way} to {Initialize} {Recurrent} {Networks} of
  {Rectified} {Linear} {Units}\BBCQ\
\newblock {\Bem {arXiv}:1504.00941 {[}cs]}.

\bibitem[\protect\BCAY{{LeCun}\ \BBA\ {Bengio}}{{LeCun}\ \BBA\
  {Bengio}}{1995}]{lecun1995convolutional}
{LeCun}, Y.\BBACOMMA\  \BBA\ {Bengio}, Y. \BBOP1995\BBCP.
\newblock \BBOQ Convolutional {Networks} for {Images}, {Speech}, and
  {Time}-{Series}\BBCQ\
\newblock In Arbib, M.~A.\BED, {\Bem The {Handbook} of {Brain} {Theory} and
  {Neural} {Networks}}. {MIT} {Press}.

\bibitem[\protect\BCAY{{LeCun}, {Bottou}, {Orr},\ \BBA\ {Muller}}{{LeCun}
  et~al.}{1998a}]{lecun1998efficient}
{LeCun}, Y., {Bottou}, L., {Orr}, G., \BBA\ {Muller}, K. \BBOP1998a\BBCP.
\newblock \BBOQ Efficient {BackProp}\BBCQ\
\newblock In Orr, G.\BBACOMMA\  \BBA\ {K}, M.\BEDS, {\Bem Neural {Networks}:
  {Tricks} of the trade}. Springer.

\bibitem[\protect\BCAY{Lecun, {Bottou}, {Bengio},\ \BBA\ {Haffner}}{Lecun
  et~al.}{1998b}]{lecun1998gradient}
Lecun, Y., {Bottou}, L., {Bengio}, Y., \BBA\ {Haffner}, P. \BBOP1998b\BBCP.
\newblock
\newblock \BBOQ Gradient {Based} {Learning} {Applied} to {Pattern}
  {Recognition}\BBCQ.

\bibitem[\protect\BCAY{{LeCun}, {Chopra}, {Hadsell}, {Ranzato},\ \BBA\
  {Huang}}{{LeCun} et~al.}{2006}]{lecun2006tutorial}
{LeCun}, Y., {Chopra}, S., {Hadsell}, R., {Ranzato}, M., \BBA\ {Huang}, F.
  \BBOP2006\BBCP.
\newblock \BBOQ {A} tutorial on energy-based learning\BBCQ\
\newblock {\Bem Predicting structured data}, {\Bem 1}, 0.

\bibitem[\protect\BCAY{{LeCun}\ \BBA\ {Huang}}{{LeCun}\ \BBA\
  {Huang}}{2005}]{lecun2005loss}
{LeCun}, Y.\BBACOMMA\  \BBA\ {Huang}, F. \BBOP2005\BBCP.
\newblock \BBOQ Loss functions for discriminative training of energybased
  models\BBCQ.
\newblock {AIStats}.

\bibitem[\protect\BCAY{Levy\ \BBA\ {Goldberg}}{Levy\ \BBA\
  {Goldberg}}{2014a}]{levy2014dependencybased}
Levy, O.\BBACOMMA\  \BBA\ {Goldberg}, Y. \BBOP2014a\BBCP.
\newblock \BBOQ Dependency-{Based} {Word} {Embeddings}\BBCQ\
\newblock In {\Bem Proceedings of the 52nd {Annual} {Meeting} of the
  {Association} for {Computational} {Linguistics} ({Volume} 2: {Short}
  {Papers})}, \BPGS\ 302--308, Baltimore, {Maryland}. Association for
  {Computational} {Linguistics}.

\bibitem[\protect\BCAY{Levy\ \BBA\ {Goldberg}}{Levy\ \BBA\
  {Goldberg}}{2014b}]{levy2014neural}
Levy, O.\BBACOMMA\  \BBA\ {Goldberg}, Y. \BBOP2014b\BBCP.
\newblock \BBOQ Neural {Word} {Embedding} as {Implicit} {Matrix}
  {Factorization}\BBCQ\
\newblock In Ghahramani, Z., {Welling}, M., {Cortes}, C., {Lawrence}, N.~D.,
  \BBA\ {Weinberger}, K.~Q.\BEDS, {\Bem Advances in {Neural} {Information}
  {Processing} {Systems} 27}, \BPGS\ 2177--2185. Curran {Associates}, {Inc}.

\bibitem[\protect\BCAY{Levy, {Goldberg},\ \BBA\ {Dagan}}{Levy
  et~al.}{2015}]{levy2015improving}
Levy, O., {Goldberg}, Y., \BBA\ {Dagan}, I. \BBOP2015\BBCP.
\newblock \BBOQ Improving {Distributional} {Similarity} with {Lessons}
  {Learned} from {Word} {Embeddings}\BBCQ\
\newblock {\Bem Transactions of the {Association} for {Computational}
  {Linguistics}}, {\Bem 3\/}(0), 211--225.

\bibitem[\protect\BCAY{Lewis\ \BBA\ {Steedman}}{Lewis\ \BBA\
  {Steedman}}{2014}]{lewis2014improved}
Lewis, M.\BBACOMMA\  \BBA\ {Steedman}, M. \BBOP2014\BBCP.
\newblock \BBOQ Improved {CCG} {Parsing} with {Semi}-supervised
  {Supertagging}\BBCQ\
\newblock {\Bem Transactions of the {Association} for {Computational}
  {Linguistics}}, {\Bem 2\/}(0), 327--338.

\bibitem[\protect\BCAY{Li, {Li},\ \BBA\ {Hovy}}{Li
  et~al.}{2014}]{li2014recursive}
Li, J., {Li}, R., \BBA\ {Hovy}, E. \BBOP2014\BBCP.
\newblock \BBOQ Recursive {Deep} {Models} for {Discourse} {Parsing}\BBCQ\
\newblock In {\Bem Proceedings of the 2014 {Conference} on {Empirical}
  {Methods} in {Natural} {Language} {Processing} ({EMNLP})}, \BPGS\ 2061--2069,
  Doha, {Qatar}. Association for {Computational} {Linguistics}.

\bibitem[\protect\BCAY{Ling, {Dyer}, {Black},\ \BBA\ {Trancoso}}{Ling
  et~al.}{2015a}]{ling2015twotoo}
Ling, W., {Dyer}, C., {Black}, A.~W., \BBA\ {Trancoso}, I. \BBOP2015a\BBCP.
\newblock \BBOQ Two/{Too} {Simple} {Adaptations} of {Word2Vec} for {Syntax}
  {Problems}\BBCQ\
\newblock In {\Bem Proceedings of the 2015 {Conference} of the {North}
  {American} {Chapter} of the {Association} for {Computational} {Linguistics}:
  {Human} {Language} {Technologies}}, \BPGS\ 1299--1304, Denver, {Colorado}.
  Association for {Computational} {Linguistics}.

\bibitem[\protect\BCAY{Ling, {Dyer}, {Black}, {Trancoso}, {Fermandez}, {Amir},
  {Marujo},\ \BBA\ {Luis}}{Ling et~al.}{2015b}]{ling2015finding}
Ling, W., {Dyer}, C., {Black}, A.~W., {Trancoso}, I., {Fermandez}, R., {Amir},
  S., {Marujo}, L., \BBA\ {Luis}, T. \BBOP2015b\BBCP.
\newblock \BBOQ Finding {Function} in {Form}: {Compositional} {Character}
  {Models} for {Open} {Vocabulary} {Word} {Representation}\BBCQ\
\newblock In {\Bem Proceedings of the 2015 {Conference} on {Empirical}
  {Methods} in {Natural} {Language} {Processing}}, \BPGS\ 1520--1530, Lisbon,
  {Portugal}. Association for {Computational} {Linguistics}.

\bibitem[\protect\BCAY{Liu, {Wei}, {Li}, {Ji}, {Zhou},\ \BBA\ {WANG}}{Liu
  et~al.}{2015}]{liu2015dependencybased}
Liu, Y., {Wei}, F., {Li}, S., {Ji}, H., {Zhou}, M., \BBA\ {WANG}, H.
  \BBOP2015\BBCP.
\newblock \BBOQ {A} {Dependency}-{Based} {Neural} {Network} for {Relation}
  {Classification}\BBCQ\
\newblock In {\Bem Proceedings of the 53rd {Annual} {Meeting} of the
  {Association} for {Computational} {Linguistics} and the 7th {International}
  {Joint} {Conference} on {Natural} {Language} {Processing} ({Volume} 2:
  {Short} {Papers})}, \BPGS\ 285--290, Beijing, {China}. Association for
  {Computational} {Linguistics}.

\bibitem[\protect\BCAY{Ma, {Zhang},\ \BBA\ {Zhu}}{Ma
  et~al.}{2014}]{ma2014tagging}
Ma, J., {Zhang}, Y., \BBA\ {Zhu}, J. \BBOP2014\BBCP.
\newblock \BBOQ Tagging {The} {Web}: {Building} {A} {Robust} {Web} {Tagger}
  with {Neural} {Network}\BBCQ\
\newblock In {\Bem Proceedings of the 52nd {Annual} {Meeting} of the
  {Association} for {Computational} {Linguistics} ({Volume} 1: {Long}
  {Papers})}, \BPGS\ 144--154, Baltimore, {Maryland}. Association for
  {Computational} {Linguistics}.

\bibitem[\protect\BCAY{Ma, {Huang}, {Zhou},\ \BBA\ {Xiang}}{Ma
  et~al.}{2015}]{ma2015dependencybased}
Ma, M., {Huang}, L., {Zhou}, B., \BBA\ {Xiang}, B. \BBOP2015\BBCP.
\newblock \BBOQ Dependency-based {Convolutional} {Neural} {Networks} for
  {Sentence} {Embedding}\BBCQ\
\newblock In {\Bem Proceedings of the 53rd {Annual} {Meeting} of the
  {Association} for {Computational} {Linguistics} and the 7th {International}
  {Joint} {Conference} on {Natural} {Language} {Processing} ({Volume} 2:
  {Short} {Papers})}, \BPGS\ 174--179, Beijing, {China}. Association for
  {Computational} {Linguistics}.

\bibitem[\protect\BCAY{{McCallum}, {Freitag},\ \BBA\ {Pereira}}{{McCallum}
  et~al.}{2000}]{mccallum2000maximum}
{McCallum}, A., {Freitag}, D., \BBA\ {Pereira}, F.~C. \BBOP2000\BBCP.
\newblock \BBOQ Maximum {Entropy} {Markov} {Models} for {Information}
  {Extraction} and {Segmentation}.\BBCQ\
\newblock In {\Bem {ICML}}, \lowercase{\BVOL}~17, \BPGS\ 591--598.

\bibitem[\protect\BCAY{Mikolov, {Chen}, {Corrado},\ \BBA\ {Dean}}{Mikolov
  et~al.}{2013}]{mikolov2013efficient}
Mikolov, T., {Chen}, K., {Corrado}, G., \BBA\ {Dean}, J. \BBOP2013\BBCP.
\newblock \BBOQ Efficient {Estimation} of {Word} {Representations} in {Vector}
  {Space}\BBCQ\
\newblock {\Bem {arXiv}:1301.3781 {[}cs]}.

\bibitem[\protect\BCAY{Mikolov, {Joulin}, {Chopra}, {Mathieu},\ \BBA\
  {Ranzato}}{Mikolov et~al.}{2014}]{mikolov2014learning}
Mikolov, T., {Joulin}, A., {Chopra}, S., {Mathieu}, M., \BBA\ {Ranzato}, M.
  \BBOP2014\BBCP.
\newblock \BBOQ Learning {Longer} {Memory} in {Recurrent} {Neural}
  {Networks}\BBCQ\
\newblock {\Bem {arXiv}:1412.7753 {[}cs]}.

\bibitem[\protect\BCAY{Mikolov, {Karafi}{\'a}t, {Burget}, {Cernocky},\ \BBA\
  {Khudanpur}}{Mikolov et~al.}{2010}]{mikolov2010recurrent}
Mikolov, T., {Karafi}{\'a}t, M., {Burget}, L., {Cernocky}, J., \BBA\
  {Khudanpur}, S. \BBOP2010\BBCP.
\newblock \BBOQ Recurrent neural network based language model.\BBCQ\
\newblock In {\Bem {INTERSPEECH} 2010, 11th {Annual} {Conference} of the
  {International} {Speech} {Communication} {Association}, {Makuhari}, {Chiba},
  {Japan}, {September} 26-30, 2010}, \BPGS\ 1045--1048.

\bibitem[\protect\BCAY{Mikolov, {Kombrink}, {Luk{\'a}{\v s} Burget}, {\v
  C}ernocky,\ \BBA\ {Khudanpur}}{Mikolov et~al.}{2011}]{mikolov2011extensions}
Mikolov, T., {Kombrink}, S., {Luk{\'a}{\v s} Burget}, {\v C}ernocky, J.~H.,
  \BBA\ {Khudanpur}, S. \BBOP2011\BBCP.
\newblock \BBOQ Extensions of recurrent neural network language model\BBCQ\
\newblock In {\Bem Acoustics, {Speech} and {Signal} {Processing} ({ICASSP}),
  2011 {IEEE} {International} {Conference} on}, \BPGS\ 5528--5531. {IEEE}.

\bibitem[\protect\BCAY{Mikolov, {Sutskever}, {Chen}, {Corrado},\ \BBA\
  {Dean}}{Mikolov et~al.}{2013}]{mikolov2013distributed}
Mikolov, T., {Sutskever}, I., {Chen}, K., {Corrado}, G.~S., \BBA\ {Dean}, J.
  \BBOP2013\BBCP.
\newblock \BBOQ Distributed {Representations} of {Words} and {Phrases} and
  their {Compositionality}\BBCQ\
\newblock In Burges, C. J.~C., {Bottou}, L., {Welling}, M., {Ghahramani}, Z.,
  \BBA\ {Weinberger}, K.~Q.\BEDS, {\Bem Advances in {Neural} {Information}
  {Processing} {Systems} 26}, \BPGS\ 3111--3119. Curran {Associates}, {Inc}.

\bibitem[\protect\BCAY{Mikolov}{Mikolov}{2012}]{mikolov2012statistical}
Mikolov, T. \BBOP2012\BBCP.
\newblock {\Bem Statistical language models based on neural networks}.
\newblock Ph.D.\ thesis, Ph. {D}. thesis, {Brno} {University} of {Technology}.

\bibitem[\protect\BCAY{Mnih\ \BBA\ {Kavukcuoglu}}{Mnih\ \BBA\
  {Kavukcuoglu}}{2013}]{mnih2013learning}
Mnih, A.\BBACOMMA\  \BBA\ {Kavukcuoglu}, K. \BBOP2013\BBCP.
\newblock \BBOQ Learning word embeddings efficiently with noise-contrastive
  estimation\BBCQ\
\newblock In Burges, C. J.~C., {Bottou}, L., {Welling}, M., {Ghahramani}, Z.,
  \BBA\ {Weinberger}, K.~Q.\BEDS, {\Bem Advances in {Neural} {Information}
  {Processing} {Systems} 26}, \BPGS\ 2265--2273. Curran {Associates}, {Inc}.

\bibitem[\protect\BCAY{Mrk{\v s}i{\'c}, {\'O}~{S}{\'e}aghdha, {Thomson},
  {Gasic}, {Su}, {Vandyke}, {Wen},\ \BBA\ {Young}}{Mrk{\v s}i{\'c}
  et~al.}{2015}]{mrksic2015multidomain}
Mrk{\v s}i{\'c}, N., {\'O}~{S}{\'e}aghdha, D., {Thomson}, B., {Gasic}, M.,
  {Su}, P.-H., {Vandyke}, D., {Wen}, T.-H., \BBA\ {Young}, S. \BBOP2015\BBCP.
\newblock \BBOQ Multi-domain {Dialog} {State} {Tracking} using {Recurrent}
  {Neural} {Networks}\BBCQ\
\newblock In {\Bem Proceedings of the 53rd {Annual} {Meeting} of the
  {Association} for {Computational} {Linguistics} and the 7th {International}
  {Joint} {Conference} on {Natural} {Language} {Processing} ({Volume} 2:
  {Short} {Papers})}, \BPGS\ 794--799, Beijing, {China}. Association for
  {Computational} {Linguistics}.

\bibitem[\protect\BCAY{Neidinger}{Neidinger}{2010}]{neidinger2010introduction}
Neidinger, R. \BBOP2010\BBCP.
\newblock \BBOQ Introduction to {Automatic} {Differentiation} and {MATLAB}
  {Object}-{Oriented} {Programming}\BBCQ\
\newblock {\Bem {SIAM} {Review}}, {\Bem 52\/}(3), 545--563.

\bibitem[\protect\BCAY{Nguyen\ \BBA\ {Grishman}}{Nguyen\ \BBA\
  {Grishman}}{2015}]{nguyen2015event}
Nguyen, T.~H.\BBACOMMA\  \BBA\ {Grishman}, R. \BBOP2015\BBCP.
\newblock \BBOQ Event {Detection} and {Domain} {Adaptation} with
  {Convolutional} {Neural} {Networks}\BBCQ\
\newblock In {\Bem Proceedings of the 53rd {Annual} {Meeting} of the
  {Association} for {Computational} {Linguistics} and the 7th {International}
  {Joint} {Conference} on {Natural} {Language} {Processing} ({Volume} 2:
  {Short} {Papers})}, \BPGS\ 365--371, Beijing, {China}. Association for
  {Computational} {Linguistics}.

\bibitem[\protect\BCAY{Nivre}{Nivre}{2008}]{nivre2008algorithms}
Nivre, J. \BBOP2008\BBCP.
\newblock \BBOQ Algorithms for {Deterministic} {Incremental} {Dependency}
  {Parsing}\BBCQ\
\newblock {\Bem Computational {Linguistics}}, {\Bem 34\/}(4), 513--553.

\bibitem[\protect\BCAY{Okasaki}{Okasaki}{1999}]{okasaki1999purely}
Okasaki, C. \BBOP1999\BBCP.
\newblock {\Bem Purely {Functional} {Data} {Structures}}.
\newblock Cambridge {University} {Press}, Cambridge, {U}.{K}.; {New} {York}.

\bibitem[\protect\BCAY{Pascanu, {Mikolov},\ \BBA\ {Bengio}}{Pascanu
  et~al.}{2012}]{pascanu2012difficulty}
Pascanu, R., {Mikolov}, T., \BBA\ {Bengio}, Y. \BBOP2012\BBCP.
\newblock \BBOQ On the difficulty of training {Recurrent} {Neural}
  {Networks}\BBCQ\
\newblock {\Bem {arXiv}:1211.5063 {[}cs]}.

\bibitem[\protect\BCAY{Pei, {Ge},\ \BBA\ {Chang}}{Pei
  et~al.}{2015}]{pei2015effective}
Pei, W., {Ge}, T., \BBA\ {Chang}, B. \BBOP2015\BBCP.
\newblock \BBOQ An {Effective} {Neural} {Network} {Model} for {Graph}-based
  {Dependency} {Parsing}\BBCQ\
\newblock In {\Bem Proceedings of the 53rd {Annual} {Meeting} of the
  {Association} for {Computational} {Linguistics} and the 7th {International}
  {Joint} {Conference} on {Natural} {Language} {Processing} ({Volume} 1: {Long}
  {Papers})}, \BPGS\ 313--322, Beijing, {China}. Association for
  {Computational} {Linguistics}.

\bibitem[\protect\BCAY{Pennington, {Socher},\ \BBA\ {Manning}}{Pennington
  et~al.}{2014}]{pennington2014glove}
Pennington, J., {Socher}, R., \BBA\ {Manning}, C. \BBOP2014\BBCP.
\newblock \BBOQ Glove: {Global} {Vectors} for {Word} {Representation}\BBCQ\
\newblock In {\Bem Proceedings of the 2014 {Conference} on {Empirical}
  {Methods} in {Natural} {Language} {Processing} ({EMNLP})}, \BPGS\ 1532--1543,
  Doha, {Qatar}. Association for {Computational} {Linguistics}.

\bibitem[\protect\BCAY{Pollack}{Pollack}{1990}]{pollack1990recursive}
Pollack, J.~B. \BBOP1990\BBCP.
\newblock \BBOQ Recursive {Distributed} {Representations}\BBCQ\
\newblock {\Bem Artificial {Intelligence}}, {\Bem 46}, 77--105.

\bibitem[\protect\BCAY{Polyak}{Polyak}{1964}]{polyak1964methods}
Polyak, B.~T. \BBOP1964\BBCP.
\newblock \BBOQ Some methods of speeding up the convergence of iteration
  methods\BBCQ\
\newblock {\Bem {USSR} {Computational} {Mathematics} and {Mathematical}
  {Physics}}, {\Bem 4\/}(5), 1 -- 17.

\bibitem[\protect\BCAY{Qian, {Tian}, {Huang}, {Liu}, {Zhu},\ \BBA\ {Zhu}}{Qian
  et~al.}{2015}]{qian2015learning}
Qian, Q., {Tian}, B., {Huang}, M., {Liu}, Y., {Zhu}, X., \BBA\ {Zhu}, X.
  \BBOP2015\BBCP.
\newblock \BBOQ Learning {Tag} {Embeddings} and {Tag}-specific {Composition}
  {Functions} in {Recursive} {Neural} {Network}\BBCQ\
\newblock In {\Bem Proceedings of the 53rd {Annual} {Meeting} of the
  {Association} for {Computational} {Linguistics} and the 7th {International}
  {Joint} {Conference} on {Natural} {Language} {Processing} ({Volume} 1: {Long}
  {Papers})}, \BPGS\ 1365--1374, Beijing, {China}. Association for
  {Computational} {Linguistics}.

\bibitem[\protect\BCAY{Rong}{Rong}{2014}]{rong2014word2vec}
Rong, X. \BBOP2014\BBCP.
\newblock \BBOQ word2vec {Parameter} {Learning} {Explained}\BBCQ\
\newblock {\Bem {arXiv}:1411.2738 {[}cs]}.

\bibitem[\protect\BCAY{Rumelhart, {Hinton},\ \BBA\ {Williams}}{Rumelhart
  et~al.}{1986}]{rumelhart1986learning}
Rumelhart, D.~E., {Hinton}, G.~E., \BBA\ {Williams}, R.~J. \BBOP1986\BBCP.
\newblock \BBOQ Learning representations by back-propagating errors\BBCQ\
\newblock {\Bem Nature}, {\Bem 323\/}(6088), 533--536.

\bibitem[\protect\BCAY{Santos\ \BBA\ {Zadrozny}}{Santos\ \BBA\
  {Zadrozny}}{2014}]{santos2014learning}
Santos, C.~D.\BBACOMMA\  \BBA\ {Zadrozny}, B. \BBOP2014\BBCP.
\newblock \BBOQ Learning {Character}-level {Representations} for
  {Part}-of-{Speech} {Tagging}\BBCQ.
\newblock \BPGS\ 1818--1826.

\bibitem[\protect\BCAY{Schuster\ \BBA\ {Paliwal}}{Schuster\ \BBA\
  {Paliwal}}{1997}]{schuster1997bidirectional}
Schuster, M.\BBACOMMA\  \BBA\ {Paliwal}, K.~K. \BBOP1997\BBCP.
\newblock \BBOQ Bidirectional recurrent neural networks\BBCQ\
\newblock {\Bem {IEEE} {Transactions} on {Signal} {Processing}}, {\Bem
  45\/}(11), 2673--2681.

\bibitem[\protect\BCAY{Shawe-{Taylor}\ \BBA\ {Cristianini}}{Shawe-{Taylor}\
  \BBA\ {Cristianini}}{2004}]{shawe-taylor2004kernel}
Shawe-{Taylor}, J.\BBACOMMA\  \BBA\ {Cristianini}, N. \BBOP2004\BBCP.
\newblock {\Bem Kernel {Methods} for {Pattern} {Analysis}}.
\newblock Cambridge {University} {Press}.

\bibitem[\protect\BCAY{Smith}{Smith}{2011}]{smith2011linguistic}
Smith, N.~A. \BBOP2011\BBCP.
\newblock {\Bem Linguistic {Structure} {Prediction}}.
\newblock Synthesis {Lectures} on {Human} {Language} {Technologies}. Morgan and
  {Claypool}.

\bibitem[\protect\BCAY{Socher}{Socher}{2014}]{socher2014recursive}
Socher, R. \BBOP2014\BBCP.
\newblock {\Bem Recursive {Deep} {Learning} {For} {Natural} {Language}
  {Processing} and {Computer} {Vision}}.
\newblock Ph.D.\ thesis, Stanford {University}.

\bibitem[\protect\BCAY{Socher, {Bauer}, {Manning},\ \BBA\ {Andrew}~{Y}.}{Socher
  et~al.}{2013}]{socher2013parsing}
Socher, R., {Bauer}, J., {Manning}, C.~D., \BBA\ {Andrew}~{Y}., N.
  \BBOP2013\BBCP.
\newblock \BBOQ Parsing with {Compositional} {Vector} {Grammars}\BBCQ\
\newblock In {\Bem Proceedings of the 51st {Annual} {Meeting} of the
  {Association} for {Computational} {Linguistics} ({Volume} 1: {Long}
  {Papers})}, \BPGS\ 455--465, Sofia, {Bulgaria}. Association for
  {Computational} {Linguistics}.

\bibitem[\protect\BCAY{Socher, {Huval}, {Manning},\ \BBA\ {Ng}}{Socher
  et~al.}{2012}]{socher2012semantic}
Socher, R., {Huval}, B., {Manning}, C.~D., \BBA\ {Ng}, A.~Y. \BBOP2012\BBCP.
\newblock \BBOQ Semantic {Compositionality} through {Recursive}
  {Matrix}-{Vector} {Spaces}\BBCQ\
\newblock In {\Bem Proceedings of the 2012 {Joint} {Conference} on {Empirical}
  {Methods} in {Natural} {Language} {Processing} and {Computational} {Natural}
  {Language} {Learning}}, \BPGS\ 1201--1211, Jeju {Island}, {Korea}.
  Association for {Computational} {Linguistics}.

\bibitem[\protect\BCAY{Socher, {Lin}, {Ng},\ \BBA\ {Manning}}{Socher
  et~al.}{2011}]{socher2011parsing}
Socher, R., {Lin}, C. C.-Y., {Ng}, A.~Y., \BBA\ {Manning}, C.~D.
  \BBOP2011\BBCP.
\newblock \BBOQ Parsing {Natural} {Scenes} and {Natural} {Language} with
  {Recursive} {Neural} {Networks}\BBCQ\
\newblock In Getoor, L.\BBACOMMA\  \BBA\ {Scheffer}, T.\BEDS, {\Bem Proceedings
  of the 28th {International} {Conference} on {Machine} {Learning}, {ICML}
  2011, {Bellevue}, {Washington}, {USA}, {June} 28 - {July} 2, 2011}, \BPGS\
  129--136. Omnipress.

\bibitem[\protect\BCAY{Socher, {Manning},\ \BBA\ {Ng}}{Socher
  et~al.}{2010}]{socher2010learning}
Socher, R., {Manning}, C., \BBA\ {Ng}, A. \BBOP2010\BBCP.
\newblock \BBOQ Learning {Continuous} {Phrase} {Representations} and
  {Syntactic} {Parsing} with {Recursive} {Neural} {Networks}\BBCQ\
\newblock In {\Bem Proceedings of the {Deep} {Learning} and {Unsupervised}
  {Feature} {Learning} {Workshop} of \{{NIPS}\} 2010}, \BPGS\ 1--9.

\bibitem[\protect\BCAY{Socher, {Perelygin}, {Wu}, {Chuang}, {Manning}, {Ng},\
  \BBA\ {Potts}}{Socher et~al.}{2013}]{socher2013recursive}
Socher, R., {Perelygin}, A., {Wu}, J., {Chuang}, J., {Manning}, C.~D., {Ng},
  A., \BBA\ {Potts}, C. \BBOP2013\BBCP.
\newblock \BBOQ Recursive {Deep} {Models} for {Semantic} {Compositionality}
  {Over} a {Sentiment} {Treebank}\BBCQ\
\newblock In {\Bem Proceedings of the 2013 {Conference} on {Empirical}
  {Methods} in {Natural} {Language} {Processing}}, \BPGS\ 1631--1642, Seattle,
  {Washington}, {USA}. Association for {Computational} {Linguistics}.

\bibitem[\protect\BCAY{Sordoni, {Galley}, {Auli}, {Brockett}, {Ji}, {Mitchell},
  {Nie}, {Gao},\ \BBA\ {Dolan}}{Sordoni et~al.}{2015}]{sordoni2015neural}
Sordoni, A., {Galley}, M., {Auli}, M., {Brockett}, C., {Ji}, Y., {Mitchell},
  M., {Nie}, J.-Y., {Gao}, J., \BBA\ {Dolan}, B. \BBOP2015\BBCP.
\newblock \BBOQ {A} {Neural} {Network} {Approach} to {Context}-{Sensitive}
  {Generation} of {Conversational} {Responses}\BBCQ\
\newblock In {\Bem Proceedings of the 2015 {Conference} of the {North}
  {American} {Chapter} of the {Association} for {Computational} {Linguistics}:
  {Human} {Language} {Technologies}}, \BPGS\ 196--205, Denver, {Colorado}.
  Association for {Computational} {Linguistics}.

\bibitem[\protect\BCAY{Sundermeyer, {Alkhouli}, {Wuebker},\ \BBA\
  {Ney}}{Sundermeyer et~al.}{2014}]{sundermeyer2014translation}
Sundermeyer, M., {Alkhouli}, T., {Wuebker}, J., \BBA\ {Ney}, H. \BBOP2014\BBCP.
\newblock \BBOQ Translation {Modeling} with {Bidirectional} {Recurrent}
  {Neural} {Networks}\BBCQ\
\newblock In {\Bem Proceedings of the 2014 {Conference} on {Empirical}
  {Methods} in {Natural} {Language} {Processing} ({EMNLP})}, \BPGS\ 14--25,
  Doha, {Qatar}. Association for {Computational} {Linguistics}.

\bibitem[\protect\BCAY{Sundermeyer, {Schl}{\"u}ter,\ \BBA\ {Ney}}{Sundermeyer
  et~al.}{2012}]{sundermeyer2012lstm}
Sundermeyer, M., {Schl}{\"u}ter, R., \BBA\ {Ney}, H. \BBOP2012\BBCP.
\newblock \BBOQ {LSTM} {Neural} {Networks} for {Language} {Modeling}.\BBCQ\
\newblock In {\Bem {INTERSPEECH}}.

\bibitem[\protect\BCAY{Sutskever, {Martens}, {Dahl},\ \BBA\ {Hinton}}{Sutskever
  et~al.}{2013}]{sutskever2013importance}
Sutskever, I., {Martens}, J., {Dahl}, G., \BBA\ {Hinton}, G. \BBOP2013\BBCP.
\newblock \BBOQ On the importance of initialization and momentum in deep
  learning\BBCQ\
\newblock In {\Bem Proceedings of the 30th international conference on machine
  learning ({ICML}-13)}, \BPGS\ 1139--1147.

\bibitem[\protect\BCAY{Sutskever, {Martens},\ \BBA\ {Hinton}}{Sutskever
  et~al.}{2011}]{sutskever2011generating}
Sutskever, I., {Martens}, J., \BBA\ {Hinton}, G.~E. \BBOP2011\BBCP.
\newblock \BBOQ Generating text with recurrent neural networks\BBCQ\
\newblock In {\Bem Proceedings of the 28th {International} {Conference} on
  {Machine} {Learning} ({ICML}-11)}, \BPGS\ 1017--1024.

\bibitem[\protect\BCAY{Sutskever, {Vinyals},\ \BBA\ {Le}}{Sutskever
  et~al.}{2014}]{sutskever2014sequence}
Sutskever, I., {Vinyals}, O., \BBA\ {Le}, Q. V.~V. \BBOP2014\BBCP.
\newblock \BBOQ Sequence to {Sequence} {Learning} with {Neural}
  {Networks}\BBCQ\
\newblock In Ghahramani, Z., {Welling}, M., {Cortes}, C., {Lawrence}, N.~D.,
  \BBA\ {Weinberger}, K.~Q.\BEDS, {\Bem Advances in {Neural} {Information}
  {Processing} {Systems} 27}, \BPGS\ 3104--3112. Curran {Associates}, {Inc}.

\bibitem[\protect\BCAY{Tai, {Socher},\ \BBA\ {Manning}}{Tai
  et~al.}{2015}]{tai2015improved}
Tai, K.~S., {Socher}, R., \BBA\ {Manning}, C.~D. \BBOP2015\BBCP.
\newblock \BBOQ Improved {Semantic} {Representations} {From}
  {Tree}-{Structured} {Long} {Short}-{Term} {Memory} {Networks}\BBCQ\
\newblock In {\Bem Proceedings of the 53rd {Annual} {Meeting} of the
  {Association} for {Computational} {Linguistics} and the 7th {International}
  {Joint} {Conference} on {Natural} {Language} {Processing} ({Volume} 1: {Long}
  {Papers})}, \BPGS\ 1556--1566, Beijing, {China}. Association for
  {Computational} {Linguistics}.

\bibitem[\protect\BCAY{Tamura, {Watanabe},\ \BBA\ {Sumita}}{Tamura
  et~al.}{2014}]{tamura2014recurrent}
Tamura, A., {Watanabe}, T., \BBA\ {Sumita}, E. \BBOP2014\BBCP.
\newblock \BBOQ Recurrent {Neural} {Networks} for {Word} {Alignment}
  {Model}\BBCQ\
\newblock In {\Bem Proceedings of the 52nd {Annual} {Meeting} of the
  {Association} for {Computational} {Linguistics} ({Volume} 1: {Long}
  {Papers})}, \BPGS\ 1470--1480, Baltimore, {Maryland}. Association for
  {Computational} {Linguistics}.

\bibitem[\protect\BCAY{Tieleman\ \BBA\ {Hinton}}{Tieleman\ \BBA\
  {Hinton}}{2012}]{tieleman2012lecture}
Tieleman, T.\BBACOMMA\  \BBA\ {Hinton}, G. \BBOP2012\BBCP.
\newblock \BBOQ Lecture 6.5---{RmsProp}: {Divide} the gradient by a running
  average of its recent magnitude\BBCQ\
\newblock COURSERA: Neural Networks for Machine Learning.

\bibitem[\protect\BCAY{{Van de Cruys}}{{Van de
  Cruys}}{2014}]{vandecruys2014neural}
{Van de Cruys}, T. \BBOP2014\BBCP.
\newblock \BBOQ {A} {Neural} {Network} {Approach} to {Selectional} {Preference}
  {Acquisition}\BBCQ\
\newblock In {\Bem Proceedings of the 2014 {Conference} on {Empirical}
  {Methods} in {Natural} {Language} {Processing} ({EMNLP})}, \BPGS\ 26--35,
  Doha, {Qatar}. Association for {Computational} {Linguistics}.

\bibitem[\protect\BCAY{Vaswani, {Zhao}, {Fossum},\ \BBA\ {Chiang}}{Vaswani
  et~al.}{2013}]{vaswani2013decoding}
Vaswani, A., {Zhao}, Y., {Fossum}, V., \BBA\ {Chiang}, D. \BBOP2013\BBCP.
\newblock \BBOQ Decoding with {Large}-{Scale} {Neural} {Language} {Models}
  {Improves} {Translation}\BBCQ\
\newblock In {\Bem Proceedings of the 2013 {Conference} on {Empirical}
  {Methods} in {Natural} {Language} {Processing}}, \BPGS\ 1387--1392, Seattle,
  {Washington}, {USA}. Association for {Computational} {Linguistics}.

\bibitem[\protect\BCAY{Wager, {Wang},\ \BBA\ {Liang}}{Wager
  et~al.}{2013}]{wager2013dropout}
Wager, S., {Wang}, S., \BBA\ {Liang}, P.~S. \BBOP2013\BBCP.
\newblock \BBOQ Dropout {Training} as {Adaptive} {Regularization}\BBCQ\
\newblock In Burges, C. J.~C., {Bottou}, L., {Welling}, M., {Ghahramani}, Z.,
  \BBA\ {Weinberger}, K.~Q.\BEDS, {\Bem Advances in {Neural} {Information}
  {Processing} {Systems} 26}, \BPGS\ 351--359. Curran {Associates}, {Inc}.

\bibitem[\protect\BCAY{Wang, {Xu}, {Xu}, {Liu}, {Zhang}, {Wang},\ \BBA\
  {Hao}}{Wang et~al.}{2015a}]{wang2015semantic}
Wang, P., {Xu}, J., {Xu}, B., {Liu}, C., {Zhang}, H., {Wang}, F., \BBA\ {Hao},
  H. \BBOP2015a\BBCP.
\newblock \BBOQ Semantic {Clustering} and {Convolutional} {Neural} {Network}
  for {Short} {Text} {Categorization}\BBCQ\
\newblock In {\Bem Proceedings of the 53rd {Annual} {Meeting} of the
  {Association} for {Computational} {Linguistics} and the 7th {International}
  {Joint} {Conference} on {Natural} {Language} {Processing} ({Volume} 2:
  {Short} {Papers})}, \BPGS\ 352--357, Beijing, {China}. Association for
  {Computational} {Linguistics}.

\bibitem[\protect\BCAY{Wang, {Liu}, {SUN}, {Wang},\ \BBA\ {Wang}}{Wang
  et~al.}{2015b}]{wang2015predicting}
Wang, X., {Liu}, Y., {SUN}, C., {Wang}, B., \BBA\ {Wang}, X. \BBOP2015b\BBCP.
\newblock \BBOQ Predicting {Polarities} of {Tweets} by {Composing} {Word}
  {Embeddings} with {Long} {Short}-{Term} {Memory}\BBCQ\
\newblock In {\Bem Proceedings of the 53rd {Annual} {Meeting} of the
  {Association} for {Computational} {Linguistics} and the 7th {International}
  {Joint} {Conference} on {Natural} {Language} {Processing} ({Volume} 1: {Long}
  {Papers})}, \BPGS\ 1343--1353, Beijing, {China}. Association for
  {Computational} {Linguistics}.

\bibitem[\protect\BCAY{Watanabe\ \BBA\ {Sumita}}{Watanabe\ \BBA\
  {Sumita}}{2015}]{watanabe2015transitionbased}
Watanabe, T.\BBACOMMA\  \BBA\ {Sumita}, E. \BBOP2015\BBCP.
\newblock \BBOQ Transition-based {Neural} {Constituent} {Parsing}\BBCQ\
\newblock In {\Bem Proceedings of the 53rd {Annual} {Meeting} of the
  {Association} for {Computational} {Linguistics} and the 7th {International}
  {Joint} {Conference} on {Natural} {Language} {Processing} ({Volume} 1: {Long}
  {Papers})}, \BPGS\ 1169--1179, Beijing, {China}. Association for
  {Computational} {Linguistics}.

\bibitem[\protect\BCAY{Weiss, {Alberti}, {Collins},\ \BBA\ {Petrov}}{Weiss
  et~al.}{2015}]{weiss2015structured}
Weiss, D., {Alberti}, C., {Collins}, M., \BBA\ {Petrov}, S. \BBOP2015\BBCP.
\newblock \BBOQ Structured {Training} for {Neural} {Network}
  {Transition}-{Based} {Parsing}\BBCQ\
\newblock In {\Bem Proceedings of the 53rd {Annual} {Meeting} of the
  {Association} for {Computational} {Linguistics} and the 7th {International}
  {Joint} {Conference} on {Natural} {Language} {Processing} ({Volume} 1: {Long}
  {Papers})}, \BPGS\ 323--333, Beijing, {China}. Association for
  {Computational} {Linguistics}.

\bibitem[\protect\BCAY{Werbos}{Werbos}{1990}]{werbos1990backpropagation}
Werbos, P.~J. \BBOP1990\BBCP.
\newblock \BBOQ Backpropagation through time: {What} it does and how to do
  it.\BBCQ\
\newblock {\Bem Proceedings of the {IEEE}}, {\Bem 78\/}(10), 1550 -- 1560.

\bibitem[\protect\BCAY{Weston, {Bordes}, {Yakhnenko},\ \BBA\ {Usunier}}{Weston
  et~al.}{2013}]{weston2013connecting}
Weston, J., {Bordes}, A., {Yakhnenko}, O., \BBA\ {Usunier}, N. \BBOP2013\BBCP.
\newblock \BBOQ Connecting {Language} and {Knowledge} {Bases} with {Embedding}
  {Models} for {Relation} {Extraction}\BBCQ\
\newblock In {\Bem Proceedings of the 2013 {Conference} on {Empirical}
  {Methods} in {Natural} {Language} {Processing}}, \BPGS\ 1366--1371, Seattle,
  {Washington}, {USA}. Association for {Computational} {Linguistics}.

\bibitem[\protect\BCAY{Xu, {Auli},\ \BBA\ {Clark}}{Xu et~al.}{2015}]{xu2015ccg}
Xu, W., {Auli}, M., \BBA\ {Clark}, S. \BBOP2015\BBCP.
\newblock \BBOQ {CCG} {Supertagging} with a {Recurrent} {Neural}
  {Network}\BBCQ\
\newblock In {\Bem Proceedings of the 53rd {Annual} {Meeting} of the
  {Association} for {Computational} {Linguistics} and the 7th {International}
  {Joint} {Conference} on {Natural} {Language} {Processing} ({Volume} 2:
  {Short} {Papers})}, \BPGS\ 250--255, Beijing, {China}. Association for
  {Computational} {Linguistics}.

\bibitem[\protect\BCAY{Yin\ \BBA\ {Sch}{\"u}tze}{Yin\ \BBA\
  {Sch}{\"u}tze}{2015}]{yin2015convolutional}
Yin, W.\BBACOMMA\  \BBA\ {Sch}{\"u}tze, H. \BBOP2015\BBCP.
\newblock \BBOQ Convolutional {Neural} {Network} for {Paraphrase}
  {Identification}\BBCQ\
\newblock In {\Bem Proceedings of the 2015 {Conference} of the {North}
  {American} {Chapter} of the {Association} for {Computational} {Linguistics}:
  {Human} {Language} {Technologies}}, \BPGS\ 901--911, Denver, {Colorado}.
  Association for {Computational} {Linguistics}.

\bibitem[\protect\BCAY{Zaremba, {Sutskever},\ \BBA\ {Vinyals}}{Zaremba
  et~al.}{2014}]{zaremba2014recurrent}
Zaremba, W., {Sutskever}, I., \BBA\ {Vinyals}, O. \BBOP2014\BBCP.
\newblock \BBOQ Recurrent {Neural} {Network} {Regularization}\BBCQ\
\newblock {\Bem {arXiv}:1409.2329 {[}cs]}.

\bibitem[\protect\BCAY{Zeiler}{Zeiler}{2012}]{zeiler2012adadelta}
Zeiler, M.~D. \BBOP2012\BBCP.
\newblock \BBOQ {ADADELTA}: {An} {Adaptive} {Learning} {Rate} {Method}\BBCQ\
\newblock {\Bem {arXiv}:1212.5701 {[}cs]}.

\bibitem[\protect\BCAY{Zeng, {Liu}, {Lai}, {Zhou},\ \BBA\ {Zhao}}{Zeng
  et~al.}{2014}]{zeng2014relation}
Zeng, D., {Liu}, K., {Lai}, S., {Zhou}, G., \BBA\ {Zhao}, J. \BBOP2014\BBCP.
\newblock \BBOQ Relation {Classification} via {Convolutional} {Deep} {Neural}
  {Network}\BBCQ\
\newblock In {\Bem Proceedings of {COLING} 2014, the 25th {International}
  {Conference} on {Computational} {Linguistics}: {Technical} {Papers}}, \BPGS\
  2335--2344, Dublin, {Ireland}. Dublin {City} {University} and {Association}
  for {Computational} {Linguistics}.

\bibitem[\protect\BCAY{Zhou, {Zhang}, {Huang},\ \BBA\ {Chen}}{Zhou
  et~al.}{2015}]{zhou2015neural}
Zhou, H., {Zhang}, Y., {Huang}, S., \BBA\ {Chen}, J. \BBOP2015\BBCP.
\newblock \BBOQ {A} {Neural} {Probabilistic} {Structured}-{Prediction} {Model}
  for {Transition}-{Based} {Dependency} {Parsing}\BBCQ\
\newblock In {\Bem Proceedings of the 53rd {Annual} {Meeting} of the
  {Association} for {Computational} {Linguistics} and the 7th {International}
  {Joint} {Conference} on {Natural} {Language} {Processing} ({Volume} 1: {Long}
  {Papers})}, \BPGS\ 1213--1222, Beijing, {China}. Association for
  {Computational} {Linguistics}.

\bibitem[\protect\BCAY{Zhu, {Qiu}, {Chen},\ \BBA\ {Huang}}{Zhu
  et~al.}{2015a}]{zhu2015reranking}
Zhu, C., {Qiu}, X., {Chen}, X., \BBA\ {Huang}, X. \BBOP2015a\BBCP.
\newblock \BBOQ {A} {Re}-ranking {Model} for {Dependency} {Parser} with
  {Recursive} {Convolutional} {Neural} {Network}\BBCQ\
\newblock In {\Bem Proceedings of the 53rd {Annual} {Meeting} of the
  {Association} for {Computational} {Linguistics} and the 7th {International}
  {Joint} {Conference} on {Natural} {Language} {Processing} ({Volume} 1: {Long}
  {Papers})}, \BPGS\ 1159--1168, Beijing, {China}. Association for
  {Computational} {Linguistics}.

\bibitem[\protect\BCAY{Zhu, {Sobhani},\ \BBA\ {Guo}}{Zhu
  et~al.}{2015b}]{zhu2015long}
Zhu, X., {Sobhani}, P., \BBA\ {Guo}, H. \BBOP2015b\BBCP.
\newblock \BBOQ Long {Short}-{Term} {Memory} {Over} {Tree} {Structures}\BBCQ\
\newblock {\Bem {arXiv}:1503.04881 {[}cs]}.

\end{thebibliography}

\end{document}